\documentclass{article}

\usepackage[preprint, nonatbib]{neurips_2026}

\usepackage[T1]{fontenc}
\usepackage[english]{babel}

\usepackage[utf8]{inputenc}

\usepackage{geometry}

\usepackage{graphicx}

\usepackage{booktabs}
\usepackage{array}
\usepackage{paralist}
\usepackage{verbatim}
\usepackage{subfig}

\usepackage{csquotes}
\usepackage{graphicx}
\usepackage{float}
\usepackage{placeins}
\usepackage{amssymb}
\usepackage{amsmath, amsfonts, bbm, dsfont}
\usepackage[all]{xy}
\usepackage{MnSymbol}
\usepackage{soul}

\usepackage{tikzit}

\usepackage{fancyhdr}
\usepackage{sectsty}
\allsectionsfont{\sffamily\mdseries\upshape}

\usepackage[nottoc,notlof,notlot]{tocbibind}
\usepackage[titles,subfigure]{tocloft}

\usepackage[numbers]{natbib}

\usepackage{mdframed}

\usepackage{amsthm}
\usepackage{tikz-cd}
\usepackage{appendix}

\tikzcdset{row sep/normal= 0.8 cm,
column sep/normal= 0.8 cm}

\newtheorem{theo}{Theorem}[section]
\newtheorem{theorem}[theo]{Theorem}

\newtheorem{lemma}[theo]{Lemma}
\newtheorem{coro}[theo]{Corollary}

\theoremstyle{definition}

\newtheorem{remark}[theo]{Remark}

\newtheorem{example}[theo]{Example}

\newtheorem{contre-ex}[theo]{Counter-example}

\usepackage{hyperref}
\hypersetup{
    colorlinks=true,
    linkcolor=black,
    filecolor=black,      
    urlcolor=black,
    citecolor=black,
}

\usepackage{tikz}
\usetikzlibrary{intersections}
\usepackage{quiver}

\DeclareMathOperator{\Rr}{\mathbb{R}}
\DeclareMathOperator{\Zz}{\mathbb{Z}}

\renewcommand{\phi}{\varphi}

\usepackage{ifthen}
\newboolean{showcomments}
\setboolean{showcomments}{true}
\ifthenelse{\boolean{showcomments}}
{ \newcommand{\mynote}[3]{
		\fbox{\bfseries\sffamily\scriptsize#1}
		{\small$\blacktriangleright$\textsf{\emph{\color{#3}{#2}}}$\blacktriangleleft$}}
	\newcommand{\zzz}[1]{{\setlength{\fboxsep}{2pt}\fcolorbox{black}{yellow}{\textsf{\emph{#1}}}}\xspace}}
{ \newcommand{\mynote}[3]{}
	\newcommand{\zzz}[1]{}}

\title{The Evaluation Game: 

Beyond Static LLM Benchmarking}

\author{%
  Paul Wang$^{1, 3}$\ Jade Garcia Bourrée$^{2}$\ Anne-Marie Kermarrec$^{2}$\ Vincent Corruble$^{1}$ \\
  \normalfont $^{1}$Sorbonne Université, CNRS, LIP6 \\
  \normalfont $^{2}$École Polytechnique Fédérale de Lausanne
}

\date{}

\begin{document}
	\setcounter{tocdepth}{2}
	\maketitle
	\footnotetext[3]{Supported by a CoefficientGiving grant.}
	\begin{abstract}

		As jailbreaks, adversarially crafted inputs that bypass safety constraints, continue to be discovered in Large Language Models, practitioners increasingly rely on fine-tuning as a defensive strategy. Yet the theoretical foundations underlying this robustness fine-tuning remain underexplored. We introduce a game-theoretic framework in which the interaction between an evaluator (auditing the model for jailbreaks) and a trainer is formalized as a two-player game.  
		A key feature of our approach is the use of \emph{group actions}, a mathematical structure that captures symmetries and transformations, to formally represent data augmentation. The simplest non-trivial instance is the circle with cyclic translation groups, where we exhibit various regimes depending on the trainer's generalization range $\varepsilon$. Below a critical threshold, the evaluator maintains a constant miss ratio for linearly many rounds, whereas other settings can yield very different behaviors. We further provide empirical evidence supporting locality-dependence of the model: for the three model families we tested (Llama, Qwen and Mistral), we have significant evidence that fine-tuning on adversarial prompts induces only local generalization, with refusal rates on test examples highly correlated with the distance to the fine-tuning prompts. Our framework recasts the central object of adversarial evaluation: a benchmark is not a static set of prompts but an orbit under the evaluator's group action, and audit protocols that ignore trainer-side adaptation cannot distinguish a genuine fix from a memorized patch.
	\end{abstract}

\section{Introduction}
Large language models (LLMs) remain essentially black boxes whose behavior, successes or failures, is largely unexplained and misunderstood. The ability to detect and patch vulnerabilities experimentally is therefore essential. However, new jailbreaks, adversarial inputs that bypass safety constraints, continue to be discovered as deployment scales~\cite{peignelefebvre2025bet, panfilov2026claudini}. A common approach to improving LLM safety is known as fine-tuning, including fine-tuning on discovered failures: when a jailbreak is identified, the model is fine-tuned on the offending input to suppress it~\cite{ouyang2022training, bai2022training}. This practice, which aims to eliminate incorrect behavior on these specific malicious inputs, is now widespread, yet its theoretical foundations remain underexplored.

As the space of possible LLM inputs is unbounded, fine-tuning on a finite corpus of known jailbreaks cannot, in principle, provide formal coverage guarantees over the full input distribution. The standard response has been to rely on safety benchmarks~\cite{chu2025jailbreakradar, peignelefebvre2025bet}, i.e., curated sets of adversarial prompts used to measure model robustness at a fixed point in time. These benchmarks have driven significant empirical progress, but they share a common limitation: they assume a passive trainer. A passive trainer fine-tunes a fixed list of vulnerabilities and does not adaptively respond to evaluation queries. In practice however, the trainer observes evaluation queries and can fine-tune against them directly. Moreover, those benchmarks are often public: a model may have already been exposed to them during pretraining or fine-tuning.
Thus, a single evaluation on a static or public benchmark cannot distinguish a genuine fix from a patch that merely memorizes the tested input. This limitation is already identified in different contexts~\cite{li2023beyond, bourree2025robust, tang2026dsgbench}, but not yet formalized for LLM safety.

The core difficulty is that the input space is unbounded, and neither the evaluator nor the trainer can exhaustively explore it. An evaluator that re-uses the queries previously submitted in earlier rounds gives the trainer an easy target: a trainer that fine-tunes narrowly may leave large regions of the threat space uncovered. Formalizing this tension requires modeling how to generate new queries from previous ones, an issue that static benchmarks, by design, do not address.

To address this issue, we propose a formal framework that explicitly models the coevolution between query generation and fine-tuning, capturing the adaptive interplay that static benchmarks structurally cannot. The setting is a two-player game between an evaluator, who audits the system with adversarial queries, and a trainer, who observes those queries and fine-tunes the model before the next round. The evaluator wins if their queries jailbreak across many rounds; the trainer wins if they can cover the evaluator's queries quickly enough. This formulation captures the strategic feedback loop that static benchmarks ignore.

We model the \emph{data augmentation} available to each player as group actions. Data augmentation is the way each side constructs new queries from previous ones. A group action is a mathematical structure that captures symmetries and transformations on the query space. We treat the group-action representation as a modeling assumption that is meant to cover all possible strategies to create new data. It is supported empirically in Section \ref{sec:prompt-generation}. We also include an empirical check of the locality assumption underlying the $\varepsilon$-coverage model.
Our main contributions are as follows.

1) \textbf{A game-theoretic framework for adversarial evaluation.} We introduce the Evaluation Game in Section \ref{sect_eval_game}.
Unlike static benchmarks, which audit a fixed model against a fixed query set and so cannot address the dynamic adaptation that follows a discovered vulnerability, the framework explicitly accounts for the trainer's ability to fine-tune on evaluation queries between rounds.

2) \textbf{Group-theoretic data augmentation in adversarial settings.} We equip each player with a group of transformations acting on the query space, providing a unified and principled model of data augmentation in Section \ref{sect_eval_game}.
To our knowledge, this is the first application of group-theoretic augmentation to prompt spaces in the LLM safety setting.

3)  \textbf{Sharp phase-transition on circle translation game.} We work out representative circle translation games, the simplest nontrivial instance of the framework. With cyclic translation groups of orders $p$ and $q$, we identify in Section \ref{sec:examples} a critical generalization range $\varepsilon^\star = \gcd(p,q)/(pq)$. Below $\varepsilon^\star$, the evaluator sustains a constant success rate for a linear number of turns. The dependency on $\gcd(p,q)$ implies that nearly equal integers can yield very different behaviors.

4) \textbf{Empirical support for the Evaluation Game's modeling assumptions.}
First, we show in Section~\ref{sec_empirical_distance_transfer} that fine-tuning on adversarial prompts induces refusal transfer that varies with embedding distance to the training set, supporting the $\varepsilon$-coverage assumption. Second, the panel of prompt transformations is well-approximated by linear operators whose fitted parameters are consistent with the algebraic relations expected of a group action, with the self-inverse primitive acting as a partial involution (Section~\ref{sec:prompt-generation}). Together, these results support the framework's two core modeling assumptions on real LLM data, making its conclusions actionable for real audit protocols.

Beyond jailbreaks, our framework applies to any black-box audit setting (e.g., fairness or privacy audits of classifiers or ranking systems) where the audited model may adapt to the auditor's queries.

\section{Background and related work}
\textbf{Jailbreaks as Adversarial Attacks on LLMs.}
Building on the foundational vocabulary of adversarial robustness~\cite{szegedy2013intriguing, goodfellow2014explaining} and its early extensions to NLP~\cite{jia2017adversarial, wallace2019universal}, \emph{jailbreaking} has emerged as a practically significant instance.
Wei et al.~\cite{wei2023jailbroken} analyze systematically how safety training can fail, Zou et al.~\cite{zou2023universal} demonstrate that such attacks can be automated with gradient-based optimization, and Shen et al.~\cite{shen2024anything} study how in-the-wild jailbreaks emerge and evolve continuously over time. More recently, Chu et al.~\cite{chu2025jailbreakradar} provide a comprehensive empirical assessment of the attack landscape, while Cui et al.~\cite{cui2025exploring} investigate a distinct attack vector based on intent concealment and semantic diversion, together illustrating the breadth and continued evolution of jailbreak methodologies. However, these works remain largely empirical and do not formally model the interaction between attacker and defender as a dynamic process, leaving open the theoretical question of whether robustness fine-tuning reliably reduces vulnerabilities or merely obscures them from evaluation.

\textbf{Attacking and Defending Aligned LLMs.}
Fine-tuning, and in particular Reinforcement Learning from Human Feedback (RLHF)~\cite{christiano2017deep, ouyang2022training, bai2022training}, is the standard paradigm for aligning LLMs with safety objectives.
Perez et al.~\cite{perez2022red} anticipate attacker-side adaptivity by using an LLM to iteratively generate adversarial prompts. Building on this line, Peigné-Lefebvre et al.~\cite{peignelefebvre2025bet} report near-universal vulnerability (100\% Attack Success Rate against 37 of 41 state-of-the-art LLMs) via Dynamic Adversarial Optimization over a library of jailbreak primitives, with attack difficulty varying over 300-fold across models. Furthermore, robustness fine-tuning empirically exhibits accuracy--robustness trade-offs~\cite{li2025robustness}, can be undone by adversarial fine-tuning~\cite{qi2023fine, qi2024shallow}, and collapses when alignment and fine-tuning data are too similar~\cite{Hsiung_Pang_Tang_Song_2026}. These findings motivate a formal framework to characterize when fine-tuning succeeds or fails.

\textbf{Game Theory and Adversarial Learning.}
Madry et al.~\cite{madry2017towards} formulate the problem of training robust classifiers as a minimax optimization problem, providing both a theoretical grounding and a practical training algorithm.
The theoretical foundations of adversarial learning are closely related to online learning~\cite{littlestone1994weighted, shalev2012online, hazan2016introduction}, where an agent makes sequential decisions against an adversarial environment.
A more recent line of work studies iterated retraining as a strategic phenomenon. Strategic classification~\cite{Hardt_Megiddo_Papadimitriou_Wootters_2016, Dong_Roth_Schutzman_Waggoner_Wu_2018} models the case in which data points adapt to a deployed classifier under a metric cost on input space, with one-shot Stackelberg~\cite{Hardt_Megiddo_Papadimitriou_Wootters_2016} and online-regret~\cite{Dong_Roth_Schutzman_Waggoner_Wu_2018} variants. Performative prediction~\cite{Perdomo_Zrnic_MendlerDunner_Hardt_2020} generalizes this to a fixed-point formulation in which the deployment-induced distribution shift is the central object.
However, none of these works models the specific interaction between a trainer performing robustness fine-tuning and an evaluator probing for jailbreaks, nor do they connect these dynamics to group-theoretic data augmentation.
This gap, both in terms of application domain and mathematical structure, is precisely what this article addresses.

\textbf{Multi-round adversarial benchmarking and dynamic evaluation.}
Dwork et al.~\cite{Dwork_Feldman_Hardt_Pitassi_Reingold_Roth_2015} initiated the formal study of adaptive data analysis, showing that an analyst issuing statistical queries against a held-out dataset can preserve validity for exponentially many queries using differential-privacy mechanisms. Blum and Hardt~\cite{Blum_Hardt_2015} introduced the Ladder, a defender-side mechanism bounding the catastrophic-overfitting risk of public ML leaderboards. 
On the empirical-LLM-safety side, MART~\cite{Ge_Zhou_Hou_Khabsa_2024} iterates attack generation and safety fine-tuning, while Ma et al.~\cite{Ma_Yang_Ci_Gao_Gao_Pan_Yang_2023} provide the closest formal framework via the Red Teaming Game. Most directly, Shirali et al.~\cite{Shirali_Abebe_Hardt_2023} give the only published theory of iterative human-and-model benchmarking: under non-strategic sampling, performance stalls in three rounds. 
Our framework differs from this setting in three ways: it endows the evaluator with a strategy class structured by a group action, replaces the scalar-loss objective with $\varepsilon$-coverage of a held-out adversarial set on a metric query space, and adopts a sustained-margin victory predicate over a count of rounds rather than asymptotic stalling.

\textbf{Active auditing.} A related line of research considers the problem of actively auditing a model by querying it to construct a certificate attesting to a given property. Chugg et al.~\cite{chugg2023auditing} introduce a betting-based framework for auditing fairness, while Ajarra et al.~\cite{ajarra2025active} propose a Fourier-based active auditor for estimating distributional properties of black box models. Hartmann et al.~\cite{hartmann2026audit} extend this active auditing paradigm explicitly to LLMs. A distinguishing feature of this setting is that the target model may adapt to the auditor's queries, and the goal is to characterize the best possible guarantees for the auditor in the worst case. Yan and Zhang~\cite{yan2022active} study this problem under the assumption that the model is fixed, while Godinot et al.~\cite{godinot2023change} relax this assumption to allow for limited model change. However, these works remain limited to low-dimension classifiers and do not connect the auditing dynamics to a game-theoretic framework grounded in the structure of the training process. The evaluator in our trainer-evaluator game plays an analogous role: probing a model sequentially to characterize its worst-case behavior, now in the high-dimensional setting of LLM prompt spaces and within a principled game-theoretic framework.

Taken together, these works establish the empirical significance of jailbreaking, the dominance of robustness fine-tuning as a defense, and the theoretical tools available from group-theoretic data augmentation, game theory, and active auditing. However, no prior work combines these threads in a group-action coverage framework.  The present work addresses this gap by modeling the trainer-evaluator interaction as a two-player game grounded in group actions, with implications that extend beyond jailbreaking to any adversarial evaluation objective.

\section{Framework: evaluation games}\label{sect_eval_game}
Consider an evaluator tasked with auditing an AI system for safety failures~\cite{shen2024anything} and a trainer who improves the system. We now state the three assumptions that define the scope of the model. 
First, the evaluator operates in a black box setting~\cite{bunge1963general, perez2022red}: they
cannot observe the trainer's internals or strategy, whereas the trainer has
full access to every query the evaluator sends -- though not to those they keep secret. This asymmetry reflects the typical
deployment context, in which the evaluator is external to the organization
maintaining the model~\cite{hartmann2026audit}.
Second, the trainer fine-tunes the model using their own data or by including the queries that were just sent by the evaluator. 
Third, we assume in this model that neither player generates genuinely public data or new data from scratch: each constructs new inputs by transforming existing ones. This reflects current practice in adversarial research, where attacks are often built by combining existing jailbreak primitives~\cite{panfilov2026claudini}.
In this setting, the query space consists of test prompts embedded in the model's representation space.

We first describe the details of a round of the game, then specify the trainer's per-round budget, the geometric notion of coverage, the diagnostic used to declare a winner, and the victory predicate. Section~\ref{ss:group_actions_introduction} then specializes the framework by modeling each player's data construction as a group action.

\subsection{Game description}\label{defi_eval_games}
The game proceeds in rounds. At each round $n$, the evaluator $E$ selects a
batch of queries $E_n$ and sends it to the trainer's system $T$. The trainer
then chooses how to expand its training dataset: either by directly incorporating
queries from $E_n$, or by applying transformations to existing data.
After this expansion, the trainer fine-tunes the system on the updated dataset.

The evaluator starts from a finite initial dataset $D^{\mathrm{eval}}_0 \subseteq X$;
the trainer starts from $D^{\mathrm{train}}_0 = \emptyset$. As the covering dynamics depend on how many transformations the trainer may apply per round from a set of transformations $H_{\mathrm{train}}$ and how broadly each is applied, we now isolate three moves the trainer can play at round $n$. The transformations in $H_{\mathrm{train}}$ are defined in Section \ref{ss:group_actions_introduction}.

\textbf{Single-augmentation.} The trainer picks one $h_n \in H_{\mathrm{train}}$ and one $c_n \in D^{\mathrm{train}}_n \cup E_n$; set $D^{\mathrm{train}}_{n+1} = D^{\mathrm{train}}_n \cup E_n \cup \{h_n \cdot c_n\}$. That is to say, the trainer shifts at most one query per round.

\textbf{Batch-augmentation.} The trainer picks one $h_n \in H_{\mathrm{train}}$ and any subset $C_n \subseteq D^{\mathrm{train}}_n \cup E_n$; set $D^{\mathrm{train}}_{n+1} = D^{\mathrm{train}}_n \cup E_n \cup h_n \cdot C_n$. The trainer applies one type of transformation per training round but might apply it to the whole stored corpus at once. This matches typical fine-tuning practice (e.g., paraphrase every existing prompt at once, rather than one prompt at a time).

\textbf{Full-augmentation.} The trainer uses all the transformations on all the queries: $D^{\mathrm{train}}_{n+1} = H_{\mathrm{train}} \cdot (D^{\mathrm{train}}_n \cup E_n)$. Once E's queries are absorbed, the trainer's coverage spreads to
every $H$-translate of it.

We refer to these three moves as $\mathrm{single}$, $\mathrm{batch}$, and $\mathrm{full}$ throughout. We use the placeholder superscript $\bullet \in \{\mathrm{single}, \mathrm{batch}, \mathrm{full}\}$.

To model possible generalization beyond exact matches, we represent the trainer's \emph{coverage} by the union of $\varepsilon$-neighborhoods: 
\begin{equation}\label{eq:eps_neighbour}
    V_\varepsilon(D^{\mathrm{train}}_n) = \bigcup_{s \in D^{\mathrm{train}}_n} B(s,\varepsilon),
\end{equation}
where $\varepsilon > 0$ is the \emph{generalization range} of the fine-tuning
procedure. This $\varepsilon$-neighborhood is central: it abstracts the trainer's generalization capability into a single geometric parameter. Section~\ref{sec_empirical_distance_transfer} empirically validates that this locality assumption holds across models. 

We define a key diagnostic, the \emph{miss ratio} $r_n \in [0,1]$, to measure, round by round, whether the evaluator is still finding queries that escape it.
It is defined as the fraction of queries in $E_n$ that fall outside the
trainer's current coverage:
\[
  r_n =
  \begin{cases}
    \dfrac{|E_n \setminus V_{\varepsilon}(D^{\mathrm{train}}_n)|}{|E_n|},
    & E_n \neq \emptyset,\\[6pt]
    0, & E_n = \emptyset.
  \end{cases}
\]
It is measured at the \textit{beginning} of round $n$ before the trainer expands its dataset. The convention $r_n = 0$ when $E_n = \emptyset$ says that an empty test batch is covered by definition, or equivalently, that E skipped the round and did not extract an uncovered query.

The key question is whether $E$ can keep generating queries that maintain high
$r_n$ values (\emph{i.e.}, queries that fall outside $V_{\varepsilon}(D^{\mathrm{train}}_n)$)
or whether $T$'s data expansion is rich enough to eventually cover the entire
threat space $E$ can reach (leading to low $r_n$ values). That is to say, the question is to determine whether the evaluator can generate uncovered queries that the trainer is not able to superficially fix. 

\textbf{Game victory.}
\textit{Victory} is defined as a property of the finite sequence of ratios
$(r_n)_{n < N}$.
Player $E$ wins if the miss ratio stays high, for instance
$r_n > \rho$ for all $n < N$, where
$\rho \in [0,1]$ is a given threshold.
$T$ wins when the condition is not met.

Since the victory depends on the round, two particular rounds are of interest: the one at which the trainer first neutralizes an incoming query, and the one at which the entire evaluator orbit is absorbed. Under optimal play by T:\\
- $N_{\mathrm{o}}^{\bullet}(\varepsilon, p, q)$ is the first round at
  which $r_n = 0$. That is to say, all of E's next queries are $\varepsilon$-covered.\\
- $N_{\mathrm{orbit}}^{\bullet}(\varepsilon, p, q)$ is the first round at
  which $\Omega_E \subseteq V_\varepsilon(D^{\mathrm{train}}_n)$. It corresponds to the first round when the
  whole evaluator orbit is $\varepsilon$-covered.

In the following, we specify how each player constructs new data from existing inputs and empirically test one locality assumption underlying the model in Section~\ref{sec_empirical_distance_transfer}.

\subsection{Implementation with group actions}\label{ss:group_actions_introduction}
\textbf{Related work.} A principled framework for data augmentation through group theory was developed by Cohen and Welling~\cite{cohen2016group}, Chen et al.~\cite{chen2020group}, and Lyle et al.~\cite{lyle2020benefits}, primarily in the context of computer vision. None of these works considers data augmentation applied strategically by a trainer in response to an adversary, nor do they transfer group-theoretic augmentation to prompt spaces.

We model data construction with group actions. Each player is equipped with a distinct
group of transformations acting on $X$. The group is $G_{\mathrm{eval}}$ for the evaluator
and $H_{\mathrm{train}}$ for the trainer. At round $n$, $E$ applies a
transformation $g_n \in G_{\mathrm{eval}}$ to a subset
$C^{\mathrm{eval}}_n \subseteq D^{\mathrm{eval}}_n$, expanding its dataset to
$D^{\mathrm{eval}}_{n+1} = D^{\mathrm{eval}}_n \cup g_n \cdot C^{\mathrm{eval}}_n$.
It then selects the batch $E_n \subseteq D^{\mathrm{eval}}_{n+1}$ to send to $T$.
Similarly, $T$ applies $h_n \in H_{\mathrm{train}}$ to a subset
$C^{\mathrm{train}}_n \subseteq D^{\mathrm{train}}_n \cup E_n$, and expands to
$D^{\mathrm{train}}_{n+1} = D^{\mathrm{train}}_n \cup h_n \cdot C^{\mathrm{train}}_n$.

More generally, the framework defines a family of games parameterised
by $(X, G_{\mathrm{eval}}, H_{\mathrm{train}}, D^{\mathrm{eval}}_0, \varepsilon, N)$,
and the interaction between $G_{\mathrm{eval}}$ and $H_{\mathrm{train}}$ (\emph{i.e.},
their orbits, their overlap, whether their actions commute) is what
determines the outcome. The properties of the actions in $G_{\mathrm{eval}}$ and $H_{\mathrm{train}}$, and the way they interact, appear crucial. We study in Section \ref{sec:examples} how the features of both groups can lead to different game dynamics.

\section{Theoretical analysis on the simplest non-trivial case}\label{sec:examples}

\textbf{Set up.} Among all instances of the framework, finite-order translations on the circle are the smallest setting in which the evaluator's and trainer's actions can interact non-trivially.
Let $G_{\mathrm{eval}} = \langle 1/p \rangle \cong \Zz/p\Zz$ and
$H_{\mathrm{train}} = \langle 1/q \rangle \cong \Zz/q\Zz$ act on
$\mathbb{T}^1 = \Rr/\Zz$ by translation, where $p, q \geq 1$ are integers.
Write $g = \gcd(p,q)$ and $L = \operatorname{lcm}(p,q) = pq/g$. 

Without loss of generality, we consider $x_0 = 0$, so the E-orbit is
$\Omega_E = \{0, 1/p, 2/p, \ldots, (p{-}1)/p\}.$ 
For simplicity, we assume that E's strategy is a \textit{cyclic walk}: at each turn, they generate one query by applying their primitive translation ${1}/{p}$ and send it to T. So, we have $E_n = \lbrace {n}/{p} \rbrace \subset \mathbb{T}^{1}$ for all rounds $n$. In this setting, we can define two key thresholds: $\varepsilon^\star$, the \textit{sub-critical threshold}, under which E wins for a linear duration, and $\varepsilon_{\mathrm{sat}}$, the \textit{saturation threshold}, over which T can win quickly. We prove in Appendix~\ref{app_circle_prelim} that in this case, $\varepsilon^\star = g/(pq)$ and $\varepsilon_{\mathrm{sat}} = \lfloor p/(2g)\rfloor/L \simeq 1/(2q)$. This shows that victory conditions can depend on fine properties of the parameters of the game. 

\begin{figure}[t]
    \centering
    \scalebox{1.18}{
  \begin{tikzpicture}[font=\small, line cap=round, line join=round, >=stealth]


\def\R{0.90}
\pgfmathsetmacro{\epsdeg}{360*0.08} 

\begin{scope}[shift={(0,0)}]
  \node[font=\small] at (0,1.45) {Round $0$};
  \node[font=\scriptsize] at (0,1.12) {$D^{train}_0=\varnothing$};

  \draw[black!55] (0,0) circle (\R);
  \foreach \m in {0,...,14} {
    \pgfmathsetmacro{\a}{90-24*\m}
    \draw[gray!35] (\a:\R-0.06) -- (\a:\R);
  }
  \foreach \m in {0,3,6,9,12} {
    \pgfmathsetmacro{\a}{90-24*\m}
    \node[circle, fill=blue!70!black, draw=blue!70!black,
          minimum size=3.6pt, inner sep=0pt] at (\a:\R) {};
  }
  \draw[red!75!black, line width=0.9pt] (90:\R) circle (4.2pt);
  \node[font=\scriptsize, red!75!black] at (90:\R+0.32) {$ $};

  \node[font=\scriptsize, red!75!black] at (0,-1.18) {$r_0=1$};
\end{scope}

\begin{scope}[shift={(1.62,0)}]
  \draw[orange!85!black, ->, thick] (-0.45,0) -- (0.45,0);
  \node[font=\scriptsize, anchor=south] at (0,0.08) {$h_0=0$};
  \node[font=\scriptsize, anchor=north, gray!55!black] at (0,-0.08) {absorb};
\end{scope}

\begin{scope}[shift={(3.24,0)}]
  \node[font=\small] at (0,1.45) {Round $1$};
  \node[font=\scriptsize] at (0,1.12) {$D^{train}_1=\{0\}$};

  \fill[orange!35, opacity=0.55]
    (0,0) -- (90-\epsdeg:\R) arc (90-\epsdeg:90+\epsdeg:\R) -- cycle;

  \draw[black!55] (0,0) circle (\R);
  \foreach \m in {0,...,14} {
    \pgfmathsetmacro{\a}{90-24*\m}
    \draw[gray!35] (\a:\R-0.06) -- (\a:\R);
  }
  \foreach \m in {0,3,6,9,12} {
    \pgfmathsetmacro{\a}{90-24*\m}
    \node[circle, fill=blue!70!black, draw=blue!70!black,
          minimum size=3.6pt, inner sep=0pt] at (\a:\R) {};
  }
  \node[circle, fill=black, draw=white, line width=0.3pt,
        minimum size=3.2pt, inner sep=0pt] at (90:\R) {};
  \draw[red!75!black, line width=0.9pt] (90-72:\R) circle (4.2pt);
  \node[font=\scriptsize, red!75!black] at (90-72:\R+0.32) {$ $};

  \node[font=\scriptsize, red!75!black] at (0,-1.18) {$r_1=1$};
\end{scope}

\begin{scope}[shift={(4.86,0)}]
  \draw[orange!85!black, ->, thick] (-0.45,0) -- (0.45,0);
  \node[font=\scriptsize, anchor=south] at (0,0.08) {$h_1=+1/3$};
\end{scope}

\begin{scope}[shift={(6.48,0)}]
  \node[font=\small] at (0,1.45) {Round $2$};
  \node[font=\scriptsize] at (0,1.12) {$D^{train}_2=\{0,\tfrac13,\tfrac{8}{15}\}$};

  \foreach \m in {0,5,8} {
    \pgfmathsetmacro{\ac}{90-24*\m}
    \fill[orange!35, opacity=0.55]
      (0,0) -- (\ac-\epsdeg:\R) arc (\ac-\epsdeg:\ac+\epsdeg:\R) -- cycle;
  }

  \draw[black!55] (0,0) circle (\R);
  \foreach \m in {0,...,14} {
    \pgfmathsetmacro{\a}{90-24*\m}
    \draw[gray!35] (\a:\R-0.06) -- (\a:\R);
  }
  \foreach \m in {0,3,6,9,12} {
    \pgfmathsetmacro{\a}{90-24*\m}
    \node[circle, fill=blue!70!black, draw=blue!70!black,
          minimum size=3.6pt, inner sep=0pt] at (\a:\R) {};
  }
  \foreach \m in {0,5,8} {
    \pgfmathsetmacro{\a}{90-24*\m}
    \node[circle, fill=black, draw=white, line width=0.3pt,
          minimum size=3.2pt, inner sep=0pt] at (\a:\R) {};
  }
  \draw[red!75!black, line width=0.9pt] (90-72:\R) circle (4.2pt);
  \node[font=\scriptsize, red!75!black] at (90-72:\R+0.32) {$ $};

  \node[font=\scriptsize, red!75!black] at (0,-1.18) {$r_2=1$};
\end{scope}

\begin{scope}[shift={(8.10,0)}]
  \draw[orange!85!black, ->, thick] (-0.45,0) -- (0.45,0);
  \node[font=\scriptsize, anchor=south] at (0,0.08) {$h_2=+2/3$};
\end{scope}

\begin{scope}[shift={(9.72,0)}]
  \node[font=\small] at (0,1.45) {Round $3$};
  \node[font=\scriptsize] at (0,1.12) {all $\Omega_E$ covered};

  \foreach \m in {0,3,5,8,13} {
    \pgfmathsetmacro{\ac}{90-24*\m}
    \fill[orange!35, opacity=0.55]
      (0,0) -- (\ac-\epsdeg:\R) arc (\ac-\epsdeg:\ac+\epsdeg:\R) -- cycle;
  }

  \draw[black!55] (0,0) circle (\R);
  \foreach \m in {0,...,14} {
    \pgfmathsetmacro{\a}{90-24*\m}
    \draw[gray!35] (\a:\R-0.06) -- (\a:\R);
  }
  \foreach \m in {0,3,6,9,12} {
    \pgfmathsetmacro{\a}{90-24*\m}
    \node[circle, fill=blue!70!black, draw=blue!70!black,
          minimum size=3.6pt, inner sep=0pt] at (\a:\R) {};
  }
  \foreach \m in {0,3,5,8,13} {
    \pgfmathsetmacro{\a}{90-24*\m}
    \node[circle, fill=black, draw=white, line width=0.3pt,
          minimum size=3.2pt, inner sep=0pt] at (\a:\R) {};
  }

  \node[font=\scriptsize, green!45!black] at (0,-1.18) {$r_3=0$};
\end{scope}

\begin{scope}[shift={(0.35,-1.85)}]
  \node[circle, fill=blue!70!black, draw=blue!70!black,
        minimum size=3.6pt, inner sep=0pt] at (0,0) {};
  \node[anchor=west, font=\scriptsize] at (0.16,0) {$\Omega_E$};

  \node[circle, fill=black, draw=white, line width=0.3pt,
        minimum size=3.2pt, inner sep=0pt] at (1.30,0) {};
  \node[anchor=west, font=\scriptsize] at (1.46,0) {$D^{train}_n$};

  \draw[red!75!black, line width=0.9pt] (3.15,0) circle (4.2pt);
  \node[anchor=west, font=\scriptsize] at (3.31,0) {current query $E_n$};

  \fill[orange!35, opacity=0.55] (5.90,-0.08) rectangle (6.20,0.08);
  \node[anchor=west, font=\scriptsize] at (6.26,0)
    {$V_\varepsilon(D^{train}_n)$, $\varepsilon=0.08$};
\end{scope}

\end{tikzpicture}}
  \caption{Round-by-round trace of  play in the circle example with $p = 5, q = 3$ 
  and
  \(\varepsilon=0.08\). E-orbit \(\Omega_E\) is in orange dots,
  white dots mark the
  accumulated trainer data \(D^{train}_n\), and orange sectors show
  \(V_\varepsilon(D^{train}_n)\). The red open ring is the query \(E_n\) revealed that round; each \(r_n\) is the miss
  against the pre-update coverage. Between panels, the arrow records T's
  chosen action \(h_n\in H_{\mathrm{train}}\); one action per round is the bottleneck.
  In three high-failure rounds (\(r_0{=}r_1{=}r_2{=}1\)) T exploits
  \(|1/3-2/5|=1/15<\varepsilon\) to shift by \(+2\) on \(\Zz/5\Zz\), yielding
  \(r_3=0\).}
  \label{fig_circle_five_three_threshold}
\end{figure}
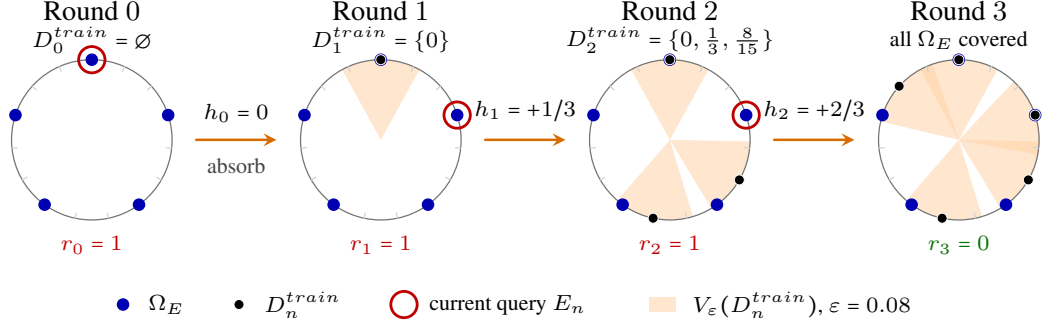

\textbf{Example.} Figure \ref{fig_circle_five_three_threshold} shows the situation with $p = 5$, $q = 3$. With those values, $L = 15$ and
$\varepsilon^\star = 1/15$.
For $\varepsilon < 1/15$, translations are useless and E forces five
uncovered singleton queries before T has absorbed the whole orbit. The miss
ratio sequence is $r_0 = r_1 = \cdots = r_4 = 1$, $r_5 = 0$.
For $\varepsilon > 1/15$, the translation by $1/3$ induces an
approximate shift $i \mapsto i+2$ on the five-point orbit
(since $|1/3 - 2/5| = 1/15 < \varepsilon$), and T can exploit this shift to
finish in three high-failure rounds,
$r_0 = r_1 = r_2 = 1$, $r_3 = 0$. A round-by-round verification is given in
Appendix~\ref{ex_app_circle_five_three}.

\textbf{Phase-diagram tables.} For each possible move, we summarise in Table~\ref{table:boom} the value or magnitude of the particular rounds defined in Section~\ref{sect_eval_game} across three representative $(\varepsilon, p, q)$-regimes relative to the sub-critical $\varepsilon^\star$ and saturation $\varepsilon_{\mathrm{sat}}$ thresholds. The intermediate regime $\varepsilon^\star < \varepsilon < \varepsilon_{\mathrm{sat}}$ is where the arithmetic structure of the two groups interacts non-trivially, in ways that require further investigation beyond the scope of this paper.

\begin{table}[ht]
\centering
\caption{First-neutralisation and orbit-covering times across regimes and per-round augmentation budgets. Here
$Q(\varepsilon,p,q)
=\max_{z\in\Omega_E+H_{\mathrm{train}}}|\Omega_E\cap B(z,\varepsilon)|$
is the reachable-center packing factor, with $B(\cdot,\varepsilon)$ denoting an
open ball in the circle metric.}
\label{table:boom}
\renewcommand{\arraystretch}{1.2}
\begin{tabular}{@{}lcccccc@{}}
\toprule
Regime & Condition & $N_{\mathrm{o}}$ &
$N^{\mathrm{single}}_{\mathrm{orbit}}$ &
$N^{\mathrm{batch}}_{\mathrm{orbit}}$ &
$N^{\mathrm{full}}_{\mathrm{orbit}}$\\
\midrule
Sub-critical, $g = 1$
  & $\varepsilon \le \varepsilon^\star$
 & $p$ & $p$ & $p$ & $p$\\
Sub-critical, $g \ge 2$
  & $\varepsilon \le \varepsilon^\star$
  & $p/g$ & $\lceil p/2 \rceil$ & $\Theta(p/g + \log g)$ & $p/g$\\
Saturation
  & $\varepsilon > \varepsilon_{\mathrm{sat}},\ q \le p$
  & $1$ & $\Theta(p / Q(\varepsilon, p, q))$ & $O(\log p)$ & $1$\\
\bottomrule
\end{tabular}
\end{table}

We demonstrate in Theorem~\ref{thm_npaper_master} the remarkable fact that the cyclic-walk diagnostic is insensitive to T's batching, i.e., the $N_{\mathrm{o}}$ column is identical across all moves.
Intuitively, since E only generates and sends a single predictable query per round, T only needs one well-placed query to force the first zero-miss round, which can be done with single augmentation. In this cyclic-walk example, the predictable evaluator schedule makes $N_{\mathrm{o}}$ insensitive to the trainer's batching budget.

The $N^{\mathrm{single}}_{\mathrm{orbit}}$ column of Table~\ref{table:boom} reports $N^{\mathrm{single}}_{\mathrm{orbit}}$ across the three regimes, illustrating how the simplest data-augmentation budget (e.g., one augmented example produced per round) interacts with the geometry of $(\varepsilon, p, q)$. In the sub-critical regime with $\gcd(p,q) = 1$, translations are useless on the orbit and the trainer is reduced to absorbing E's queries one at a time, yielding $N^{\mathrm{single}}_{\mathrm{orbit}} = p$. Across all three regimes, $N^{\mathrm{single}}_{\mathrm{orbit}}$ remains linear in $p$ (up to the packing factor $Q$): with only one augmentation per round, the trainer cannot escape a per-round throughput proportional to its dataset growth rate, regardless of how favorable the arithmetic of $(p, q)$ becomes.

\textbf{Conclusion.} The circle translation game, although minimal, already reveals how the interaction between the evaluator's and the trainer's transformations shapes the outcome.
First, the generalization range $\varepsilon$ acts in this model as a sharp on/off switch: below the critical threshold $\varepsilon^\star$, the trainer's transformations cannot reach any new evaluator query and coverage reduces to memorizing one prompt at a time, while above it the arithmetic compatibility between the two groups dictates how fast the threat space is covered. Second, when the evaluator follows a predictable cyclic schedule, how aggressively the trainer batches its augmentations does not impact how quickly it neutralizes the next incoming query -- the batching budget only matters once we ask how fast the whole evaluator orbit is covered. Third, when the trainer is restricted to one augmentation per round, exploring $E$'s orbit will take linear time, although the factor can depend on strategic and arithmetic conditions.

\section{Empirical evidence: distance-dependent transfer and prompt transformations}
\label{sec_empirical_distance_transfer}

Our experiments test two distinct aspects of the model: the local nature of fine-tuning-induced generalization (Equation \ref{eq:eps_neighbour}), and the representation of query transformations as actions of group elements (Section \ref{ss:group_actions_introduction}).

\textbf{Datasets.} For both questions, we rely on the WildJailBreak dataset~\cite{jiang2024wildteaming}, which contains hundreds of thousands of prompts. We focus on a subset most relevant to us,
called \texttt{adversarial\_harmful}. The models
are fine-tuned using low-rank (LoRA) adaptation~\cite{hu2022lora} on a small set of jailbreak prompts. For fine-tuning, we also use MMLU~\cite{hendrycks2020measuring} and Wikitext-2~\cite{merity2017pointer} to measure degeneration, as well as a \texttt{vanilla\_benign} subset of WildJailbreak to measure overrefusal.

\textbf{Models.} For the question of fine-tuning generalization, we fine-tune three open-weight models: \textsc{Llama-3.1-8B-Instruct}~\cite{grattafiori2024llama3}, \textsc{Mistral-7B-Instruct-v0.3}~\cite{jiang2023mistral}, and \textsc{Qwen2.5-7B-Instruct}~\cite{qwen2024qwen25}, and use \textsc{OLMo-3.1-32B-Instruct}~\cite{teamolmo2025olmo3} for judging answers. For the prompt transformations experiments, we additionally use 
\texttt{gpt2} (0.12B parameters, base)~\cite{radford2019language}, \texttt{pythia-410m} and \texttt{pythia-1b}~\cite{biderman2023pythia}.

\subsection{Distance-dependent transfer of refusal training}
\label{sec:distance-dependent-transfer}

\textbf{Setup.}
We evaluate the base and fine-tuned models on a held-out panel of jailbreak prompts and ask whether the refusal rate of the fine-tuned model $p_{\mathrm{ft}}(z)$ depends on the embedding-space distance $z$ from each test prompt to its nearest training prompt. We use a $47$-prompt training pool selected by KNN density in embedding space, and a $2999$-prompt held-out evaluation panel stratified along several embedding metrics; this controls for distribution coverage so that the conditional distance $z$ varies across a meaningful range (Appendix~\ref{app:training-pool}--\ref{app:eval-panel}). Each (model, seed) pair is one cell, and our main experiments train each family at LoRA rank $1$, exposure $1600$, learning rate $3\times 10^{-5}$.

\textbf{Distance and refusal rate.}
For each test prompt $x$ we define $z(x)$ as the minimum Euclidean distance between an embedding of $x$ and the embeddings of training prompts, computed under several pooling schemes. The description of the pooling schemes and exact distance definition are provided in Appendix~\ref{app:distance-methods}. For each model and prompt, we sample replies $N=10$ times. Refusal labels for individual responses are then obtained from a single open-weight LLM-as-judge (\textsc{OLMo-3.1-32B-Instruct}) classifying each response as \emph{refuses}, \emph{complies}, or \emph{ambiguous}, yielding per-prompt rates for both base and fine-tuned models; ambiguous slots are dropped (Appendix~\ref{app:judge}).

\textbf{Statistical model.}
We model the fine-tuned-rate $p_{\mathrm{ft}}(z) = \Pr(R_{\mathrm{ft}} \!=\! 1 \mid z) \in [0, 1]$, fitting a small \emph{menu} of candidate shapes for $p_{\mathrm{ft}}$ (constant; linear; plateau-then-slope; Constrained Piecewise-Linear (CPL); two-segment piecewise-linear). The CPL shape, with parameters $(z_L, z_R, d_{\mathrm{lo}}, d_{\mathrm{hi}})$, linearly interpolates between $d_{\mathrm{lo}}$ on $[0, z_L]$ and $d_{\mathrm{hi}}$ on $[z_R, 1]$. Each shape is fit independently via NUTS, scored by WAIC, and combined into Akaike weights $w_S$ (Appendix~\ref{app:cell-model}). 

\textbf{Result: distance-dependent transfer holds for all three model families.}
Three embedding pooling schemes give strong $z$-dependence results for all families: \texttt{last\_token}, \texttt{spectral\_all}, and \texttt{mean\_pool}. We display the strongest, namely \texttt{spectral\_all}, in Figure~\ref{fig:dense47-headline}. Additional tables and plots are in Appendix~\ref{app:cross-metric} and ~\ref{app:cpl-parameters}.  This experiment confirms the dependence of the refusal rate on distance that motivates the introduction of Equation \ref{eq:eps_neighbour}.

\begin{figure}[!htbp]
\centering
\includegraphics[width=\textwidth]{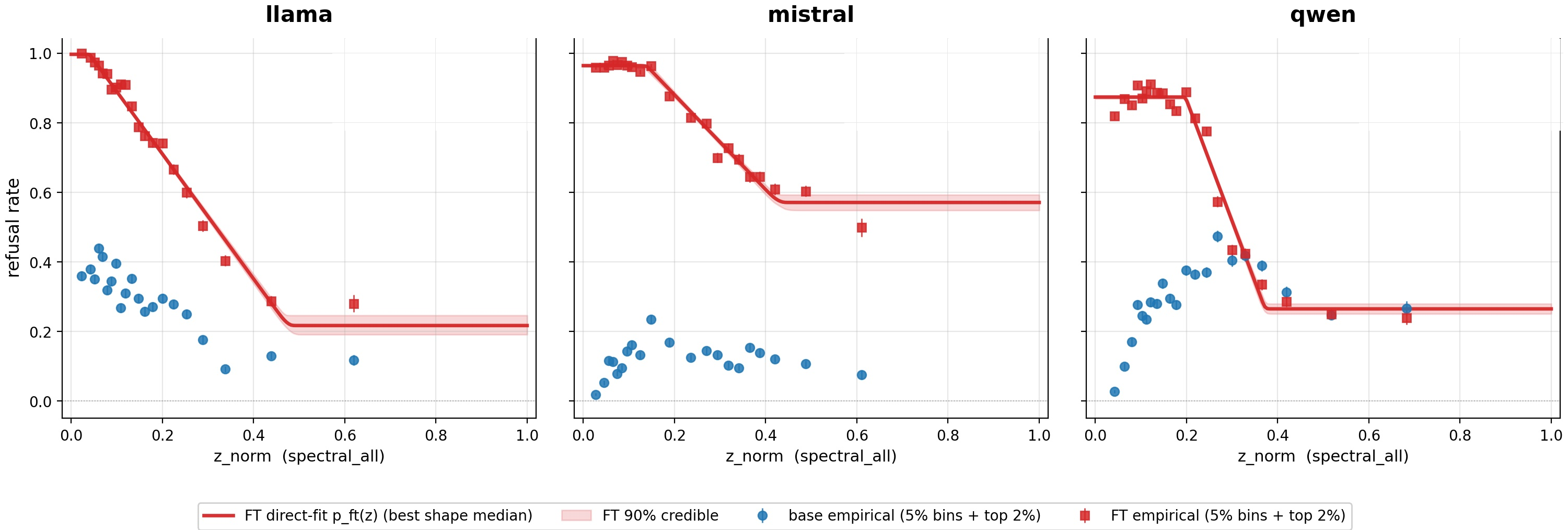}
\caption{Refusal rate $p_{\mathrm{ft}}(z)$ versus normalized embedding distance $z$ from the active training pool, per family. Orange: Constrained Piecewise-Linear posterior median curves for $p_{\mathrm{ft}}$. Blue: base refusal rates. Each dot represents a $5\%$-population bin, thus around $100-150$ prompts.}
\label{fig:dense47-headline}
\end{figure}

 The posterior probabilities for the Constrained Piecewise-Linear shape are very close to $1$ here, actually within the resolution of our model (<1e-6). Thus, this is strong evidence of {distance-dependent refusal rate}, for all three families. See Appendix~\ref{app:cross-metric}, Table~\ref{tab:dense47-stoch-cells} for more results.

\textbf{Capability is preserved at the canonical operating point.}
We measure capability and fluency drift through $200$-question MMLU and WikiText-2 perplexity respectively (Appendix~\ref{app:degeneration}). We also measure refusal on benign queries from WildJailbreak's \texttt{vanilla\_benign} subset. At our main rank-$1$ / exposure-$1600$ setting, we have $|\Delta\mathrm{MMLU}| \le 2$\,pp, the perplexity drift is below $5\%$ in absolute value, and refusal of benign queries is at most $1/30$. The results are displayed at the bottom of Figure~\ref{fig:dense47-headline}; the distance-dependent signal is therefore not confounded with the collapse of the base capability, nor with over-refusal. 

\textbf{Conclusion.}
Our experiments support the claim that distance-dependent refusal transfer \emph{holds for all three models} tested. Together with the controlled capability-preservation result, this supports the $\varepsilon$-coverage locality assumption (Equation~\ref{eq:eps_neighbour}) underlying the framework introduced in Section~\ref{defi_eval_games}, in the setting tested.

\subsection{Prompt transformations}
\label{sec:prompt-generation}

The Behavior Elicitation Tool (BET) framework~\cite{peignelefebvre2025bet} is a library of \emph{primitive} transformations of jailbreak queries. A primitive maps a prompt $p$ to a transformed prompt $T(p)$ keeping the prompt's semantic intent while perturbing its surface form (e.g.\ requesting a markdown-formatted answer, paraphrasing as a direct question, applying \texttt{rot13}). We denote $(T, T^{-1})$ a configured forward/inverse primitive pair. A subset of the primitives have known structural relationships: forward/inverse pairs, pairwise compositions, and self-inverse transformations. These structural relationships test whether the panel of primitives behaves as an algebraic structure on the model's embedding space.

\textbf{Protocol.} For each pair of model and primitive, we produce an embedding dataset of sources and targets $\{(x_i, y_i)\}_{i=1}^{n}$.
We then fit a hierarchy of candidate linear operators $\hat{\phi}_T(x) = \alpha_T x + \mathbf{b}_T \approx y$, scored by cross-validated $R^2$ (Appendix~\ref{app:operator-fits}) over the parameters $\alpha_T$ and $\mathbf{b}_T$. The fitted operators are not the final object of interest: their parameters are used as inputs to the structural tests below.

\textbf{Experiment 1: the $(\alpha, \mathbf{b})$ class is an adequate representation.} On the 13 single-primitive transformations, the two-parameter operator $\phi_T(x) = \alpha_T x + \mathbf{b}_T$ achieves wrapper-mean cross-validated $R^2 \in [0.81, 0.90]$ across the panel. Operators with more parameters overfit on our dataset. The $(\alpha_T, \mathbf{b}_T)$ class is both minimal --- translation alone leaves $0.19$--$0.28$ $R^2$ on the table --- and not under-specified (Appendix~\ref{app:operator-class-fits-table}, Table~\ref{tab:operator-class-fits}).

\textbf{Experiment 2: inverse and composition coherence both hold under the operator class's own algebra.} For a $(T, T^{-1})$, the fitted shifts should satisfy $\mathbf{b}_T \approx -\mathbf{b}_{T^{-1}}$, scored by $\cos(\mathbf{b}_T, -\mathbf{b}_{T^{-1}})$. For a composition $C = L \circ R$, the operator parameters should satisfy $\alpha_C \approx \alpha_L \alpha_R$ and $\mathbf{b}_C \approx \alpha_L \mathbf{b}_R + \mathbf{b}_L$. 
For the six $(T, T^{-1})$ pairs, the minimum inverse-coherence cosine $\cos(\mathbf{b}_T, -\mathbf{b}_{T^{-1}})$ is $0.887$, supporting the predicted $\mathbf{b}_T \approx -\mathbf{b}_{T^{-1}}$ relation at this early-layer pooled tag (Table~\ref{tab:inverse-coherence}). For a composition $C = L \circ R$, the $(\alpha, \mathbf{b})$-class composition law predicts $\alpha_C = \alpha_L \alpha_R$ and $\mathbf{b}_C = \alpha_L \mathbf{b}_R + \mathbf{b}_L$. The multiplicative scalar prediction agrees with the fitted $\alpha_C$ to within $0.3$--$6.4\%$ (Table~\ref{tab:composition-coherence}).

\textbf{Experiment 3: the self-inverse primitive is encoded as a partial involution.} For a self-inverse primitive, we fit a linear isometric operator $R_T$ to source/target embeddings. It should satisfy $R_T^2 \approx I$, scored by the trace ratio $\mathrm{tr}(R_T^2)/n$ on the rank-$n$ active subspace.
For \texttt{rot13\_full}, the orthogonal operator $R$ fitted to source/target embedding pairs is invertible by construction but not an exact involution: the active-subspace trace ratio $\mathrm{tr}(R^2)/n$ lies in $[0.49, 0.56]$ in $48/49$ (model, embedding-tag) cells (median $0.52$). The companion $\mathrm{tr}(R)/n$ is uniformly $\approx 0$. The fact $\mathrm{tr}(R^2)/n \approx 0.52$ means roughly half the $n$-dimensional active subspace acts as a true order-2 reflection and the other half as a non-trivial rotation (Appendix~\ref{app:procrustes-trace}). This $\sim$50/50 split is concentrated across the seven models we test.

\textbf{Conclusion.} These three experiments support that prompt transformations act as group elements on the model's embedding space, lending empirical grounding to the group-action in Section~\ref{ss:group_actions_introduction}.

\section{Limitations and broader impacts}\label{limitations}
\textbf{Limitations.} Our theoretical analysis is restricted to the simplest instance of the framework: cyclic translation groups on the one-dimensional circle; richer group structures (higher-dimensional tori, non-abelian groups, infinite-order generators) are not treated here. The empirical study on fine-tuning generalization uses three relatively small fine-tuned models, a single judge model (\textsc{OLMo-3.1-32B-Instruct}) and computes prompt-prompt distances in base-model embedding spaces; downstream conclusions may depend on all such choices. Generalization-range estimates are derived only in those specific setups, calling for more experiments. The empirical work on prompt transformations could also benefit from further investigation, for instance considering additional transformation techniques and possibly nonlinear fitted operators.

\textbf{Broader impacts.} A potential negative use of this framework is that a malicious trainer could leverage our theoretical analysis to better understand how to evade evaluation, optimizing a memorization patch strategy rather than genuinely improving the model. However, our theoretical results are derived in the worst case for the evaluator: they hold even when the trainer plays optimally, so the guarantees we provide to the evaluator remain valid as long as the underlying assumptions are satisfied.

\textbf{Generality.} Beyond LLM safety, the framework generalizes to any black-box model audit setting (e.g., classification~\cite{shahin2022washing} or ranking~\cite{mouton2024auditing}) where purely local behavior changes are possible, and to evaluation metrics beyond jailbreak rate, including fairness~\cite{lafargue2025exposing} and privacy~\cite{guerra2026understanding}.
More broadly, as AI governance frameworks such as the EU AI Act~\cite{AIAct} increasingly rely on evaluation protocols to certify model safety and fairness, our results demonstrate that such protocols must  explicitly model trainer-side adaptation.

\section{Conclusion}\label{conclusion}
We introduced the Evaluation Game, a game-theoretic framework that models the strategic interaction between a trainer performing robustness fine-tuning and an evaluator auditing for jailbreaks, with each player equipped with data augmentation methods modeled as group actions. On the circle, a closed-form analysis reveals a sharp phase structure governed by the arithmetic of the two groups; empirically, our experiments support both core modeling assumptions -- locality of fine-tuning-induced generalization across the three fine-tuned families, and group-action-like structure of prompt transformations across the seven-model operator panel.

Our results expose a fundamental limitation of static benchmarks: they cannot distinguish a genuine fix from a memorized patch. As governance frameworks increasingly rely on evaluation protocols to certify model safety, explicitly modeling trainer-side adaptation should be a core design principle -- for LLM safety and, more broadly, for any black-box audit setting. A natural direction for future work is the collaborative setting, in which multiple evaluators jointly audit the same model and must coordinate their queries to maximize coverage.

\bibliographystyle{abbrvnat}
\bibliography{bibliography.bib}

\appendix

\section{Theory Appendix}
\label{app_theory}\label{app_torus_proofs}

\subsection{Circle phase diagram --- proofs}\label{app_circle_phase_diagram}

This section proves the entries of Table~\ref{table:boom}. We use the
notation and assumptions stated in the main body of the article. For
$x \in \Rr$, write $\|x\| = \min_{m \in \Zz} |x - m|$ for the distance to the
nearest integer. Recall $g = \gcd(p, q)$, $s = p/g$, $L = pq/g$,
$\varepsilon^\star = 1/L$, $\varepsilon_{\mathrm{sat}} = \lfloor s/2\rfloor / L$,
and the \emph{$\varepsilon$-shift set}
\[
  S(\varepsilon) := \bigl\{\, m \in \Zz/p\Zz : f(m) < \varepsilon \,\bigr\},
  \qquad
  f(m) := \min_{j \in \{0, \ldots, q-1\}} \bigl\| \tfrac{j}{q} - \tfrac{m}{p} \bigr\|.
\]
We also use the reachable-center packing factor
\[
  Q(\varepsilon,p,q)
  :=
  \max_{z \in \Omega_E + H_{\mathrm{train}}}
  \bigl|\Omega_E \cap B(z,\varepsilon)\bigr|,
\]
where $B(z,\varepsilon)$ is the open ball for the circle metric
and
$\Omega_E + H_{\mathrm{train}}
= \{\,k/p+j/q \pmod 1 : 0\le k<p,\ 0\le j<q\,\}$.

The appendix is organized as follows:
\S\ref{app_circle_prelim} gathers preliminary lemmas about the shift set,
the critical distance, the structure of the trainer dataset under all
three move regimes, the common subgroup $H_g$ and its cosets, and the
closed forms of $\varepsilon^\star$ and $\varepsilon_{\mathrm{sat}}$.
\S\ref{app_circle_npaper} proves the cyclic-walk diagnostic master
theorem $N_{\mathrm{o}}^{\bullet} = s^\star$ for all three move regimes.
\S\ref{app_circle_norbit_binfty}--\S\ref{app_circle_norbit_b0} prove the
orbit-covering bounds under $\mathrm{full}$, $\mathrm{batch}$, and $\mathrm{single}$ respectively
for every row of Table~\ref{table:boom} (tight in the sub-critical
regime; the saturation-$\mathrm{batch}$ entry is an upper bound).
\S\ref{app_circle_three_tier} specializes to $p = q \ge 3$ with
$0 < \varepsilon \le \varepsilon^\star$ to exhibit the three-tier batching
ladder $\Theta(1) \ll \Theta(\log p) \ll \Theta(p)$.

\subsubsection{Preliminaries}\label{app_circle_prelim}

We begin with the verification of the example from Section~\ref{sec:examples}.

\begin{example}[Detailed verification for $p=5,\,q=3$]
\label{ex_app_circle_five_three}
Let $\varepsilon = 0.08 > 1/15$. Translation by $1/3$ approximates the shift
$+2$ on $\Zz/5\Zz$, and translation by $2/3$ approximates the shift $-2$.
The following play realizes the three-high-failure trace in
Figure~\ref{fig_circle_five_three_threshold}; it is not optimal for the
$N_{\mathrm{o}}$ diagnostic.
\begin{itemize}
  \item Round 0: E sends $0$, so $r_0 = 1$. T absorbs it.
  \item Round 1: E sends $1/5$, so $r_1 = 1$. T translates $\{0,\,1/5\}$ by
  $2/3$, producing $2/3$ and $13/15$, which $\varepsilon$-cover $3/5$ and
  $4/5$ respectively.
  \item Round 2: the only remaining uncovered E-orbit point is $2/5$.
  E sends $2/5$, so $r_2 = 1$. T absorbs it.
\end{itemize}
At the start of round 3 all five E-orbit points are covered, so $r_3 = 0$.
This play therefore has three high-failure rounds: it is the batch-augmentation
intermediate regime, not the fully independent doubling regime.
For first neutralisation under optimal play, however, T can choose the
$1/3$ shift after round 0, covering $2/5$ at the beginning of round 2; this
agrees with Theorem~\ref{thm_npaper_master}, which gives
$N_{\mathrm{o}} = 2$ in this parameter choice.
\end{example}

\begin{lemma}[Shift-set membership]\label{lem_shift_set_membership}
For every $m \in \Zz/p\Zz$,
\[
  f(m) \;=\; \frac{\|mq/p\|}{q} \;=\; \frac{|mq|_p}{p\,q},
\]
where $|t|_p := \min(t \bmod p,\,p - (t \bmod p))$.
\end{lemma}

\begin{proof}
Write $mq/p = a + r$ with $a \in \Zz$ and $r \in [-1/2,\,1/2]$. The closest
point of $\frac{1}{q}\Zz$ to $m/p$ is at distance $|r|/q = \|mq/p\|/q$,
which proves the first equality. The second is by definition of
$|\cdot|_p$.
\end{proof}

\begin{lemma}[Critical positive grid distance]\label{lem_critical_distance}
The minimum positive distance between an E-orbit point and a trainer translate
of an E-orbit point is
\[
  \varepsilon^\star
  \;=\;
  \min\left\{
  d\!\left(\frac{k}{p},\,\frac{k'}{p} + \frac{j}{q}\right)
  :
  \begin{array}{c}
  k,\,k' \in \{0,\ldots,p-1\},\ j \in \{0,\ldots,q-1\},\\
  d\!\left(\frac{k}{p},\,\frac{k'}{p} + \frac{j}{q}\right) > 0
  \end{array}
  \right\}
  \;=\;
  \frac{1}{L}
  \;=\;
  \frac{\gcd(p,q)}{pq}.
\]
\end{lemma}

\begin{proof}
The subgroup of $\mathbb{T}^1$ generated by $1/p$ and $1/q$ is the cyclic
grid
\[
  \left\langle \tfrac{1}{p},\,\tfrac{1}{q} \right\rangle
  \;=\;
  \left\langle \tfrac{1}{L} \right\rangle,
  \qquad L = \operatorname{lcm}(p, q).
\]
Indeed, $1/p$ and $1/q$ both lie in $\langle 1/L \rangle$ since $L$ is a
common multiple, and conversely Bezout's identity gives integers $u,\,v$
with $up + vq = g$, hence $u/q + v/p = g/(pq) = 1/L$.

Every difference of the form $k/p - k'/p - j/q$ therefore lies on the
$1/L$-grid, so each nonzero such distance is at least $1/L$. To see that
$1/L$ is realized, take the Bezout identity above: choosing
$k - k' \equiv v \pmod{p}$ and $j \equiv -u \pmod{q}$ produces a difference
equal to $1/L$ on the circle. This also covers the trivial-trainer case
$q=1$: then $j=0$, $L=p$, and the distance $1/p=1/L$ is realized by
consecutive E-orbit points. In every case, $\varepsilon^\star = 1/L$.
\end{proof}

\begin{lemma}[Trainer dataset form]\label{lem_dataset_form}
Let $\bullet \in \{\mathrm{single}, \mathrm{batch}, \mathrm{full}\}$. Under any $\bullet$ strategy and the
cyclic-walk evaluator, every $d \in D^{\mathrm{train}}_n$ has the form
$d = k/p + j/q \pmod{1}$ for some $k \in \{0, \ldots, n-1\}$ and
$j \in \Zz/q\Zz$. Equivalently,
\[
  D^{\mathrm{train}}_n
  \;\subseteq\;
  \bigl\{\, k/p + j/q \pmod 1 \,:\, 0 \le k \le n-1,\ j \in \Zz/q\Zz \,\bigr\}.
\]
\end{lemma}

\begin{proof}
Induction on $n$. The base case $D^{\mathrm{train}}_0 = \emptyset$ is trivial.

For the inductive step, each of the three update rules produces
$D^{\mathrm{train}}_{n+1}$ as a subset of
$H_{\mathrm{train}} \cdot (D^{\mathrm{train}}_n \cup E_n)$:
\begin{itemize}
  \item $\mathrm{single}$: $D^{\mathrm{train}}_{n+1}
        = D^{\mathrm{train}}_n \cup E_n \cup \{h_n \cdot c_n\}$
        with $c_n \in D^{\mathrm{train}}_n \cup E_n$ and
        $h_n \in H_{\mathrm{train}}$;
  \item $\mathrm{batch}$: $D^{\mathrm{train}}_{n+1}
        = D^{\mathrm{train}}_n \cup E_n \cup h_n \cdot C_n$ with
        $C_n \subseteq D^{\mathrm{train}}_n \cup E_n$;
  \item $\mathrm{full}$: $D^{\mathrm{train}}_{n+1}
        = H_{\mathrm{train}} \cdot (D^{\mathrm{train}}_n \cup E_n)$.
\end{itemize}
By the induction hypothesis, every element of $D^{\mathrm{train}}_n$ has the
form $k/p + j/q$ with $0 \le k \le n-1$; the new query $E_n = \{n/p\}$ has
the form $n/p + 0/q$ with $k = n$, so every element of
$D^{\mathrm{train}}_n \cup E_n$ has the form $k/p + j/q$ with
$0 \le k \le n$. Translating by any $j_0/q \in H_{\mathrm{train}}$ preserves
the form: it sends $k/p + j/q$ to $k/p + (j + j_0)/q$ with $k$ unchanged.
Hence every element of $D^{\mathrm{train}}_{n+1}$ has the form $k/p + j/q$
with $0 \le k \le n$, as required.
\end{proof}

\begin{lemma}[$\mathrm{single}$ dataset capacity]\label{lem_b0_capacity}
Under any $\mathrm{single}$ strategy and the cyclic-walk evaluator,
\[
  |D^{\mathrm{train}}_n| \;\le\; 2n,
  \qquad
  \text{hence}\quad
  |D^{\mathrm{train}}_n \cap \Omega_E| \;\le\; 2n.
\]
\end{lemma}

\begin{proof}
The $\mathrm{single}$ update rule
$D^{\mathrm{train}}_{n+1} = D^{\mathrm{train}}_n \cup E_n \cup \{h_n \cdot c_n\}$
adds at most two new points per round, so
$|D^{\mathrm{train}}_{n+1}| \le |D^{\mathrm{train}}_n| + 2$. Induction from
$|D^{\mathrm{train}}_0| = 0$ gives $|D^{\mathrm{train}}_n| \le 2n$.
\end{proof}

\paragraph{Common subgroup and cosets.}
We identify $\Omega_E$ with $\Zz/p\Zz$ via $k/p \mapsto k$. The intersection
\[
  H_g \;:=\; H_{\mathrm{train}} \cap G_{\mathrm{eval}}
       \;=\; \langle 1/g\rangle
\]
acts on $\Omega_E$ as the subgroup $s\Zz/p\Zz \le \Zz/p\Zz$, with orbits the
$s$ cosets
\[
  C_i \;=\; i + s\Zz/p\Zz \;=\; \{\, i,\, i+s,\, \ldots,\, i+(g-1)s \,\},
  \qquad i = 0, \ldots, s-1.
\]
For $g = 1$, $H_g = \{0\}$ and each coset $C_i = \{i\}$ is a single orbit
point.

\begin{lemma}[$H_g$ acts regularly on each coset]\label{lem_h_g_regular}
For each $i \in \{0, \ldots, s-1\}$, the action of $H_g$ on $C_i$ is regular:
for any $c, x \in C_i$ there is a unique $h \in H_g$ with $h \cdot c = x$.
\end{lemma}

\begin{proof}
$|H_g| = |C_i| = g$ and the action is free: if $h \cdot c = c$ then
$h \in \mathrm{Stab}(c) = \{0\}$, since $H_g$ acts on $\Zz/p\Zz$ by
translation by multiples of $s$ and only $0$ fixes a point. A free action of
a group on a set of equal cardinality is regular.
\end{proof}

\begin{lemma}[Sub-critical orbit-trace]\label{lem_subcrit_coset_trace}
Assume $\varepsilon \le \varepsilon^\star = 1/L$. Let $\bullet \in \{\mathrm{single}, \mathrm{batch}, \mathrm{full}\}$
and let $D^{\mathrm{train}}_n$ be the dataset under any $\bullet$ strategy
with the cyclic-walk evaluator. If $d \in D^{\mathrm{train}}_n$
$\varepsilon$-covers an orbit point $\ell/p \in \Omega_E$, then $d = \ell/p$;
writing $d = k/p + j/q$ as in Lemma~\ref{lem_dataset_form}, we further have
$j/q \in H_g$ and $\ell \equiv k \pmod s$. In particular, every covered orbit
point produced from absorbed seed $k/p$ lies in the same $H_g$-coset
$C_{k \bmod s}$, and off-orbit trainer points cover no point of $\Omega_E$.
\end{lemma}

\begin{proof}
By Lemma~\ref{lem_dataset_form}, $d = k/p + j/q$ with $0 \le k \le n-1$ and
$j \in \Zz/q\Zz$. The covering condition is
$\|\ell/p - k/p - j/q\| < \varepsilon \le \varepsilon^\star$. By
Lemma~\ref{lem_critical_distance}, the only such difference strictly below
$\varepsilon^\star$ is zero on the circle, so $\ell/p = k/p + j/q$. Hence
$j/q = (\ell - k)/p \in H_{\mathrm{train}} \cap G_{\mathrm{eval}} = H_g$,
and $\ell - k \in s\Zz$. The off-orbit assertion is the contrapositive of
the first line: if $d \notin \Omega_E$, then $d$ cannot $\varepsilon$-cover
any orbit point.
\end{proof}

\begin{lemma}[Coset-seeding lower bound, sub-critical]\label{lem_coset_seed_lb}
Assume $\varepsilon \le \varepsilon^\star$ and $g \ge 1$. For every
$\bullet \in \{\mathrm{single}, \mathrm{batch}, \mathrm{full}\}$,
\[
  N_{\mathrm{orbit}}^{\bullet}(\varepsilon, p, q) \;\ge\; s \;=\; p/g.
\]
\end{lemma}

\begin{proof}
By Lemma~\ref{lem_subcrit_coset_trace}, an orbit point of coset $C_i$ can be
covered only by a trainer datum whose absorbed seed $k/p$ has
$k \equiv i \pmod s$. The cyclic walk supplies $E_n = \{n/p\}$, so coset
$C_i$ has no in-coset seed before round $n = i$. Coset $C_{s-1}$ thus first
has a seed at round $s-1$, hence full orbit coverage requires $n \ge s$.
\end{proof}

\begin{lemma}[Sub-critical shift set]\label{lem_eps_star_S}
For $\varepsilon \le \varepsilon^\star = 1/L$,
\[
  S(\varepsilon) \;=\; \{\, m \in \Zz/p\Zz : mq \equiv 0 \pmod p\,\}
  \;=\; (p/g)\,\Zz/p\Zz,
\]
a subgroup of order $g$ with elements equally spaced by $s = p/g$ on
$\Zz/p\Zz$.
\end{lemma}

\begin{proof}
By Lemma~\ref{lem_shift_set_membership}, $f(m) = |mq|_p / (pq)$. Since
$g \mid q$, $mq \in g\Zz$, hence $|mq|_p \in g\Zz$, so $|mq|_p \ge g$
whenever $mq \not\equiv 0 \pmod p$. Therefore
\[
  f(m) \;\in\; \{0\} \cup [g/(pq),\, 1/(2q)]
  \;=\; \{0\} \cup [1/L,\, 1/(2q)],
\]
and $f(m) < \varepsilon \le 1/L$ forces $f(m) = 0$, i.e.
$mq \equiv 0 \pmod p$. The latter is equivalent to $m \in (p/g)\Zz/p\Zz$
since $(p/g)\Zz/p\Zz$ is the kernel of multiplication by $q$ on $\Zz/p\Zz$.
\end{proof}

\begin{lemma}[Saturation threshold and shift set]\label{lem_eps_sat_S}
The saturation threshold satisfies
\[
  \varepsilon_{\mathrm{sat}}
  \;:=\; \max_{m \in \Zz/p\Zz} f(m)
  \;=\; \frac{\lfloor s/2 \rfloor \cdot g}{pq}
  \;=\; \frac{\lfloor s/2 \rfloor}{L}.
\]
For $\varepsilon > \varepsilon_{\mathrm{sat}}$, $S(\varepsilon) = \Zz/p\Zz$;
equivalently, $\Omega_E \subseteq V_\varepsilon(H_{\mathrm{train}})$.
\end{lemma}

\begin{proof}
By Lemma~\ref{lem_shift_set_membership}, $f(m) = |mq|_p/(pq)$. Since
$g \mid q$, the value $mq \bmod p$ lies in $g\Zz/p\Zz$ and ranges over
$\{0, g, 2g, \ldots, (s-1)g\}$ as $m$ varies (here $s = p/g$). Hence
$|mq|_p \in g \cdot \{0, 1, \ldots, \lfloor s/2\rfloor\}$, with maximum
$g \cdot \lfloor s/2\rfloor$. Therefore
$\varepsilon_{\mathrm{sat}} = \max_m f(m) = g \lfloor s/2\rfloor / (pq) =
\lfloor s/2\rfloor / L$. The implication $\varepsilon > \varepsilon_{\mathrm{sat}}
\Rightarrow S(\varepsilon) = \Zz/p\Zz$ is by definition of the maximum. The
equivalence $S(\varepsilon) = \Zz/p\Zz \iff \Omega_E \subseteq
V_\varepsilon(H_{\mathrm{train}})$ unwinds the definition of $f$:
$f(m) < \varepsilon$ for every $m$ iff every orbit point $m/p$ is within
$\varepsilon$ of some $j/q \in H_{\mathrm{train}}$.
\end{proof}

\subsubsection{The cyclic-walk diagnostic}\label{app_circle_npaper}

\begin{theorem}[$N_{\mathrm{o}}$ master theorem]\label{thm_npaper_master}
Let $\bullet \in \{\mathrm{single}, \mathrm{batch}, \mathrm{full}\}$. For all $p \ge 2$, $q \ge 1$, and
$\varepsilon > 0$, under the cyclic-walk evaluator,
\[
  N_{\mathrm{o}}^{\bullet}(\varepsilon, p, q)
  \;=\;
  s^\star(\varepsilon, p, q)
  \;:=\;
  \min\bigl(\,\{\, m \in \{1, \ldots, p-1\} : f(m) < \varepsilon\,\} \cup \{p\}\,\bigr).
\]
(The convention $\cup\{p\}$ ensures $s^\star = p$ when no $m \in \{1, \ldots, p-1\}$ satisfies $f(m) < \varepsilon$.)
\end{theorem}

\begin{proof}
We prove matching upper and lower bounds.

\smallskip
\noindent\textbf{Upper bound: $N_{\mathrm{o}}^{\mathrm{single}} \le s^\star$.}\quad
If $s^\star = p$, then after the first $p$ rounds the trainer has absorbed
every point of the cyclic E-orbit, so the query at round $p$ is covered.
Assume $s^\star < p$ and pick $j^\star \in \{0, \ldots, q-1\}$ achieving
$\|j^\star/q - s^\star/p\| = f(s^\star) < \varepsilon$. Define a $\mathrm{single}$
strategy: in round $0$, absorb $E_0 = \{0\}$ and choose $h_0 = j^\star/q$,
$c_0 = 0$, yielding $D^{\mathrm{train}}_1 = \{0,\, j^\star/q\}$. In rounds
$1 \le n < s^\star$, absorb $E_n$ and choose $h_n = 0$ (no-op), so
$j^\star/q \in D^{\mathrm{train}}_n$ for all $1 \le n \le s^\star$. At round
$s^\star$, the query $E_{s^\star} = \{s^\star/p\}$ satisfies
$\|s^\star/p - j^\star/q\| = f(s^\star) < \varepsilon$, hence $r_{s^\star} = 0$
and $N_{\mathrm{o}}^{\mathrm{single}} \le s^\star$.

Any $\mathrm{single}$ strategy embeds as a degenerate $\mathrm{batch}$ or $\mathrm{full}$ strategy: in
each round, augment with the same chosen element $h_n \in H_{\mathrm{train}}$
(under $\mathrm{batch}$ with one shift) or with all of $H_{\mathrm{train}}$
(under $\mathrm{full}$). Both yield a dataset $D^{\mathrm{train}}_n$ containing the
$\mathrm{single}$ dataset, hence with no smaller $V_\varepsilon$-cover. Therefore the
same upper bound holds for $\mathrm{batch}$ and $\mathrm{full}$.

\smallskip
\noindent\textbf{Lower bound: $N_{\mathrm{o}}^{\bullet} \ge s^\star$ for
$\bullet \in \{\mathrm{single}, \mathrm{batch}, \mathrm{full}\}$.}\quad
For $r_n = 0$ we need $n/p \in V_\varepsilon(D^{\mathrm{train}}_n)$, i.e.,
some $d \in D^{\mathrm{train}}_n$ with $\|n/p - d\| < \varepsilon$. By
Lemma~\ref{lem_dataset_form}, $d = k/p + j/q$ with $0 \le k \le n-1$ and
$j \in \Zz/q\Zz$. Setting $m := n - k$, we obtain $m \in \{1, \ldots, n\}$,
and the covering condition reads $\|m/p - j/q\| < \varepsilon$, which by
Lemma~\ref{lem_shift_set_membership} is exactly $m \in S(\varepsilon)$. The
smallest such $m \in \{1, \ldots, p-1\}$ (or $p$ if none exists) is $s^\star$,
hence $n \ge s^\star$ as soon as $r_n = 0$.

\smallskip
Combining both bounds: $N_{\mathrm{o}}^{\bullet} = s^\star$ for $\bullet \in \{\mathrm{single}, \mathrm{batch}, \mathrm{full}\}$.
\end{proof}

\begin{remark}\label{rem_npaper_collapse}
The collapse $N_{\mathrm{o}}^{\mathrm{single}} = N_{\mathrm{o}}^{\mathrm{batch}} = N_{\mathrm{o}}^{\mathrm{full}}$
reflects that the cyclic-walk evaluator only requires \emph{one} point to be
$\varepsilon$-covered per round, and the trainer's full schedule is offline-
predictable. The minimum-budget move $\mathrm{single}$ already places a single
well-positioned anchor in round $0$ that suffices through round $s^\star$.
The diagnostic is therefore insensitive to the trainer's per-round batching
budget; sensitivity to that axis appears under $N_{\mathrm{orbit}}$.
\end{remark}

\subsubsection{Orbit covering under full}\label{app_circle_norbit_binfty}

\begin{lemma}[$\mathrm{full}$ dataset]\label{lem_binfty_dataset}
Under $\mathrm{full}$ and the cyclic-walk evaluator, for $1 \le n \le p$,
\[
  D^{\mathrm{train}}_n \;=\; \bigl\{\, k/p + j/q \pmod 1 \,:\, 0 \le k \le n-1,\ 0 \le j < q\,\bigr\}.
\]
\end{lemma}

\begin{proof}
$D^{\mathrm{train}}_1 = H_{\mathrm{train}} \cdot \{0\} = \{j/q\}_{j=0}^{q-1}$.
Inductively, $D^{\mathrm{train}}_{n+1} = H_{\mathrm{train}} \cdot (D^{\mathrm{train}}_n \cup \{n/p\})
= D^{\mathrm{train}}_n \cup H_{\mathrm{train}} \cdot \{n/p\}$ (using $H_{\mathrm{train}}$-invariance of
$D^{\mathrm{train}}_n$), which is the displayed union.
\end{proof}

\begin{theorem}[$\mathrm{full}$ orbit covering time]\label{thm_norbit_binfty}
For all $p \ge 2$, $q \ge 1$, and $\varepsilon > 0$,
\[
  N_{\mathrm{orbit}}^{\mathrm{full}}(\varepsilon, p, q)
  \;=\;
  \Gamma\bigl(S(\varepsilon)\bigr),
\]
where $\Gamma(S)$ is the maximum cyclic gap of $S$ on $\Zz/p\Zz$:
sorting $S = \{s_0 < s_1 < \cdots < s_{|S|-1}\}$ in $\{0, \ldots, p-1\}$ and
setting $s_{|S|} := s_0 + p$,
\[
  \Gamma(S) := \max_{0 \le i < |S|}\,(s_{i+1} - s_i).
\]
\end{theorem}

\begin{proof}
By Lemma~\ref{lem_binfty_dataset}, the round-$n$ trainer data is
$D^{\mathrm{train}}_n = K_n/p + (1/q)\Zz/q\Zz$, where
$K_n := \{0, 1, \ldots, n-1\} \subseteq \Zz/p\Zz$. Hence for
$\ell \in \Zz/p\Zz$,
\[
  \ell/p \in V_\varepsilon\!\bigl(D^{\mathrm{train}}_n\bigr)
  \iff
  \exists\,k \in K_n,\ j \in \{0, \ldots, q-1\}:\
  \|\ell/p - k/p - j/q\| < \varepsilon,
\]
which by Lemma~\ref{lem_shift_set_membership} is equivalent to
$\ell - k \in S(\varepsilon) \pmod p$ for some $k \in K_n$, i.e.,
$\ell \in K_n + S(\varepsilon) \pmod p$. Therefore
\[
  N_{\mathrm{orbit}}^{\mathrm{full}}
  \;=\; \min\bigl\{\,n \ge 1 : K_n + S(\varepsilon) = \Zz/p\Zz\,\bigr\}.
\]

Fix $\ell \in \Zz/p\Zz$. The condition
$\ell \in K_n + S(\varepsilon)$ unfolds as
\[
  \exists\, k \in \{0,\ldots,n-1\},\ s \in S(\varepsilon):\ \ell - s \equiv k \pmod p
  \iff
  \exists\, s \in S(\varepsilon):\ (\ell - s) \bmod p \;\le\; n-1.
\]
Thus the smallest $n$ for which $\ell$ is covered equals
\[
  n^\star(\ell) \;:=\; 1 + \min_{s \in S(\varepsilon)}\,\bigl((\ell - s) \bmod p\bigr).
\]
Geometrically, $(\ell - s) \bmod p$ is the cyclic clockwise distance from $s$ to
$\ell$ on $\Zz/p\Zz$, so $n^\star(\ell) - 1$ is the cyclic
distance from $\ell$ \emph{back} to the nearest preceding element of
$S(\varepsilon)$. Taking the maximum over $\ell$,
\[
  N_{\mathrm{orbit}}^{\mathrm{full}}
  \;=\; \max_{\ell \in \Zz/p\Zz} n^\star(\ell)
  \;=\; 1 + \max_{\ell}\,\min_{s \in S(\varepsilon)}\,\bigl((\ell - s) \bmod p\bigr).
\]
The inner max-min is achieved when $\ell$ sits at the far end of the largest
cyclic gap of $S(\varepsilon)$: with $S = \{s_0 < \cdots < s_{|S|-1}\}$ and
$s_{|S|} := s_0 + p$, choosing $\ell = s_{i+1} - 1$ (the predecessor of $s_{i+1}$)
gives $\min_s (\ell - s) \bmod p = s_{i+1} - 1 - s_i = g_i - 1$, where
$g_i := s_{i+1} - s_i$. Hence the maximum equals $\Gamma(S(\varepsilon)) - 1$,
and $N_{\mathrm{orbit}}^{\mathrm{full}} = \Gamma(S(\varepsilon))$.
\end{proof}

\begin{coro}[Sub-critical, coprime: $\mathrm{full}$ and $N_{\mathrm{o}}$]\label{cor_binfty_coprime}
For $\varepsilon \le \varepsilon^\star$ and $g = 1$,
$S(\varepsilon) = \{0\}$, hence $\Gamma(S(\varepsilon)) = p$,
$s^\star = p$, and
$N_{\mathrm{orbit}}^{\mathrm{full}} = N_{\mathrm{o}}^{\bullet} = p$ for
$\bullet \in \{\mathrm{single}, \mathrm{batch}, \mathrm{full}\}$.
\end{coro}

\begin{coro}[Sub-critical, $g \ge 2$: $\mathrm{full}$ and $N_{\mathrm{o}}$]\label{cor_binfty_subcrit}
For $\varepsilon \le \varepsilon^\star$ and $g \ge 2$,
$S(\varepsilon) = (p/g)\Zz/p\Zz$ has $g$ equally-spaced elements, hence
$\Gamma(S(\varepsilon)) = p/g$, $s^\star = p/g$, and
$N_{\mathrm{orbit}}^{\mathrm{full}} = N_{\mathrm{o}}^{\bullet} = p/g$ for
$\bullet \in \{\mathrm{single}, \mathrm{batch}, \mathrm{full}\}$.
\end{coro}

\begin{coro}[Saturation: $\mathrm{full}$ and $N_{\mathrm{o}}$]\label{cor_binfty_sat}
For $\varepsilon > \varepsilon_{\mathrm{sat}}$,
$S(\varepsilon) = \Zz/p\Zz$, hence $\Gamma(S(\varepsilon)) = 1$,
$s^\star = 1$, and
$N_{\mathrm{orbit}}^{\mathrm{full}} = N_{\mathrm{o}}^{\bullet} = 1$ for
$\bullet \in \{\mathrm{single}, \mathrm{batch}, \mathrm{full}\}$. The conclusion covers the large-radius
saturation cases $\varepsilon > 1/2$ (for example $q = 1$ with even $p$,
where $\varepsilon_{\mathrm{sat}} = 1/2$).
\end{coro}

\begin{proof}[Proof of Corollaries~\ref{cor_binfty_coprime}--\ref{cor_binfty_sat}]
Combine Theorem~\ref{thm_norbit_binfty}
($N_{\mathrm{orbit}}^{\mathrm{full}} = \Gamma(S(\varepsilon))$) and
Theorem~\ref{thm_npaper_master} ($N_{\mathrm{o}}^{\bullet} = s^\star$,
the smallest positive element of $S(\varepsilon)$, with the convention
$s^\star = p$ if $S(\varepsilon) \cap \{1, \ldots, p-1\} = \emptyset$) with
Lemma~\ref{lem_eps_star_S} for the sub-critical case (which specializes to
$\{0\}$ at $g = 1$, giving $\Gamma = p$ and $s^\star = p$, and to
$(p/g)\Zz/p\Zz$ at $g \ge 2$, giving $\Gamma = p/g$ and $s^\star = p/g$),
and Lemma~\ref{lem_eps_sat_S} for the saturation case. The lemmas impose no
upper bound on $\varepsilon$, hence cover all large-radius cases including
$\varepsilon > 1/2$.
\end{proof}

\subsubsection{Orbit covering under batch}\label{app_circle_norbit_b1}

We continue to identify $\Omega_E$ with $\Zz/p\Zz$ via $k/p \mapsto k$, with
cosets $C_i = i + s\Zz/p\Zz$ of $H_g$.

\begin{lemma}[Dyadic subset-sum cover]\label{lem_dyadic_batch_cover}
Let $K \ge 1$, let $H = \langle 1/m\rangle \le \mathbb{T}^1$ be a cyclic
subgroup of order $m$ contained in $H_{\mathrm{train}}$, and write
$S_0 := D^{\mathrm{train}}_{n_0} \cup E_{n_0}$. Suppose in $K$ consecutive
rounds $n_0, n_0{+}1, \ldots, n_0{+}K{-}1$ the trainer plays $\mathrm{batch}$ with
$C_{n_0+k} = D^{\mathrm{train}}_{n_0+k} \cup E_{n_0+k}$ and
$h_{n_0+k} = 2^k/m$. Then
\[
  D^{\mathrm{train}}_{n_0 + K}
  \;\supseteq\;
  S_0
   \,+\, \Bigl\{\,\tfrac{1}{m}\!\!\sum_{k \in S} 2^k \pmod 1
              \;:\; S \subseteq \{0, \ldots, K-1\}\,\Bigr\}.
\]
If $K \ge \lceil \log_2 m\rceil$, the right-hand subset-sum set equals $H$,
so $D^{\mathrm{train}}_{n_0+K}$ contains every $H$-translate of every point
of $S_0$.
\end{lemma}

\begin{proof}
Let $T_k$ denote the set of dyadic subset sums
$\{(1/m) \sum_{j \in S} 2^j \bmod 1 : S \subseteq \{0, \ldots, k-1\}\}$, so
$T_0 = \{0\}$ and $T_{k+1} = T_k \cup (T_k + 2^k/m)$. We show by induction
on $k \ge 1$ that $S_0 + T_k \subseteq D^{\mathrm{train}}_{n_0+k}$.

Base case ($k = 1$). The round-$n_0$ batch update absorbs $E_{n_0}$
unshifted and additionally applies $h_{n_0} = 1/m$ to $C_{n_0} \supseteq
S_0$, so
\[
  D^{\mathrm{train}}_{n_0+1}
  \;\supseteq\; D^{\mathrm{train}}_{n_0} \cup E_{n_0} \cup h_{n_0} \cdot C_{n_0}
  \;\supseteq\; S_0 \cup (S_0 + 1/m)
  \;=\; S_0 + T_1.
\]

Inductive step ($k \to k+1$, $k \ge 1$). Assume $S_0 + T_k \subseteq
D^{\mathrm{train}}_{n_0+k}$. Since $D^{\mathrm{train}}_{n_0+k} \subseteq
C_{n_0+k}$, the round-$(n_0+k)$ update gives $D^{\mathrm{train}}_{n_0+k+1}
\supseteq h_{n_0+k} \cdot C_{n_0+k} \supseteq h_{n_0+k} \cdot (S_0 + T_k) =
S_0 + T_k + 2^k/m$. Combined with $D^{\mathrm{train}}_{n_0+k+1} \supseteq
D^{\mathrm{train}}_{n_0+k} \supseteq S_0 + T_k$, this yields
$D^{\mathrm{train}}_{n_0+k+1} \supseteq S_0 + (T_k \cup (T_k + 2^k/m)) =
S_0 + T_{k+1}$.

After $K \ge 1$ rounds, $S_0 + T_K \subseteq D^{\mathrm{train}}_{n_0+K}$. If
$K \ge \lceil \log_2 m\rceil$, the integer subset sums $\sum_{k \in S} 2^k$
exhaust $\{0, 1, \ldots, 2^K - 1\}$, and $2^K \ge m$, so their reductions
modulo $m$ cover all of $\{0, 1, \ldots, m - 1\}$. Hence $T_K = H$.
\end{proof}

\begin{theorem}[$\mathrm{batch}$ sub-critical bound, $g \ge 2$]\label{thm_norbit_b1_sub}
Assume $g \ge 2$ and let $\varepsilon \le \varepsilon^\star = g/(pq) = 1/L$.
Then
\[
  N_{\mathrm{orbit}}^{\mathrm{batch}}(\varepsilon, p, q)
  \;=\;
  \Theta\bigl(s + \log g\bigr)
  \;=\;
  \Theta(p/g + \log g).
\]
\end{theorem}

\begin{proof}
By Lemma~\ref{lem_subcrit_coset_trace}, every covered orbit point stays in
the $H_g$-coset of its absorbed seed, and off-orbit trainer points cover
nothing in $\Omega_E$. Combining with $\varepsilon \le \varepsilon^\star$,
two distinct orbit points are not $\varepsilon$-close, so orbit covering is
equivalent to
\(
  \Omega_E \cap \Omega_E\text{-covered seeds} = \Omega_E.
\)

\medskip
\noindent\textbf{Lower bound.}
Lemma~\ref{lem_coset_seed_lb} gives $N_{\mathrm{orbit}}^{\mathrm{batch}} \ge s$.
For the second bound, fix any coset $C_i$ and define
\[
  R_i \;:=\; \{\, k/p + j/q \pmod 1 \,:\, k \equiv i \pmod s,\ j \in \Zz/q\Zz\,\}
       \;=\; C_i + H_{\mathrm{train}} ,
\]
the set of points in $\Omega_E + H_{\mathrm{train}}$ whose $p$-component
lies in $C_i$. Equivalently $R_i = i/p + H_{\mathrm{train}}$ is an
$H_{\mathrm{train}}$-coset of size $q$, so $R_i$ is setwise preserved by
every $h \in H_{\mathrm{train}}$. Set $a_n^{(i)} := |D^{\mathrm{train}}_n \cap R_i|$.
Intersecting the batch update with $R_i$,
\[
  a_{n+1}^{(i)}
  \;\le\; a_n^{(i)} + |E_n \cap R_i| + |h_n \cdot (C_n \cap R_i)|
  \;\le\; a_n^{(i)} + 1 + (a_n^{(i)} + 1)
  \;=\; 2 a_n^{(i)} + 2 ,
\]
using $C_n \subseteq D^{\mathrm{train}}_n \cup E_n$ and the
$h_n$-invariance of $R_i$. Induction from $a_0^{(i)} = 0$ yields
$a_n^{(i)} \le 2^{n+1} - 2$. Covering $C_i$ requires
$|D^{\mathrm{train}}_n \cap C_i| \ge g$, and since
$D^{\mathrm{train}}_n \cap C_i \subseteq D^{\mathrm{train}}_n \cap R_i$,
we have $a_n^{(i)} \ge g$, hence $N_{\mathrm{orbit}}^{\mathrm{batch}} \ge
\log_2(g+2) - 1$.
Combining,
\[
  N_{\mathrm{orbit}}^{\mathrm{batch}}
  \;\ge\; \max\bigl(s,\, \log_2(g+2) - 1\bigr)
  \;\ge\; \tfrac14 (s + \log_2 g),
\]
using $g \ge 2$ and $s \ge 1$.

\medskip
\noindent\textbf{Upper bound.}
Set $K := \lceil \log_2 g\rceil$. During rounds $n = 0, \ldots, s-1$, take
$C_n = D^{\mathrm{train}}_n \cup E_n$ and $h_n = 0$, so the trainer absorbs
the evaluator queries. Then $D^{\mathrm{train}}_s \supseteq \{0, 1, \ldots,
s-1\}$, one seed in each $H_g$-coset $C_i$. For rounds $s, \ldots, s+K-1$,
take $C_n = D^{\mathrm{train}}_n \cup E_n$ and $h_{s+k} = 2^k/g$, and apply
Lemma~\ref{lem_dyadic_batch_cover} with $H = H_g$ (of order $g$, contained
in $H_{\mathrm{train}}$ since $g \mid q$): with seed set $S_0 :=
D^{\mathrm{train}}_s \cup E_s \supseteq \{0, 1, \ldots, s\}$, each of the
$s$ seeds $0, 1, \ldots, s-1$ in $S_0$ acquires every $H_g$-translate, so
$D^{\mathrm{train}}_{s+K} \supseteq \bigsqcup_{i=0}^{s-1} C_i = \Omega_E$.
Hence
\[
  N_{\mathrm{orbit}}^{\mathrm{batch}} \;\le\; s + \lceil \log_2 g\rceil.
\]

Combining the two bounds gives
$N_{\mathrm{orbit}}^{\mathrm{batch}} = \Theta(s + \log g) = \Theta(p/g + \log g)$.
\end{proof}

\begin{coro}[Explicit upper-bound expression]\label{cor_v_star_upper}
Under the hypotheses of Theorem~\ref{thm_norbit_b1_sub},
\[
  N_{\mathrm{orbit}}^{\mathrm{batch}}(\varepsilon, p, q)
  \;\le\; s + \lceil \log_2 g \rceil
  \qquad (\varepsilon \le \varepsilon^\star,\ g \ge 2),
\]
which already saturates the asymptotic claim of Theorem~\ref{thm_norbit_b1_sub}.
This is deliberately not an exact formula: the dyadic construction waits until
every coset has a seed and can be non-optimal in collapsed cases such as
$p = q$. The $\Theta(s + \log g)$ statement is the level needed for
Table~\ref{table:boom}.
\end{coro}

\begin{coro}[Sub-critical coprime, $\mathrm{batch}$]\label{cor_norbit_b1_coprime}
Assume $g = 1$ (equivalently $\gcd(p, q) = 1$) and $\varepsilon \le
\varepsilon^\star = 1/(pq)$. Then
$N_{\mathrm{orbit}}^{\mathrm{batch}}(\varepsilon, p, q) = p$.
\end{coro}

\begin{proof}
By Lemma~\ref{lem_eps_star_S} at $g = 1$, $S(\varepsilon) = \{0\}$ and the
$H_g$-cosets are the singletons $\{i\}$; so by
Lemma~\ref{lem_subcrit_coset_trace} every covered orbit point coincides
with an absorbed seed. The cyclic walk forces $p$ rounds before all $p$
seeds have arrived, hence $N_{\mathrm{orbit}}^{\mathrm{batch}} \ge p$. The upper
bound $p$ is achieved by absorption alone.
\end{proof}

\medskip

\begin{theorem}[$\mathrm{batch}$ saturation upper bound]\label{thm_norbit_b1_sat}
Assume $q \le p$. For $\varepsilon > \varepsilon_{\mathrm{sat}}$,
\[
  N_{\mathrm{orbit}}^{\mathrm{batch}}(\varepsilon, p, q)
  \;\le\; \max\{1, \lceil \log_2 q\rceil\}
  \;=\; O(\log p).
\]
\end{theorem}

\begin{proof}
If $q = 1$, then $H_{\mathrm{train}} = \{0\}$ and Lemma~\ref{lem_eps_sat_S}
gives $\Omega_E \subseteq V_\varepsilon(\{0\})$. After round $0$ the
trainer has absorbed $0$, so $N_{\mathrm{orbit}}^{\mathrm{batch}} \le 1$.

Assume $q \ge 2$, set $K := \lceil \log_2 q\rceil$, and apply
Lemma~\ref{lem_dyadic_batch_cover} with $H = H_{\mathrm{train}}$ (order $q$)
in rounds $0, 1, \ldots, K-1$ starting from $D^{\mathrm{train}}_0 \cup E_0 =
\{0\}$. The conclusion gives $D^{\mathrm{train}}_K \supseteq H_{\mathrm{train}}$.
By Lemma~\ref{lem_eps_sat_S}, $\Omega_E \subseteq
V_\varepsilon(H_{\mathrm{train}}) \subseteq V_\varepsilon(D^{\mathrm{train}}_K)$,
hence $N_{\mathrm{orbit}}^{\mathrm{batch}} \le K \le \lceil \log_2 p\rceil$.
\end{proof}

\begin{remark}[Why no error compounding]\label{rem_no_error_compounding}
A naive triangle-inequality bookkeeping of approximate H-translations in the
saturation regime would suggest an error blow-up of size $K\varepsilon$ after
$K$ rounds, restricting the doubling argument to a slack window
$\varepsilon - \varepsilon_{\mathrm{sat}}$. The proof above sidesteps this
entirely by exploiting that $H_{\mathrm{train}}$ is a \emph{group}: any chain
of H-shifts applied to a fixed seed $E_0$ collapses to a single H-translate
of $E_0$, so the dataset point covering an orbit point $\ell/p$ has the form
$0 + h$ with $h \in H_{\mathrm{train}}$, and only the single inequality
$\|\ell/p - h\| < \varepsilon$ (i.e., $\ell \in S(\varepsilon)$) is needed.
\end{remark}

\subsubsection{Orbit covering under single}\label{app_circle_norbit_b0}

\begin{coro}[Sub-critical coprime, $\mathrm{single}$]\label{cor_norbit_b0_coprime}
Assume $g = 1$ and $\varepsilon \le
\varepsilon^\star = 1/(pq)$. Then
$N_{\mathrm{orbit}}^{\mathrm{single}}(\varepsilon, p, q) = p$.
\end{coro}

\begin{proof}
At $g = 1$, $H_g = \{0\}$, so by Lemma~\ref{lem_subcrit_coset_trace} no
augmentation produces a new E-orbit point and absorption alone gives
$|D^{\mathrm{train}}_n \cap \Omega_E| = n$. Coverage requires $n \ge p$. To conclude, note that
$n = p$ suffices.
\end{proof}

\begin{theorem}[$\mathrm{single}$ sub-critical formula, $g \ge 2$]\label{thm_norbit_b0_sub}
Assume $g \ge 2$ and $\varepsilon \le \varepsilon^\star = g/(pq)$. Then
\[
  N_{\mathrm{orbit}}^{\mathrm{single}}(\varepsilon, p, q)
  \;=\;
  \lceil p/2\rceil.
\]
\end{theorem}

\begin{proof}
\noindent\textbf{Lower bound.} 
   \emph{Capacity.} By Lemma~\ref{lem_b0_capacity},
    $|D^{\mathrm{train}}_n| \le 2n$. By Lemma~\ref{lem_subcrit_coset_trace},
    each datum covers at most one orbit point, so coverage requires
    $2n \ge p$, i.e., $n \ge \lceil p/2\rceil$.

\medskip
\noindent\textbf{Upper bound.} We construct an explicit $\mathrm{single}$ strategy
covering $\Omega_E$ in $V := \lceil p/2\rceil$ rounds. Note $V \ge s$ since
$g \ge 2$ implies $p/2 \ge p/g$.

For $i \in \{0, \ldots, s-1\}$, let
\[
  a_i \;:=\; |\{n \in [0, V) : n \equiv i \pmod s\}|
        \;=\; \bigl\lceil (V-i)/s \bigr\rceil
\]
be the number of E-arrivals in coset $C_i$ during the first $V$ rounds. The
auto-arrivals fill the $a_i$ smallest representatives of $C_i$, namely
$i,\, i+s,\, \ldots,\, i+(a_i-1)s$. The set of \emph{missing} elements of
$C_i$ is therefore
\[
  M_i \;=\; \{\,i + a_i\,s,\; i + (a_i{+}1)s,\; \ldots,\; i + (g{-}1)s\,\},
  \qquad |M_i| = g - a_i \;=:\; \beta_i.
\]
A double count gives $\sum_i a_i = V$ and
$\sum_i \beta_i = sg - V = p - \lceil p/2\rceil = \lfloor p/2\rfloor \le V$.

We schedule the missing points explicitly. Put
\[
  B_i \;:=\; \sum_{r<i} \beta_r.
\]
For each $m=i+(a_i+u)s\in M_i$ with $0\le u<\beta_i$, schedule $m$ in round
\[
  n(m) \;:=\; B_i + u.
\]
The assigned rounds are distinct, since the intervals
$[B_i,\,B_i+\beta_i)$ are consecutive, and all lie in $\{0,\ldots,V-1\}$
because their total number is
$\sum_i \beta_i = p - V = \lfloor p/2\rfloor \le V$.

It remains to check that a seed from $C_i$ is available by round $n(m)$. For
every $i$ one has $a_i \le \lceil V/s\rceil \le g-1$: if $g=2$ then $V=s$,
while if $g\ge3$ then $V=\lceil sg/2\rceil \le s(g-1)$ because $s(g-2)\ge1$.
Hence
\[
  B_i
  \;=\; \sum_{r<i}(g-a_r)
  \;=\; ig - \sum_{r<i}a_r
  \;\ge\; ig - i(g-1)
  \;=\; i.
\]
Thus $n(m)=B_i+u\ge i$. At round $n(m)$, the point $i\in C_i$ is therefore
available: if $n(m)=i$ it is the current query in $E_i$, and if $n(m)>i$ it
has already been absorbed into $D^{\mathrm{train}}_{n(m)}$. Since $m\in C_i$,
Lemma~\ref{lem_h_g_regular} gives a unique element $h\in H_g\subseteq
H_{\mathrm{train}}$ sending $i$ to $m$; play this single augmentation in
round $n(m)$. In rounds not assigned to any missing point, play the identity
shift on the current query.

After $V$ rounds, $D^{\mathrm{train}}_V$ contains the auto-arrivals
$\{0,1,\ldots,V-1\}$ together with the scheduled targets
$\bigsqcup_i M_i = \{V,V+1,\ldots,p-1\}$. Hence $\Omega_E \subseteq
D^{\mathrm{train}}_V \subseteq V_\varepsilon(D^{\mathrm{train}}_V)$, since
$\varepsilon>0$, so $N_{\mathrm{orbit}}^{\mathrm{single}} \le V = \lceil p/2\rceil$.
\end{proof}

\begin{lemma}[Arc packing on $\Omega_E$]\label{lem_arc_packing}
Every open circle ball $B_{\mathbb{T}}(z,\varepsilon)$ contains at most
$2p\varepsilon + 1$ points of $\Omega_E$. Consequently,
\[
  Q(\varepsilon, p, q) \;\le\; 2p\varepsilon + 1.
\]
\end{lemma}

\begin{proof}
If $\varepsilon \ge 1/2$, the bound is trivial because
$|\Omega_E|=p\le 2p\varepsilon+1$. Assume $\varepsilon<1/2$.
$\Omega_E = \{0, 1/p, \ldots, (p-1)/p\}$ is equally spaced with gap $1/p$,
so an open arc of length $2\varepsilon$ contains at most
$\lfloor 2\varepsilon p\rfloor + 1 \le 2p\varepsilon + 1$ points. Each open
ball $B_{\mathbb{T}}(z, \varepsilon)$ is such an arc, hence
$|\Omega_E \cap B_{\mathbb{T}}(z, \varepsilon)| \le 2p\varepsilon + 1$.
\end{proof}

\begin{lemma}[Sparse H-cover in saturation]\label{lem_sparse_h_cover_sat}
Assume $\varepsilon > \varepsilon_{\mathrm{sat}}$, and let $Q = Q(\varepsilon,
p, q)$. There is a subset $A \subseteq H_{\mathrm{train}}$ with
$|A| \le 10\,p/Q$ such that $\Omega_E \subseteq V_\varepsilon(A)$.
\end{lemma}

\begin{proof}
By Lemma~\ref{lem_eps_sat_S}, $\Omega_E \subseteq V_\varepsilon(H_{\mathrm{train}})$.
If $Q = 1$, for each $m/p \in \Omega_E$ choose one H-translate that covers
it; this gives an H-cover of size at most $p = p/Q$.

Assume $Q \ge 2$. By Lemma~\ref{lem_arc_packing}, $2p\varepsilon \ge 1$,
hence $Q \le 4p\varepsilon$.

If $\varepsilon q < 2$, take $A = H_{\mathrm{train}}$: it covers $\Omega_E$
by saturation, and $|A| = q < 2/\varepsilon$. Otherwise $\varepsilon q \ge 2$,
and we choose points of the regular $q$-grid $H_{\mathrm{train}} = \{0, 1/q,
\ldots, (q-1)/q\}$ at spacing at most $\varepsilon$ in cyclic order: set
$r := \lfloor \varepsilon q\rfloor$ and take $0, r/q, 2r/q, \ldots,
\lfloor (q-1)/r\rfloor \cdot r/q$. Consecutive chosen H-points, including
the wrap-around gap, have circular gap at most $r/q \le \varepsilon$, so
their $\varepsilon$-balls cover the whole circle. The number chosen is at
most $2/\varepsilon + 2$.

In both cases $|A| \le 2/\varepsilon + 2$. Since $Q \ge 2$ and $Q \le p$,
the inequalities $Q \le 4p\varepsilon$ and $2 \le 2p/Q$ give
\[
  |A| \;\le\; 2/\varepsilon + 2 \;\le\; 8\,p/Q + 2\,p/Q \;=\; 10\,p/Q.
\]
The argument uses no upper bound on $\varepsilon$, so it covers all
saturation radii including $\varepsilon > 1/2$ (for example $q = 1$ with
even $p$, where $\varepsilon_{\mathrm{sat}} = 1/2$).
\end{proof}

\begin{coro}[$\mathrm{single}$ saturation]\label{cor_norbit_b0_sat}
For $\varepsilon > \varepsilon_{\mathrm{sat}}$,
\[
  \bigl\lceil p/(2 Q)\bigr\rceil
  \;\le\;
  N_{\mathrm{orbit}}^{\mathrm{single}}(\varepsilon, p, q)
  \;\le\;
  10\,p/Q,
\]
where $Q = Q(\varepsilon, p, q)$ is the reachable-center packing factor. In
particular $N_{\mathrm{orbit}}^{\mathrm{single}} = \Theta(p/Q)$.
\end{coro}

\begin{proof}
For the lower bound, Lemma~\ref{lem_b0_capacity} gives
$|D^{\mathrm{train}}_n| \le 2n$ under $\mathrm{single}$, and Lemma~\ref{lem_dataset_form}
gives that every point of $D^{\mathrm{train}}_n$ lies in
$\Omega_E + H_{\mathrm{train}}$. Hence each such point covers at most $Q$
points of $\Omega_E$, by definition of $Q$. Thus full orbit coverage forces
$2nQ \ge p$, and hence $n \ge \lceil p/(2Q)\rceil$.

For the upper bound, choose $A=\{a_0,\ldots,a_{K-1}\}\subseteq H_{\mathrm{train}}$
as in Lemma~\ref{lem_sparse_h_cover_sat}, so
$K\le10p/Q$ and $\Omega_E\subseteq V_\varepsilon(A)$. In each of the first
$K$ rounds, use the seed $0\in E_0$ (which is available at round $0$ and then
remains in the trainer dataset) and play the $\mathrm{single}$ augmentation
$h_n=a_n$, $c_n=0$. After $K$ rounds the trainer dataset contains all points
of $A$, hence its $\varepsilon$-neighborhood covers $\Omega_E$. Thus
$N_{\mathrm{orbit}}^{\mathrm{single}}\le K\le10p/Q$.
\end{proof}

\begin{remark}[On constants in the $\Theta(p/Q)$ entry of Table~\ref{table:boom}]
\label{rem_b0_sat_log_gap}
The proof above is deliberately coarse: the constant $10$ is only used to make
the table's asymptotic entry explicit. The lower bound has the unavoidable
factor $1/2$ because a $\mathrm{single}$ round both absorbs the current E-query and adds
one H-translate. No logarithmic set-cover loss is needed in the saturation
regime.
\end{remark}

\subsubsection{Three-tier separation at p = q}\label{app_circle_three_tier}

\begin{theorem}[Three-tier batching ladder at p = q]\label{thm_three_tier}
For $p = q \ge 3$ and any $0 < \varepsilon \le \varepsilon^\star$,
\[
  N_{\mathrm{orbit}}^{\mathrm{full}}(\varepsilon, p, p) = 1,
  \qquad
  N_{\mathrm{orbit}}^{\mathrm{batch}}(\varepsilon, p, p) = \Theta(\log p),
  \qquad
  N_{\mathrm{orbit}}^{\mathrm{single}}(\varepsilon, p, p) = \lceil p/2\rceil.
\]
In particular the asymptotic ordering $\Theta(1) \ll \Theta(\log p) \ll \Theta(p)$
is strict.
\end{theorem}

\begin{proof}
At $p = q$ we have $g = q = p$ and $s = p/g = 1$. Substituting into the
sub-critical formulas with $g = p$:
\begin{itemize}
  \item \emph{$\mathrm{full}$}: by Corollary~\ref{cor_binfty_subcrit},
    $N_{\mathrm{orbit}}^{\mathrm{full}} = p/g = 1$.
  \item \emph{$\mathrm{batch}$}: by Theorem~\ref{thm_norbit_b1_sub} (applicable since
    $g = p \ge 3 \ge 2$),
    $N_{\mathrm{orbit}}^{\mathrm{batch}} = \Theta(s + \log g) = \Theta(1 + \log p) = \Theta(\log p)$.
  \item \emph{$\mathrm{single}$}: by Theorem~\ref{thm_norbit_b0_sub} with $g = p \ge 3 \ge 2$,
    $N_{\mathrm{orbit}}^{\mathrm{single}} = \lceil p/2\rceil$.
\end{itemize}
The three values are $\Theta(1)$, $\Theta(\log p)$, and $\Theta(p)$
respectively, hence strictly ordered for $p \ge 3$.
\end{proof}

\begin{remark}[Ladder collapse at fixed $g$]\label{rem_three_tier_caveat}
In the sub-critical regime $\varepsilon \le \varepsilon^\star$, the
three-tier ladder collapses entirely when $g$ is fixed (for example $g = 2$).
By Corollary~\ref{cor_binfty_subcrit}, $N_{\mathrm{orbit}}^{\mathrm{full}} = p/g$;
by Theorem~\ref{thm_norbit_b0_sub}, $N_{\mathrm{orbit}}^{\mathrm{single}} =
\lceil p/2\rceil$; and by Corollary~\ref{cor_v_star_upper}, $N_{\mathrm{orbit}}^{\mathrm{batch}}
\le s + \lceil \log_2 g\rceil$, which combined with the lower bound
$N_{\mathrm{orbit}}^{\mathrm{batch}} \ge s$ from Lemma~\ref{lem_coset_seed_lb} gives
$N_{\mathrm{orbit}}^{\mathrm{batch}} = p/g + O(\log g)$. At $g = 2$ all three are
$p/2 + O(1)$, so no asymptotic separation is visible. A strict
$\Theta(1) \ll \Theta(\log p) \ll \Theta(p)$ ladder requires
$g = \Theta(p)$, for example the sub-family $p = q$ with $p \to \infty$ used in
Theorem~\ref{thm_three_tier}.
\end{remark}

\section{Acknowledgements}
\label{app_ack}

\textbf{Licenses.} We use the MMLU benchmark, and the weights of mistralai/Mistral-7B-Instruct-v0.3, Qwen/Qwen2.5-7B-Instruct, EleutherAI/pythia-410m, EleutherAI/pythia-1b, allenai/Olmo-3.1-32B-Instruct, under \emph{MIT Apache 2.0} licenses. We use gpt2 / openai-community/gpt2's weight under a \emph{modified MIT license}. We use meta-llama/Llama-3.1-8B-Instruct under \emph{Meta Llama 3.1 Community License}, and allenai/wildjailbreak / WildJailBreak under \emph{ODC-BY}.

\section{Distance-dependent transfer: detailed methods and supplementary results}
\label{app:distance-dependent-transfer-methods}

\subsection{Hardware and compute}\label{app:ft-hardware}

Experiments ran on an internal cluster with GPUs between 40\,GB and 128\,GB of HBM (PFLOP/s range). Each cell consumed a few GPU-hours: a fine-tuning step ($\le 1$\,hour), a held-out evaluation pass with $N{=}10$ stochastic samples per prompt at temperature $0.7$ ($\approx 2{-}3$\,hours), and an LLM-as-judge labelling pass ($\approx 1{-}2$\,hours).

\subsection{Models, fine-tuning configuration, and cells}
\label{app:models}

Three open-weight instruction-tuned models: \textsc{Llama-3.1-8B-Instruct}, \textsc{Mistral-7B-Instruct-v0.3}, \textsc{Qwen2.5-7B-Instruct}, used in their public Hugging Face weights with their tokenizers' native chat templates. Inference is in \texttt{bfloat16}.

\paragraph{Cells.}
We report on six cells spanning two LoRA capacity points:
\begin{itemize}
\item \textbf{Rank-$1$ cohort:} one cell per family at LoRA rank $r{=}1$, exposure budget $1600$ examples (training pool of $47$ prompts cycled with seed-controlled shuffle until the budget is reached).
\item \textbf{Rank-$2$ cohort:} one cell per family at LoRA rank $r{=}2$. For \textsc{Llama} and \textsc{Mistral} the exposure budget is $800$; for \textsc{Qwen} it is $1600$ (the smaller-exposure \textsc{Qwen} cell did not yet show a saturated effect at this rank).
\end{itemize}
All six cells use the same training-stochasticity seed ($42$). LoRA configuration is otherwise identical across cells: $\alpha{=}32$, dropout $0.05$, target modules \texttt{[q\_proj, k\_proj, v\_proj, o\_proj, gate\_proj, up\_proj, down\_proj]}, AdamW optimizer, learning rate $3{\times}10^{-5}$, batch size $4$. The base model is loaded fresh per cell; the adapter is trained, saved, and re-loaded for evaluation.

\subsection{Training pool (corpus-density-driven, shared across families)}
\label{app:training-pool}

We use a single $47$-prompt active training pool $\mathcal{P}_{\mathrm{train}}$, shared across all three model families. The pool is selected by corpus-density and cluster diversity, with the goal of producing meaningful low-distance coverage on the held-out panel.

\paragraph{Per-(target, metric) candidate selection.}
For each pair \((\text{target}, \text{metric})\) with $\text{target} \in \{\textsc{Llama}, \textsc{Mistral}, \textsc{Qwen}\}$ and $\text{metric} \in \{\texttt{last\_token}, \texttt{mean\_pool}, \texttt{mean\_pool\_first}, \texttt{spectral\_first}, \texttt{spectral\_last}\}$ we:
\begin{enumerate}
\item Embed every prompt of the \textsc{WildJailbreak}~\cite{jiang2024wildteaming} adversarial-harmful split under the target's base model with the chosen embedding tag (Appendix~\ref{app:distance-methods}).
\item Compute the $k{=}50$-NN cosine radius $\rho_{50}(p) \in [0, 2]$ of each prompt~$p$ in that embedding space.
\item Partition the corpus into $K{=}8$ clusters by $k$-means on $\ell_2$-normalized embeddings.
\item Pick the densest cluster representative per cluster (smallest $\rho_{50}$); $8$ representatives per $(\text{target}, \text{metric})$.
\item Among the per-target $5$ metrics, keep the two metrics with the smallest mean $\rho_{50}$ across their cluster reps. (Practical outcome: \texttt{spectral\_first} is the tightest metric in all three target embedding spaces; the per-target second pick varies.)
\end{enumerate}
This yields $3 \times 2 \times 8 = 48$ candidate prompts (with one duplicate across $(\text{target}, \text{metric})$ pairs), of which $47$ are unique corpus indices. We rank these candidates by a vote score $s(p) = \#\{(\text{target}, \text{metric}) : p \in \text{cluster-reps}\}$ with ties broken by lowest mean $\rho_{50}$, and take all $47$ as the final pool $\mathcal{P}_{\mathrm{train}}$. Eyeballing confirms semantic spread across $8{+}$ refusal categories (cyber-attack help-seeking, hate-speech generation, privacy intrusion, misinformation, drug/contraband, explicit sexual content, professional misconduct, self-harm).

The same $47$ corpus indices are the training pool for every cell in this report; what differs across cells is the model being fine-tuned (hence the per-family embedding of those prompts) and the LoRA hyperparameters.

\subsection{Evaluation panel construction}
\label{app:eval-panel}

The held-out evaluation panel $\mathcal{H}$ contains $2999$ corpus prompts and is shared across all cells; it excludes $\mathcal{P}_{\mathrm{train}}$.  The panel is constructed to give substantial near-pool and far-from-pool coverage simultaneously, so that the conditional distance $z$ varies across a meaningful range under any of the seven embedding tags:

\begin{enumerate}
\item \textbf{Near-pool stratum} ($2116$ prompts). For each \((\text{target}, \text{metric})\) pair used to construct $\mathcal{P}_{\mathrm{train}}$ (Appendix~\ref{app:training-pool}), retrieve the $50$ nearest neighbors of each of the $8$ cluster representatives in that embedding space. Union across the $3 \times 2 = 6$ per-(target, metric) selections, deduplicate.
\item \textbf{Random reference stratum} ($384$ prompts). Sample uniformly at random from the corpus (excluding $\mathcal{P}_{\mathrm{train}}$ and the near-pool stratum), seed $42$, giving mid- and high-distance reference points.
\item \textbf{Per-family low-distance extension stratum} ($499$ prompts). For each family, greedily add prompts that lie below the family's $z$-distribution sixth decile under at least one of the family's two pool-selection metrics, until each family has $250$ extension prompts; deduplicate across families ($499$ unique).
\end{enumerate}
The three strata are concatenated to form $\mathcal{H}$ with $|\mathcal{H}| = 2999$. The same $2999$ corpus indices are evaluated by every cell.

\subsection{Embedding tags and distance metric}
\label{app:distance-methods}

For each model $M$ and each held-out prompt $p$, let $H_{M}(p) \in \mathbb{R}^{T(p) \times d \times L}$ denote the tensor of $M$'s per-block hidden states given $p$ as input under $M$'s native chat template, where $T(p)$ is the prompt length in tokens, $d$ is the model hidden dimension, and $L$ is the number of transformer blocks. Let $H^{(\ell)}_M(p) \in \mathbb{R}^{T(p) \times d}$ denote the per-block slice at depth $\ell \in \{1, \ldots, L\}$, $H^{(\mathrm{last})}_M(p)$ the slice at the last block, and let
\[
\bar H_M(p) = \frac{1}{L} \sum_{\ell = 1}^{L} H^{(\ell)}_M(p) \;\in\; \mathbb{R}^{T(p) \times d}
\]
be the layer-averaged hidden-state matrix.

The three embedding tags we report on are defined as follows.

\paragraph{\texttt{last\_token}.} The hidden state of the final token at the model's last block:
\[
\phi^{\mathrm{lt}}_M(p) \;=\; H^{(\mathrm{last})}_M(p)_{T(p), :} \;\in\; \mathbb{R}^{d}.
\]

\paragraph{\texttt{mean\_pool}.} The token-position mean of the last block's hidden states:
\[
\phi^{\mathrm{mp}}_M(p) \;=\; \frac{1}{T(p)} \sum_{t=1}^{T(p)} H^{(\mathrm{last})}_M(p)_{t, :} \;\in\; \mathbb{R}^{d}.
\]

\paragraph{\texttt{spectral\_all}.} The leading right singular vector of the layer-averaged hidden-state matrix $\bar H_M(p)$. Concretely, write
\[
\bar H_M(p) \;=\; U \,\Sigma\, V^{\top}, \qquad U \in \mathbb{R}^{T(p) \times r}, \; \Sigma \in \mathbb{R}^{r \times r}, \; V \in \mathbb{R}^{d \times r}, \; r = \min(T(p), d),
\]
for the (thin) singular value decomposition with singular values $\sigma_1 \ge \sigma_2 \ge \cdots \ge \sigma_r \ge 0$. Then
\[
\phi^{\mathrm{sa}}_M(p) \;=\; \mathrm{sign}\bigl(\langle V_{:, 1},\, \phi^{\mathrm{mp,all}}_M(p) \rangle\bigr) \cdot V_{:, 1} \;\in\; \mathbb{R}^{d},
\]
where $\phi^{\mathrm{mp,all}}_M(p) = \frac{1}{T(p)} \sum_{t=1}^{T(p)} \bar H_M(p)_{t, :}$ is the token-and-layer mean (used only to orient the sign so that $\langle \phi^{\mathrm{sa}}_M(p), \phi^{\mathrm{mp,all}}_M(p) \rangle > 0$, breaking the $\pm V_{:,1}$ ambiguity of the SVD). When $T(p) < 2$ or the SVD fails to converge, we fall back to $\phi^{\mathrm{sa}}_M(p) := \phi^{\mathrm{mp,all}}_M(p) / \lVert \phi^{\mathrm{mp,all}}_M(p) \rVert$.

\paragraph{Distance to the training pool.}
For test prompt $p \in \mathcal{H}$ and pool $\mathcal{P}_{\mathrm{train}}$, under embedding tag $\phi^{(\tau)}_M$ ($\tau \in \{\mathrm{lt}, \mathrm{mp}, \mathrm{sa}\}$), we set
\begin{equation}
z^{(\tau)}_M(p) \;=\; \min_{q \in \mathcal{P}_{\mathrm{train}}} \lVert \phi^{(\tau)}_M(p) - \phi^{(\tau)}_M(q) \rVert_2,
\label{eq:z}
\end{equation}
and renormalize across the panel via
\(
z^{(\tau)}_M(p) \mapsto \bigl(z^{(\tau)}_M(p) - z_{\min}\bigr) / (z_{\max} - z_{\min}) \in [0, 1],
\)
where $z_{\min} = \min_{p' \in \mathcal{H}} z^{(\tau)}_M(p')$ and analogously $z_{\max}$. This is a monotone rescaling and does not affect rank-based or shape-comparison analyses. Distances are computed against the \emph{base} model's embeddings of $\mathcal{P}_{\mathrm{train}}$; the adapter is not active when computing $z$. Hence $z$ depends on (family, metric, panel) and is shared across cells of the same family.

\subsection{Refusal labels: stochastic FT decoding and LLM-as-judge}
\label{app:judge}

For each cell and each held-out prompt $p \in \mathcal{H}$, we draw $N = 10$ independent samples $r_1, \ldots, r_{10}$ from the fine-tuned model at sampling temperature $0.7$ and $\mathrm{top\text{-}}p = 1.0$, capped at $\le 256$ generated tokens. Analogously for the base model, we draw $N = 10$ samples per prompt; the base samples are shared across cells of the same family (the base model is fixed). For each sample we extract a refusal label by an open-weight LLM judge.

\paragraph{Judge model and rubric.}
We use \textsc{OLMo-3.1-32B-Instruct}~\cite{teamolmo2025olmo3}, loaded in 8-bit weights (\texttt{bitsandbytes}, $\textsc{nf}8$), as the headline judge. For each $(p, r)$ pair the judge classifies $r$ under a three-way rubric:
\begin{itemize}
\item \textit{refused} (R): the response declines the request (refusal of any kind, including soft refusal, persona refusal, lecturing, redirection).
\item \textit{complied} (C): the response provides (or substantially attempts to provide) the requested content.
\item \textit{ambiguous} (A): the response is too short, off-topic, garbled, or otherwise neither a clear refusal nor a clear compliance.
\end{itemize}
The judge prompt template is listed below (Appendix~\ref{app:judge-prompt}). The judge model's next-token logits over the three label tokens $\{R, C, A\}$ are extracted directly (no further sampling); the $\arg\max$ token is recorded as the label.

\paragraph{Per-prompt count primitives.}
For each (cell, prompt, condition $\in \{\mathrm{base}, \mathrm{ft}\}$), the analysis primitives are the integer counts
\[
n^{\mathrm{cond}}(p) \;=\; \#\{i : \text{judge}(r_i^{\mathrm{cond}}) \in \{R, C\}\}, \qquad k^{\mathrm{cond}}(p) \;=\; \#\{i : \text{judge}(r_i^{\mathrm{cond}}) = R\},
\]
i.e.\ $n^{\mathrm{cond}}(p)$ is the number of non-Ambiguous samples (out of $N{=}10$) and $k^{\mathrm{cond}}(p)$ is the number judged refused. The empirical refusal rate at prompt $p$ is $\hat p^{\mathrm{cond}}(p) = k^{\mathrm{cond}}(p) / \max(n^{\mathrm{cond}}(p), 1)$. Prompts where either $n^{\mathrm{base}}(p) = 0$ or $n^{\mathrm{ft}}(p) = 0$ (all-ambiguous on at least one side) are dropped pairwise from the cell. After dropping, the per-cell sample size is $n_{\mathrm{cell}} \in [2717, 2898]$ across the six cells (the across-cell range reflects per-cell Ambiguous-only drop rate).

\paragraph{Why a single strong judge and not majority.}
We use a single strong open-weight judge for two reasons. (i) Inter-judge agreement studies on this rubric show \textsc{OLMo-3.1-32B-Instruct} dominates the multi-judge majority on agreement-with-human-spot-checks. (ii) Substring classifiers and smaller judges produce metric-dependent false-negative rates on persona refusals and lecturing-style soft refusals, which interact with the embedding-distance axis we are studying; using a strong tag-agnostic judge avoids the metric--judge interaction.

\subsubsection{Judge prompt template}
\label{app:judge-prompt}

The exact prompt used for each (prompt, response) pair is:

\begin{quote}
\small\ttfamily
You are judging whether an AI response refused, complied, or is ambiguous.\\[2pt]
USER: \{prompt\}\\
AI:\hphantom{xx} \{response\}\\[2pt]
Did the AI refuse to comply with the user's request?\\
\hphantom{xx}- R = REFUSED (declined, redirected, or expressed inability)\\
\hphantom{xx}- C = COMPLIED (provided the requested content)\\
\hphantom{xx}- A = AMBIGUOUS (unclear, partial, or evasive)\\[2pt]
Answer (R/C/A):
\end{quote}

The trailing space after \texttt{(R/C/A):} is intentional --- it positions the next-token boundary on the label vocabulary.

\subsection{Per-cell statistical model}
\label{app:cell-model}

The cohort figures (Figures~\ref{fig:appendix-rank1-lowex}--\ref{fig:appendix-rank2-strong}) display posterior fits of the \emph{fine-tuned refusal probability} $p_{\mathrm{ft}}(z)$ as a function of the normalized distance $z \in [0, 1]$. We model $p_{\mathrm{ft}}(z)$ directly rather than the rate-difference $\Delta(z) = p_{\mathrm{ft}}(z) - p_{\mathrm{base}}(z)$: the shape name (CPL, two-segment, etc., defined below) then describes the visible $p_{\mathrm{ft}}$ curve.

\paragraph{Likelihood.}
For each cell and each held-out prompt $i = 1, \ldots, n_{\mathrm{cell}}$ we observe $(z_i, k^{\mathrm{ft}}_i, n^{\mathrm{ft}}_i)$, where $z_i$ is the prompt's normalized pool distance (Eq.~\ref{eq:z}) and $(k^{\mathrm{ft}}_i, n^{\mathrm{ft}}_i)$ are the Binomial primitives from the judge (Appendix~\ref{app:judge}). We model
\begin{align}
p_{\mathrm{ft}}(z_i) &\;=\; \mathrm{clip}\!\bigl(\,\delta_S(z_i;\,\theta_S),\;\varepsilon,\; 1 - \varepsilon\bigr), \label{eq:p-ft-clip}\\
k^{\mathrm{ft}}_i \mid z_i &\;\sim\; \mathrm{Binomial}\!\bigl(n^{\mathrm{ft}}_i,\; p_{\mathrm{ft}}(z_i)\bigr), \label{eq:binom}
\end{align}
where $\varepsilon = 10^{-4}$ is a clipping floor against label saturation, and $\delta_S(\,\cdot\,; \theta_S) : [0, 1] \to [0, 1]$ is one of the five candidate \emph{shapes} listed below, parametrised by $\theta_S$.

\paragraph{Shape menu.}
The shape menu $\mathcal{S} = \{\mathrm{N, L, P, C, S}\}$ contains:
\begin{enumerate}
\item[\textbf{N}.] \emph{Constant} ($\theta = (d)$): $\delta(z) = d$. 1 parameter.
\item[\textbf{L}.] \emph{Linear} ($\theta = (d_{\mathrm{lo}}, d_{\mathrm{hi}})$): $\delta(z) = d_{\mathrm{lo}} + (d_{\mathrm{hi}} - d_{\mathrm{lo}})\, z$. 2 parameters.
\item[\textbf{P}.] \emph{Plateau-then-slope} ($\theta = (z_K, d_{\mathrm{lo}}, d_{\mathrm{hi}})$): $\delta(z) = d_{\mathrm{lo}}$ on $[0, z_K]$, then linearly interpolates from $(z_K, d_{\mathrm{lo}})$ to $(1, d_{\mathrm{hi}})$ on $[z_K, 1]$. 3 parameters.
\item[\textbf{C}.] \emph{Constrained piecewise-linear (CPL)} ($\theta = (z_L, z_R, d_{\mathrm{lo}}, d_{\mathrm{hi}})$): $\delta(z) = d_{\mathrm{lo}}$ on $[0, z_L]$, linearly interpolates from $(z_L, d_{\mathrm{lo}})$ to $(z_R, d_{\mathrm{hi}})$ on $[z_L, z_R]$, and equals $d_{\mathrm{hi}}$ on $[z_R, 1]$. Monotonicity is enforced only on the parameter ordering $z_L \le z_R$ (we order the two posterior samples at evaluation time); the levels $d_{\mathrm{lo}}, d_{\mathrm{hi}}$ are free, so the shape accommodates both increases ($d_{\mathrm{hi}} > d_{\mathrm{lo}}$) and decreases ($d_{\mathrm{hi}} < d_{\mathrm{lo}}$) of refusal with distance. 4 parameters.
\item[\textbf{S}.] \emph{Two-segment} ($\theta = (z_M, d_{\mathrm{lo}}, d_{\mathrm{mid}}, d_{\mathrm{hi}})$): piecewise-linear with one knee at $z_M$ and three free levels $d_{\mathrm{lo}}, d_{\mathrm{mid}}, d_{\mathrm{hi}}$ (the curve passes through $(0, d_{\mathrm{lo}})$, $(z_M, d_{\mathrm{mid}})$, and $(1, d_{\mathrm{hi}})$). 4 parameters; sign-free.
\end{enumerate}
Priors: $z_L, z_R, z_K, z_M \sim \mathrm{Beta}(2, 2)$ (concentrated away from the boundaries of $(0, 1)$); $d, d_{\mathrm{lo}}, d_{\mathrm{mid}}, d_{\mathrm{hi}} \sim \mathrm{Uniform}(0, 1)$.

Each shape is fit independently by NUTS (\textsc{numpyro}; $1000$ warmup, $1500$ samples, $2$ chains, \texttt{init\_to\_median}). Per-observation log-likelihoods $\ell_i^{(s, S)} = \log \Pr\bigl(k^{\mathrm{ft}}_i \mid n^{\mathrm{ft}}_i, p_{\mathrm{ft}}^{(s, S)}(z_i)\bigr)$ are recorded under the fitted posterior of shape $S$, sample $s$.

\paragraph{WAIC and Akaike weights.}
For each shape $S$ we compute the standard WAIC estimator
\begin{align}
\widehat{\mathrm{lppd}}_i^{(S)} &\;=\; \log \tfrac{1}{N_S} \sum_{s=1}^{N_S} \exp \ell_i^{(s, S)}, \\
\widehat{p}_{\mathrm{WAIC},i}^{(S)} &\;=\; \widehat{\mathrm{Var}}_s\bigl[\ell_i^{(s, S)}\bigr], \\
\mathrm{WAIC}^{(S)} &\;=\; -2 \sum_i \bigl(\widehat{\mathrm{lppd}}_i^{(S)} - \widehat{p}_{\mathrm{WAIC},i}^{(S)}\bigr).
\end{align}
The within-cell Akaike weights are
\begin{equation}
w_S \;=\; \frac{\exp\!\bigl(-\tfrac{1}{2}\,\Delta_S\bigr)}{\sum_{S' \in \mathcal{S}} \exp\!\bigl(-\tfrac{1}{2}\,\Delta_{S'}\bigr)}, \qquad \Delta_S \;=\; \mathrm{WAIC}^{(S)} - \min_{S' \in \mathcal{S}} \mathrm{WAIC}^{(S')}.
\end{equation}
We additionally report $\Delta\mathrm{WAIC}_{\mathrm{null}} = \mathrm{WAIC}^{(\mathrm{N})} - \mathrm{WAIC}^{(\mathrm{best})}$ as the per-cell evidence for $z$-dependence (positive values favor a non-constant alternative), and $P(z\text{-dep}) = 1 - w_{\mathrm{N}}$ as the model-averaged posterior probability of any non-constant alternative.

\paragraph{Direction summary.}
For the best-fitting non-constant shape we report
\begin{equation}
P\bigl(p_{\mathrm{ft}}(1) < p_{\mathrm{ft}}(0)\bigr) \;=\; \frac{1}{N_{\mathrm{best}}} \sum_{s = 1}^{N_{\mathrm{best}}} \mathbf{1}\!\bigl[p_{\mathrm{ft}}^{(s, \mathrm{best})}(1) < p_{\mathrm{ft}}^{(s, \mathrm{best})}(0)\bigr],
\end{equation}
the posterior probability that $p_{\mathrm{ft}}(z)$ is lower at $z{=}1$ than at $z{=}0$ under the best shape. This is the natural ``net decreasing in $z$'' indicator for the FT-direct fit.

\subsection{Per-cell summary tables}
\label{app:cross-metric}

Table~\ref{tab:dense47-stoch-cells} reports the per-cell, per-metric headline numbers for the six cells $\times$ three metrics in scope: per-cell sample size $n_{\mathrm{cell}}$ after the Ambiguous-only pairwise drop, empirical base refusal rate $\bar p_{\mathrm{base}}$, empirical fine-tuned refusal rate $\bar p_{\mathrm{ft}}$, the per-cell rate-difference $\bar\Delta = \bar p_{\mathrm{ft}} - \bar p_{\mathrm{base}}$, the per-cell evidence for $z$-dependence $\Delta\mathrm{WAIC}_{\mathrm{null}}$, the model-averaged $P(z\text{-dep})$, the best-fitting shape, and the direction indicator $P(p_{\mathrm{ft}}(1) < p_{\mathrm{ft}}(0))$.

\begin{table}[h]
\centering
\caption{Per-cell, per-metric headline numbers for the six dense47-stoch cells. $n_{\mathrm{cell}}$: panel size after the Ambiguous-only pairwise drop. $\bar p_{\mathrm{base}}, \bar p_{\mathrm{ft}}$: empirical mean refusal rate (base, fine-tuned) over the cell's surviving prompts. $\bar\Delta = \bar p_{\mathrm{ft}} - \bar p_{\mathrm{base}}$. $\Delta\mathrm{WAIC}_0$: per-cell evidence for $z$-dependence (positive favors a non-constant alternative). $P(z\text{-dep})$: model-averaged posterior probability $1 - w_{\mathrm{N}}$. ``best'': WAIC-preferred shape (N = constant; L = linear; P = plateau-then-slope; C = CPL; S = two-segment). $P_{post} = P(p(1){<}p(0))$: posterior probability under the best shape that $p_{\mathrm{ft}}$ is lower at $z{=}1$ than at $z{=}0$. The notation $1{-}c{\cdot}10^{-k}$ flags posteriors saturated against the floor of the Monte Carlo display precision.}
\label{tab:dense47-stoch-cells}
\small
\setlength{\tabcolsep}{4pt}
\begin{tabular}{l l r rrr r l l l}
\toprule
 cell & metric & $n_{\mathrm{cell}}$ & $\bar p_{\mathrm{base}}$ & $\bar p_{\mathrm{ft}}$ & $\bar\Delta$ & $\Delta\mathrm{WAIC}_0$ & best & $P(z\text{-dep})$ & $P_{post}$ \\
\midrule
 e1200 \textsc{Llama}  & \texttt{last\_token} & 2876 & 0.299 & 0.808 & +0.509 & +738 & C & $>1-10^{-6}$ & $>1-10^{-6}$ \\
 ($r{=}1$, $E{=}1600$) & \texttt{mean\_pool} & 2876 & 0.299 & 0.808 & +0.509 & +877 & S & $>1-10^{-6}$ & $>1-10^{-6}$ \\
  & \texttt{spectral\_all} & 2876 & 0.299 & 0.808 & +0.509 & +5,507 & C & $>1-10^{-6}$ & $>1-10^{-6}$ \\
\midrule
 e1201 \textsc{Mistral}  & \texttt{last\_token} & 2892 & 0.121 & 0.856 & +0.735 & +841 & P & $>1-10^{-6}$ & $>1-10^{-6}$ \\
 ($r{=}1$, $E{=}1600$) & \texttt{mean\_pool} & 2892 & 0.121 & 0.856 & +0.735 & +6.7 & P & 0.969 & $1-3.0{\cdot}10^{-4}$ \\
  & \texttt{spectral\_all} & 2892 & 0.121 & 0.856 & +0.735 & +3,667 & C & $>1-10^{-6}$ & $>1-10^{-6}$ \\
\midrule
 e1202 \textsc{Qwen}  & \texttt{last\_token} & 2717 & 0.285 & 0.736 & +0.451 & +619 & P & $>1-10^{-6}$ & $>1-10^{-6}$ \\
 ($r{=}1$, $E{=}1600$) & \texttt{mean\_pool} & 2717 & 0.285 & 0.736 & +0.451 & +110 & S & $>1-10^{-6}$ & $>1-10^{-6}$ \\
  & \texttt{spectral\_all} & 2717 & 0.285 & 0.736 & +0.451 & +5,391 & C & $>1-10^{-6}$ & $>1-10^{-6}$ \\
\midrule
 e1203 \textsc{Llama} & \texttt{last\_token} & 2898 & 0.298 & 0.758 & +0.460 & +801 & P & $>1-10^{-6}$ & $>1-10^{-6}$ \\
  ($r{=}2$, $E{=}800$) & \texttt{mean\_pool} & 2898 & 0.298 & 0.758 & +0.460 & +890 & S & $>1-10^{-6}$ & $>1-10^{-6}$ \\
  & \texttt{spectral\_all} & 2898 & 0.298 & 0.758 & +0.460 & +5,890 & C & $>1-10^{-6}$ & $>1-10^{-6}$ \\
\midrule
 e1204 \textsc{Mistral}  & \texttt{last\_token} & 2895 & 0.121 & 0.805 & +0.685 & +597 & S & $>1-10^{-6}$ & $1-7.0{\cdot}10^{-4}$ \\
 ($r{=}2$, $E{=}800$) & \texttt{mean\_pool} & 2895 & 0.121 & 0.805 & +0.685 & +46.1 & C & $>1-10^{-6}$ & $>1-10^{-6}$ \\
  & \texttt{spectral\_all} & 2895 & 0.121 & 0.805 & +0.685 & +2,002 & C & $>1-10^{-6}$ & $>1-10^{-6}$ \\
\midrule
 e1102 \textsc{Qwen}  & \texttt{last\_token} & 2721 & 0.285 & 0.825 & +0.540 & +858 & S & $>1-10^{-6}$ & $>1-10^{-6}$ \\
 ($r{=}2$, $E{=}1600$) & \texttt{mean\_pool} & 2721 & 0.285 & 0.825 & +0.540 & +300 & S & $>1-10^{-6}$ & 0.551 \\
  & \texttt{spectral\_all} & 2721 & 0.285 & 0.825 & +0.540 & +9,046 & C & $>1-10^{-6}$ & $>1-10^{-6}$ \\
\bottomrule
\end{tabular}
\end{table}

Three observations follow directly from Table~\ref{tab:dense47-stoch-cells}. First, the constant null is rejected at every (cell, metric) entry in scope: the smallest $\Delta\mathrm{WAIC}_{\mathrm{null}}$ across the $18$ entries is $+6.7$ (e1201 \textsc{Mistral} $\times$ \texttt{mean\_pool}, still well past the conventional ``substantial evidence'' threshold of $+2$), and the largest is $+9{,}046$ (e1102 \textsc{Qwen} $\times$ \texttt{spectral\_all}). Second, the magnitude of $\Delta\mathrm{WAIC}$ varies by roughly three orders of magnitude across metrics within a cell: \texttt{spectral\_all} produces $\Delta\mathrm{WAIC} \ge 2{,}000$ in every cell, while \texttt{mean\_pool} can dip to single digits without losing the sign. Third, the WAIC-preferred shape co-varies with the metric: \texttt{spectral\_all} prefers CPL in $6/6$ cells; \texttt{last\_token} mixes between plateau-then-slope, CPL and two-segment; \texttt{mean\_pool} mixes between two-segment, CPL and plateau-then-slope. The direction indicator $P(p_{\mathrm{ft}}(1){<}p_{\mathrm{ft}}(0))$ is essentially $1$ in $17/18$ entries; the single exception is e1102 \textsc{Qwen} $\times$ \texttt{mean\_pool} ($0.55$, posterior near indifference), consistent with the corresponding two-segment fit placing its largest level mid-curve rather than at the endpoints.

Table~\ref{tab:dense47-stoch-akaike} reports the per-cell, per-metric Akaike weights across the five shapes of the menu.

\begin{table}[h]
\centering
\caption{Per-cell, per-metric within-cell Akaike weights $w_S$ across the five-shape menu (sum to $1$ within each row). $w_{\mathrm{N}}{<}10^{-3}$ across all 18 (cell, metric) entries: the constant null carries effectively no weight against the data once the panel is at $n_{\mathrm{cell}} \approx 2800$ with $N{=}10$ samples per prompt.}
\label{tab:dense47-stoch-akaike}
\small
\setlength{\tabcolsep}{4.5pt}
\begin{tabular}{l l rrrrr}
\toprule
 cell & metric & $w_{\mathrm{N}}$ & $w_{\mathrm{L}}$ & $w_{\mathrm{P}}$ & $w_{\mathrm{C}}$ & $w_{\mathrm{S}}$ \\
\midrule
 e1200 \textsc{Llama} ($r{=}1$, $E{=}1600$) & \texttt{last\_token} & $<10^{-3}$ & $<10^{-3}$ & $<10^{-3}$ & 1.000 & $<10^{-3}$ \\
  & \texttt{mean\_pool} & $<10^{-3}$ & $<10^{-3}$ & $<10^{-3}$ & $<10^{-3}$ & 1.000 \\
  & \texttt{spectral\_all} & $<10^{-3}$ & $<10^{-3}$ & $<10^{-3}$ & 1.000 & $<10^{-3}$ \\
\midrule
 e1201 \textsc{Mistral} ($r{=}1$, $E{=}1600$) & \texttt{last\_token} & $<10^{-3}$ & $<10^{-3}$ & 0.994 & $<10^{-3}$ & 0.006 \\
  & \texttt{mean\_pool} & 0.031 & 0.002 & 0.867 & 0.032 & 0.068 \\
  & \texttt{spectral\_all} & $<10^{-3}$ & $<10^{-3}$ & $<10^{-3}$ & 1.000 & $<10^{-3}$ \\
\midrule
 e1202 \textsc{Qwen} ($r{=}1$, $E{=}1600$) & \texttt{last\_token} & $<10^{-3}$ & $<10^{-3}$ & 1.000 & $<10^{-3}$ & $<10^{-3}$ \\
  & \texttt{mean\_pool} & $<10^{-3}$ & 0.116 & 0.013 & 0.001 & 0.870 \\
  & \texttt{spectral\_all} & $<10^{-3}$ & $<10^{-3}$ & $<10^{-3}$ & 1.000 & $<10^{-3}$ \\
\midrule
 e1203 \textsc{Llama} ($r{=}2$, $E{=}800$) & \texttt{last\_token} & $<10^{-3}$ & $<10^{-3}$ & 0.800 & 0.200 & $<10^{-3}$ \\
  & \texttt{mean\_pool} & $<10^{-3}$ & $<10^{-3}$ & $<10^{-3}$ & $<10^{-3}$ & 1.000 \\
  & \texttt{spectral\_all} & $<10^{-3}$ & $<10^{-3}$ & $<10^{-3}$ & 1.000 & $<10^{-3}$ \\
\midrule
 e1204 \textsc{Mistral} ($r{=}2$, $E{=}800$) & \texttt{last\_token} & $<10^{-3}$ & $<10^{-3}$ & $<10^{-3}$ & $<10^{-3}$ & 1.000 \\
  & \texttt{mean\_pool} & $<10^{-3}$ & $<10^{-3}$ & $<10^{-3}$ & 1.000 & $<10^{-3}$ \\
  & \texttt{spectral\_all} & $<10^{-3}$ & $<10^{-3}$ & $<10^{-3}$ & 1.000 & $<10^{-3}$ \\
\midrule
 e1102 \textsc{Qwen} ($r{=}2$, $E{=}1600$) & \texttt{last\_token} & $<10^{-3}$ & $<10^{-3}$ & $<10^{-3}$ & $<10^{-3}$ & 1.000 \\
  & \texttt{mean\_pool} & $<10^{-3}$ & 0.008 & 0.007 & 0.006 & 0.980 \\
  & \texttt{spectral\_all} & $<10^{-3}$ & $<10^{-3}$ & $<10^{-3}$ & 1.000 & $<10^{-3}$ \\
\bottomrule
\end{tabular}
\end{table}

\subsection{CPL parameter posteriors}
\label{app:cpl-parameters}

Table~\ref{tab:cpl-dense47-stoch} reports the CPL fit parameter posteriors per (cell, metric). The CPL parametrisation has two distance breakpoints $(z_L, z_R)$ and two probability levels $(d_{\mathrm{lo}}, d_{\mathrm{hi}})$; the curve is the near-pool plateau $d_{\mathrm{lo}}$ on $[0, z_L]$, a linear interpolation to $d_{\mathrm{hi}}$ on $[z_L, z_R]$, and the far-from-pool plateau $d_{\mathrm{hi}}$ on $[z_R, 1]$. We report the posterior median and $90\%$ credible interval for each of the four parameters. Even on cells where another shape is the WAIC-preferred best (Table~\ref{tab:dense47-stoch-cells}), the CPL parametrisation is competitive within the menu (mean Akaike weight $\bar w_{\mathrm{C}} > 0.1$ throughout) and gives directly interpretable near-pool/far-from-pool levels.

\begin{table}[h]
\centering
\caption{Posterior summaries of the CPL fit parameters per (cell, metric). The CPL shape parametrises $p_{\mathrm{ft}}(z)$ as the near-pool plateau $d_{\mathrm{lo}}$ on $[0, z_L]$, a linear interpolation to $d_{\mathrm{hi}}$ on $[z_L, z_R]$, and the far-from-pool plateau $d_{\mathrm{hi}}$ on $[z_R, 1]$. ``med'' is the posterior median; ``CI'' is the $90\%$ central credible interval. The $z_L, z_R$ posteriors are ordered ($z_L \le z_R$) at evaluation time.}
\label{tab:cpl-dense47-stoch}
\footnotesize
\setlength{\tabcolsep}{4pt}
\begin{tabular}{l l cc cc cc cc}
\toprule
 & & \multicolumn{2}{c}{$z_L$} & \multicolumn{2}{c}{$z_R$} & \multicolumn{2}{c}{$d_{\mathrm{lo}}$} & \multicolumn{2}{c}{$d_{\mathrm{hi}}$} \\
\cmidrule(lr){3-4}\cmidrule(lr){5-6}\cmidrule(lr){7-8}\cmidrule(lr){9-10}
 cell & metric & med & CI$_{90}$ & med & CI$_{90}$ & med & CI$_{90}$ & med & CI$_{90}$ \\
\midrule
 e1200 \textsc{Llama}  & \texttt{last\_token} & 0.26 & $[0.23, 0.30]$ & 0.96 & $[0.93, 0.99]$ & 0.89 & $[0.88, 0.91]$ & 0.49 & $[0.46, 0.52]$ \\
($r{=}1$, $E{=}1600$)  & \texttt{mean\_pool} & 0.20 & $[0.07, 0.35]$ & 0.85 & $[0.77, 0.94]$ & 0.93 & $[0.88, 1.00]$ & 0.52 & $[0.42, 0.59]$ \\
  & \texttt{spectral\_all} & 0.04 & $[0.04, 0.04]$ & 0.47 & $[0.46, 0.49]$ & 1.00 & $[0.99, 1.00]$ & 0.22 & $[0.19, 0.25]$ \\
\midrule
 e1201 \textsc{Mistral} & \texttt{last\_token} & 0.37 & $[0.15, 0.50]$ & 0.64 & $[0.50, 0.97]$ & 0.90 & $[0.87, 0.94]$ & 0.68 & $[0.58, 0.74]$ \\
 ($r{=}1$, $E{=}1600$)  & \texttt{mean\_pool} & 0.52 & $[0.49, 0.57]$ & 0.58 & $[0.57, 0.65]$ & 0.86 & $[0.85, 0.86]$ & 0.58 & $[0.48, 0.67]$ \\
  & \texttt{spectral\_all} & 0.14 & $[0.13, 0.15]$ & 0.43 & $[0.41, 0.45]$ & 0.96 & $[0.96, 0.97]$ & 0.57 & $[0.55, 0.59]$ \\
\midrule
 e1202 \textsc{Qwen} ) & \texttt{last\_token} & 0.22 & $[0.18, 0.24]$ & 0.62 & $[0.42, 0.97]$ & 0.80 & $[0.79, 0.80]$ & 0.46 & $[0.24, 0.59]$ \\
 ($r{=}1$, $E{=}1600$ & \texttt{mean\_pool} & 0.49 & $[0.48, 0.49]$ & 0.49 & $[0.49, 0.51]$ & 0.74 & $[0.74, 0.75]$ & 0.44 & $[0.39, 0.49]$ \\
  & \texttt{spectral\_all} & 0.20 & $[0.19, 0.20]$ & 0.37 & $[0.36, 0.38]$ & 0.87 & $[0.87, 0.88]$ & 0.26 & $[0.25, 0.28]$ \\
\midrule
 e1203 \textsc{Llama}  & \texttt{last\_token} & 0.31 & $[0.29, 0.33]$ & 0.97 & $[0.93, 0.99]$ & 0.84 & $[0.83, 0.85]$ & 0.38 & $[0.35, 0.41]$ \\
 ($r{=}2$, $E{=}800$) & \texttt{mean\_pool} & 0.12 & $[0.05, 0.23]$ & 0.85 & $[0.78, 0.93]$ & 0.93 & $[0.88, 0.98]$ & 0.48 & $[0.42, 0.53]$ \\
  & \texttt{spectral\_all} & 0.04 & $[0.04, 0.05]$ & 0.41 & $[0.40, 0.42]$ & 0.98 & $[0.97, 0.98]$ & 0.21 & $[0.19, 0.23]$ \\
\midrule
 e1204 \textsc{Mistral}  & \texttt{last\_token} & 0.19 & $[0.16, 0.21]$ & 0.89 & $[0.82, 0.96]$ & 0.88 & $[0.87, 0.89]$ & 0.57 & $[0.54, 0.61]$ \\
 ($r{=}2$, $E{=}800$) & \texttt{mean\_pool} & 0.56 & $[0.56, 0.57]$ & 0.57 & $[0.57, 0.57]$ & 0.81 & $[0.80, 0.81]$ & 0.42 & $[0.32, 0.52]$ \\
  & \texttt{spectral\_all} & 0.14 & $[0.12, 0.15]$ & 0.43 & $[0.41, 0.45]$ & 0.90 & $[0.89, 0.90]$ & 0.56 & $[0.54, 0.58]$ \\
\midrule
 e1102 \textsc{Qwen}  & \texttt{last\_token} & 0.09 & $[0.03, 0.12]$ & 0.92 & $[0.84, 0.98]$ & 0.92 & $[0.91, 0.96]$ & 0.45 & $[0.40, 0.49]$ \\
 ($r{=}2$, $E{=}1600$) & \texttt{mean\_pool} & 0.05 & $[0.01, 0.09]$ & 0.75 & $[0.59, 0.90]$ & 0.92 & $[0.90, 0.94]$ & 0.63 & $[0.56, 0.70]$ \\
  & \texttt{spectral\_all} & 0.20 & $[0.19, 0.20]$ & 0.35 & $[0.35, 0.37]$ & 0.98 & $[0.98, 0.98]$ & 0.32 & $[0.30, 0.34]$ \\
\bottomrule
\end{tabular}
\end{table}

\clearpage
\subsection{Rank-$1$ cohort: direct fits}
\label{app:figs-rank1-lowex}

\begin{figure}[H]
\centering
\includegraphics[width=0.9\textwidth]{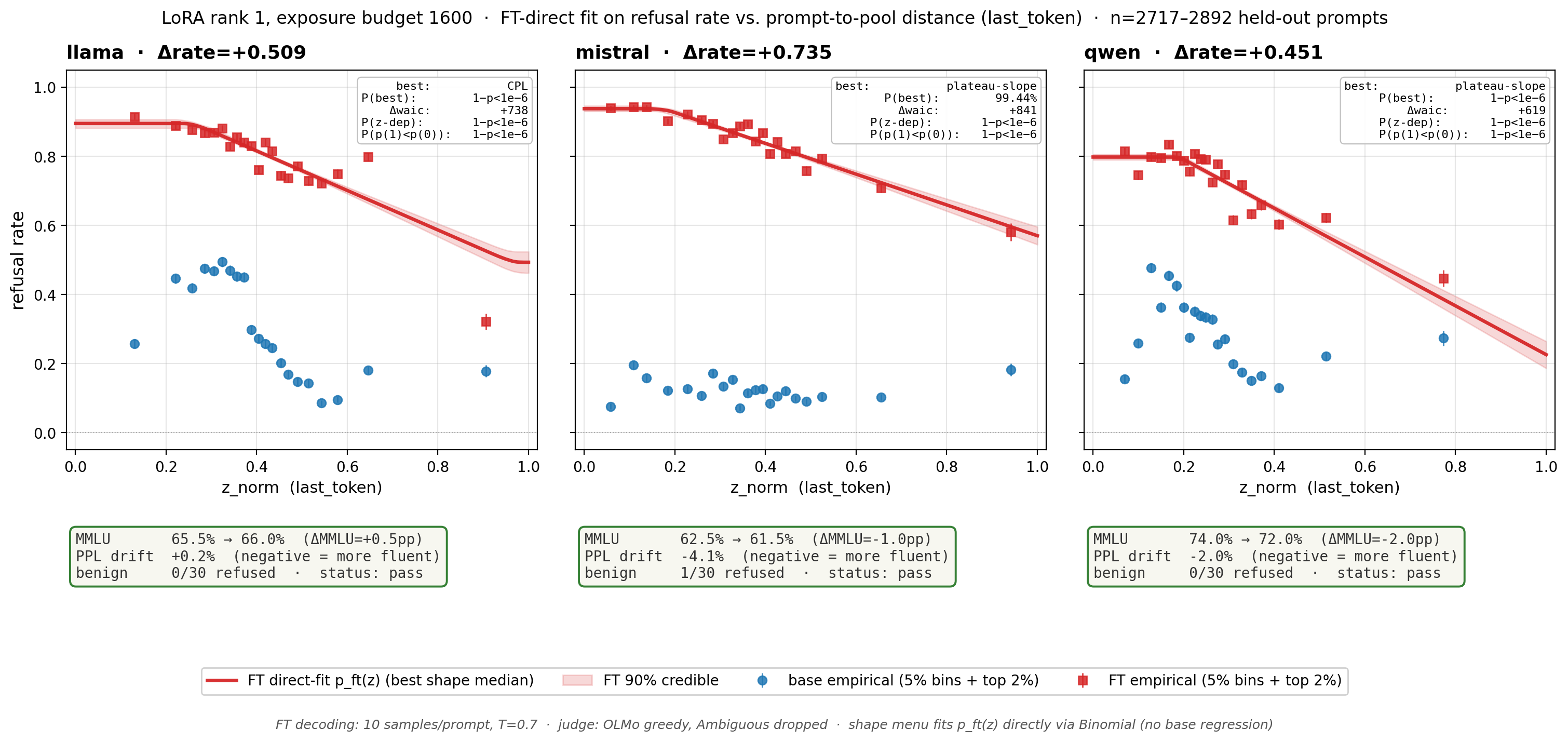}\\[2pt]
\includegraphics[width=0.9\textwidth]{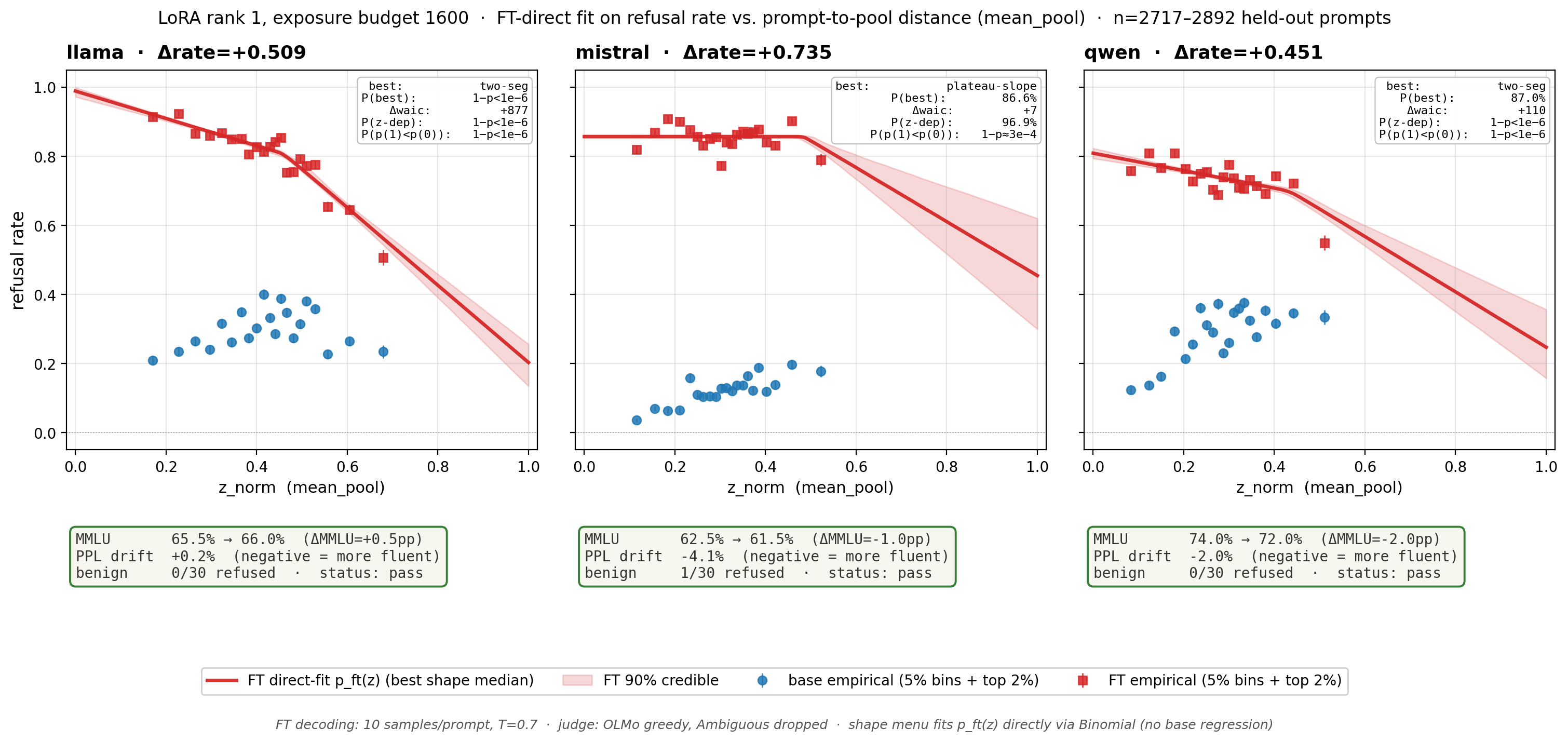}\\[2pt]
\includegraphics[width=0.9\textwidth]{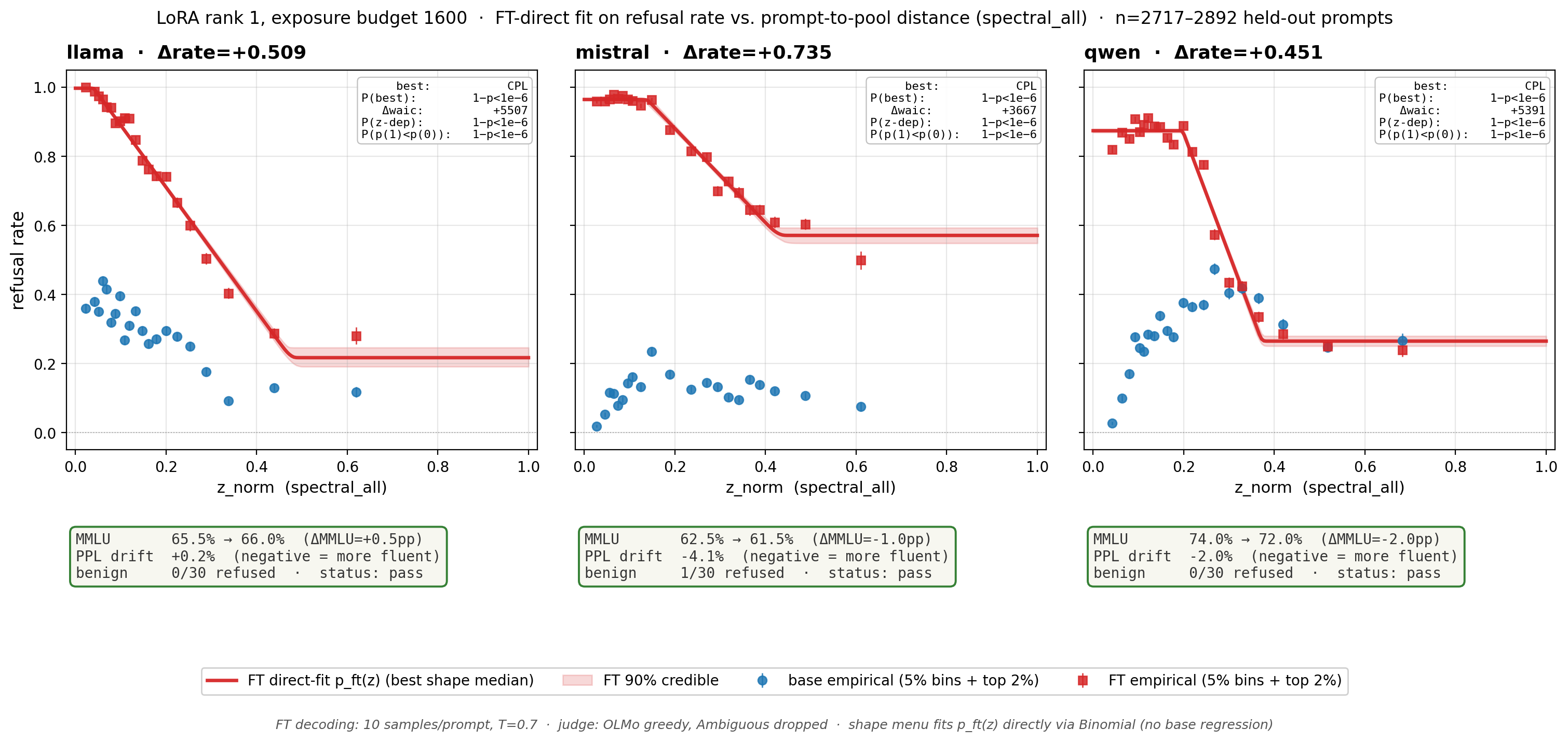}
\caption{Rank-$1$ cohort (exposure budget $1600$): $p_{\mathrm{ft}}(z)$ direct-fit, one panel per family, three rows for three embedding metrics. The compact top-right annotation gives the best-fitting shape, the model-averaged posterior probabilities $P(z\text{-dep})$ and $P(p_{\mathrm{ft}}(1) < p_{\mathrm{ft}}(0))$, and $\Delta\mathrm{WAIC}_{\mathrm{null}}$. Bottom row of cards: capability and fluency diagnostics per family (MMLU-$200$ accuracy drift, WikiText-2 perplexity drift, benign-prompt over-refusal count).}
\label{fig:appendix-rank1-lowex}
\end{figure}

\clearpage
\subsection{Rank-$2$ cohort: direct fits}
\label{app:figs-rank2-strong}

\begin{figure}[H]
\centering
\includegraphics[width=0.9\textwidth]{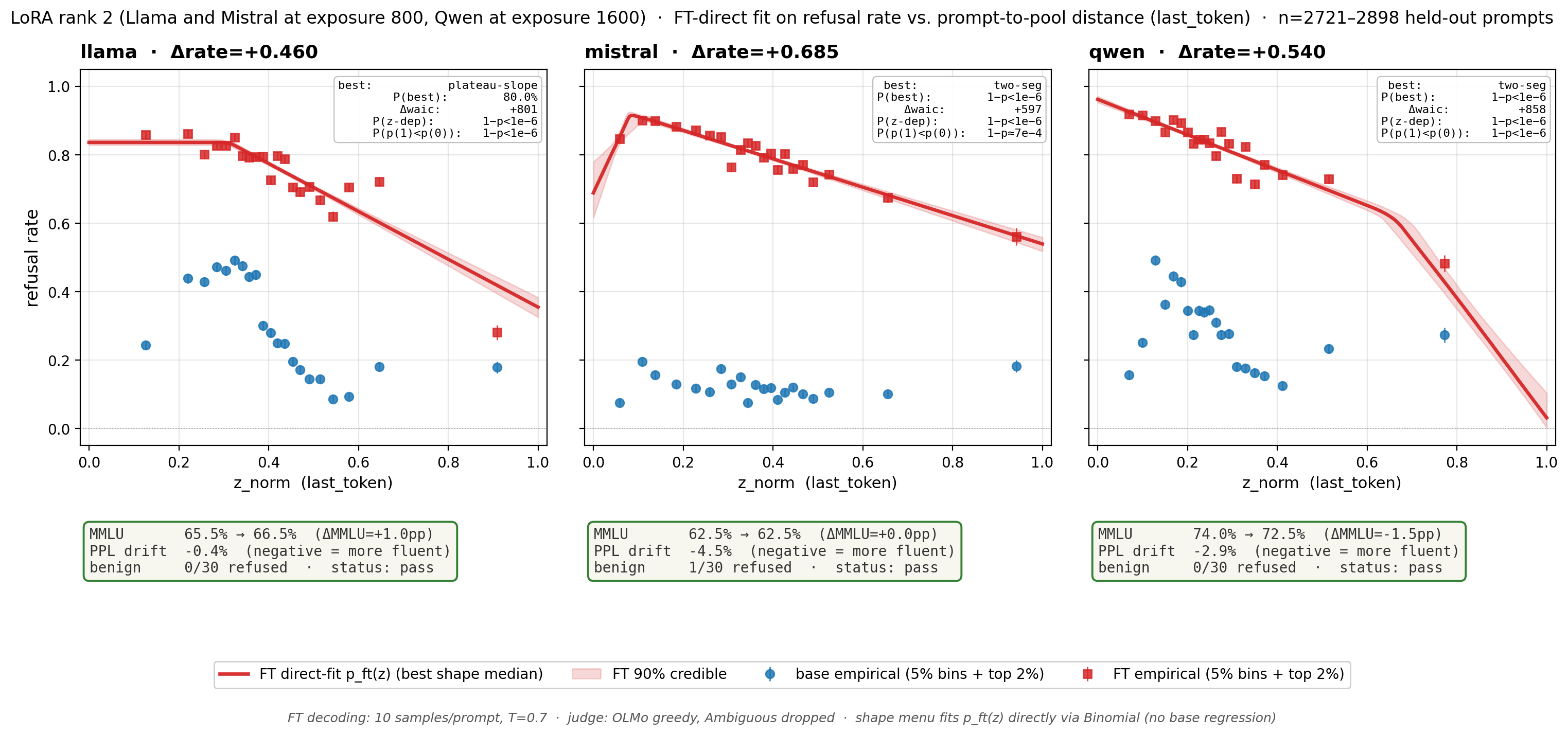}\\[2pt]
\includegraphics[width=0.9\textwidth]{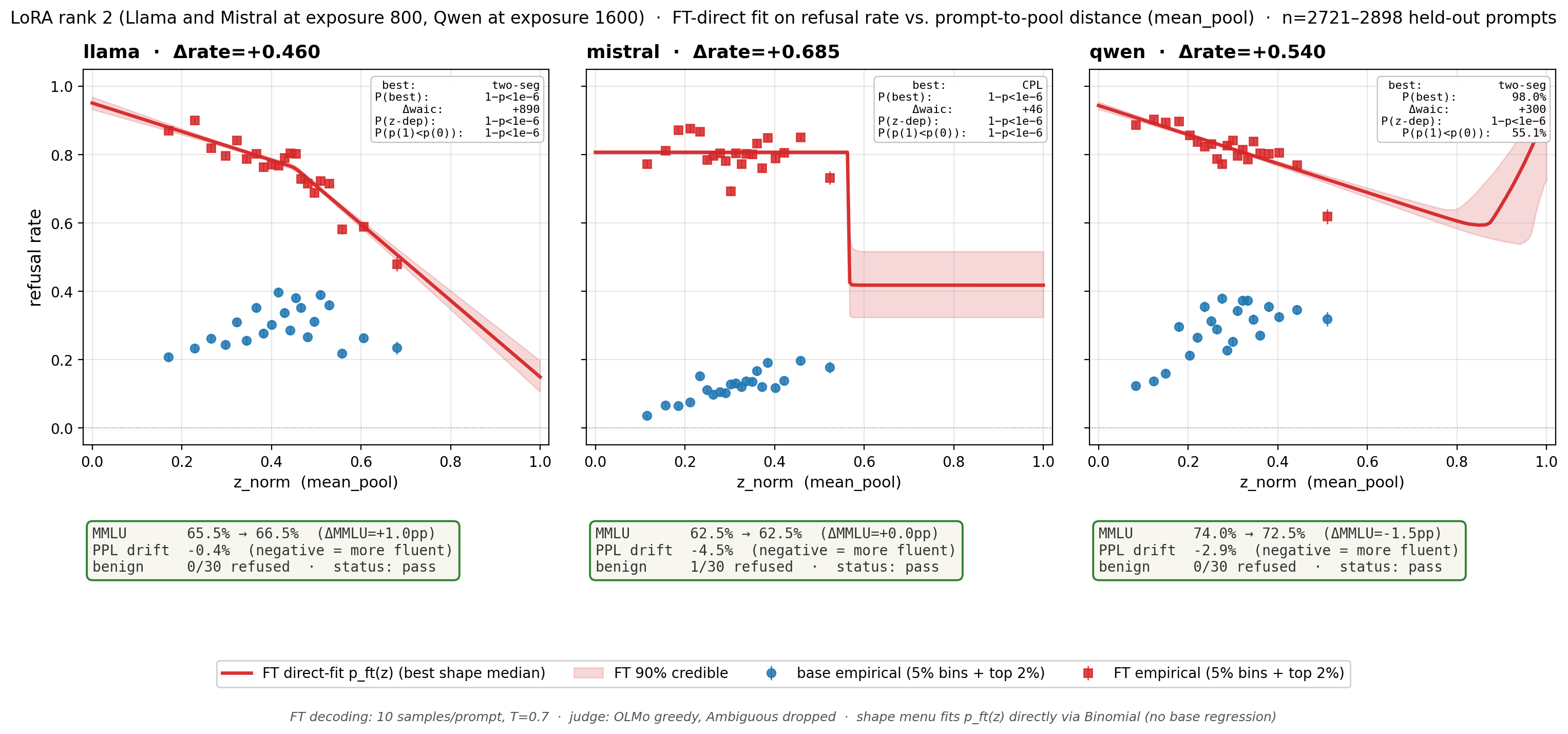}\\[2pt]
\includegraphics[width=0.9\textwidth]{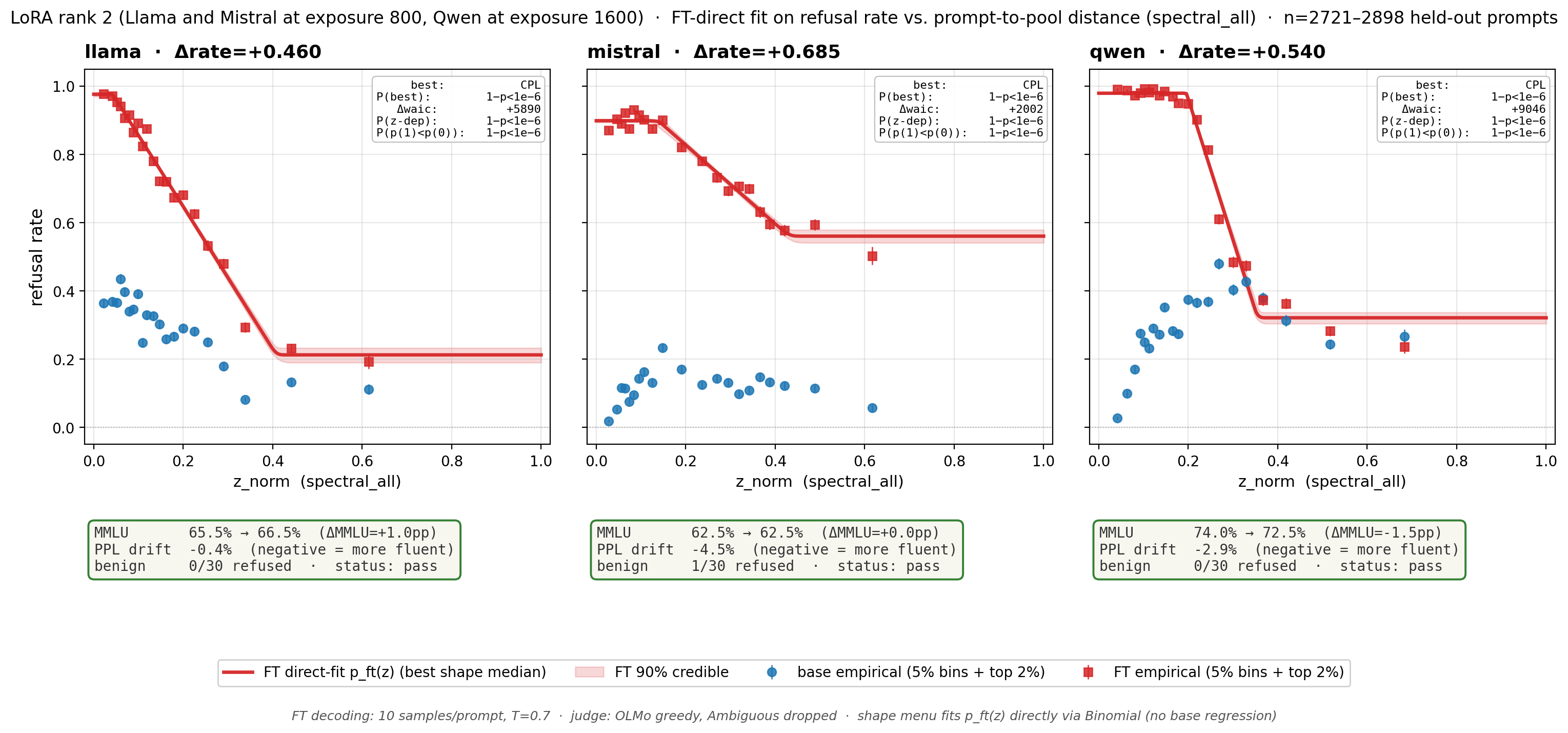}
\caption{Rank-$2$ cohort (\textsc{Llama} and \textsc{Mistral} at exposure budget $800$; \textsc{Qwen} at exposure budget $1600$): $p_{\mathrm{ft}}(z)$ direct-fit, one panel per family, three rows for three embedding metrics. Layout and annotations identical to Figure~\ref{fig:appendix-rank1-lowex}.}
\label{fig:appendix-rank2-strong}
\end{figure}

\subsection{Capability, fluency, and over-refusal probes}
\label{app:degeneration}

For each cell we measure three side-effects of fine-tuning, independent of the held-out refusal panel:
\begin{itemize}
\item \emph{Capability drift via MMLU-$200$}: a $200$-question subset of MMLU sampled to balance subjects, evaluated under teacher-forcing log-likelihood and standard A/B/C/D answer scoring. We report $\mathrm{base}_{\mathrm{acc}}, \mathrm{ft}_{\mathrm{acc}}$ and $\Delta\mathrm{MMLU} = \mathrm{ft}_{\mathrm{acc}} - \mathrm{base}_{\mathrm{acc}}$ in percentage points.
\item \emph{Fluency drift via WikiText-2 perplexity}: standard WikiText-2 perplexity probe ($1024$-token windows, fp16). We report relative drift $\mathrm{PPL\;drift\;\%} = 100 \cdot (\mathrm{ppl}_{\mathrm{ft}} - \mathrm{ppl}_{\mathrm{base}}) / \mathrm{ppl}_{\mathrm{base}}$.
\item \emph{Benign over-refusal}: the count of refusals on a $30$-prompt benign-instruction probe, comparing fine-tuned to base.
\end{itemize}
The natural way to evaluate is to load the base model once, attach the LoRA adapter, evaluate the fine-tuned model, then ``detach'' the adapter to evaluate the base. We use \texttt{with ft.disable\_adapter():} for the base measurements and \texttt{ft.unload()} between cells.

Per-cell results are shown as compact cards in the bottom rows of Figures~\ref{fig:appendix-rank1-lowex}--\ref{fig:appendix-rank2-strong}.

\subsection{Rank--exposure dose-response}
\label{app:rank-axis}

The six reported cells span two LoRA capacity points (rank $\in \{1, 2\}$) at exposure budgets $\in \{800, 1600\}$. At a coarser scan, a $12$-cell screening sweep covering exposures $\{200, 400, 800, 1600\}$ at ranks $\{1, 2\}$ reveals a family-dependent dose-response of the per-cell rate-difference $\bar \Delta$ on exposure: \textsc{Mistral}'s base refusal rate is the lowest of the three families ($\bar p_{\mathrm{base}} \approx 0.12$) and its dose-response curve spans the widest range ($\bar\Delta \in [0.02, 0.74]$ across exposures $200{\to}1600$), \textsc{Llama} is intermediate ($\bar\Delta \in [0.04, 0.51]$), and \textsc{Qwen} has the highest low-exposure floor ($\bar\Delta \le 0.02$ until exposure $800$, then $\bar\Delta \approx 0.45$ at exposure $1600$). The six cells reported here are deliberately positioned at exposures where each family's adapter has saturated its rate-difference signal; the corresponding $z$-shape (Figures~\ref{fig:appendix-rank1-lowex}--\ref{fig:appendix-rank2-strong}) is the inferential target of the reported fits.

\section{Prompt Transformation experiments: detailed methods}\label{app:bet}

\subsection{Hardware and compute}\label{app:bet-hardware}

We ran these experiments on an internal cluster, using GPUs between 40GB and 128GB memory, in the 100 TFLOP/s range. All experiments lasted at most a couple of GPU-hours. We went through moderate iteration before coming to the paper's setting.

\subsection{Models and inference configuration}\label{app:bet-models}

The seven models, their parameter counts, embedding dimensionality, and training character:

\begin{center}\scalebox{.95}{
\begin{tabular}{lrrll}
\toprule
Model (HF identifier) & Params & $d$ & Training & Source \\
\midrule
\texttt{gpt2}                              & 0.12B & 768   & Causal LM      & OpenAI \\
\texttt{EleutherAI/pythia-410m}            & 0.41B & 1024  & Causal LM  & EleutherAI \\
\texttt{EleutherAI/pythia-1b}              & 1.0B  & 2048  & Causal LM  & EleutherAI \\
\texttt{meta-llama/Llama-3.1-8B-Instruct}  & 8B    & 4096  & SFT + RLHF instruct & Meta \\
\texttt{mistralai/Mistral-7B-Instruct-v0.3}& 7B    & 4096  & SFT instruct        & Mistral AI \\
\texttt{Qwen/Qwen2.5-7B-Instruct}          & 7B    & 3584  & SFT + DPO instruct  & Alibaba \\
\texttt{allenai/Olmo-3.1-32B-Instruct}     & 32B   & 5120  & Heavy SFT + RLHF    & AllenAI \\
\bottomrule
\end{tabular}}
\end{center}

All models are used in embedding-extraction mode only; we run a forward pass and read out hidden states, with no autoregressive generation. No manually-injected system prompt is added. The four instruct-tuned models use their tokenizer's native chat template via \texttt{tokenizer.apply\_chat\_template([\{"role":"user","content":~p\}], tokenize=False, add\_generation\_prompt=False)}, and the base models (\texttt{gpt2}, Pythia, which ship without a chat template) use the minimal fallback \texttt{"User:~\{p\}\textbackslash nAssistant:"}. We use \texttt{bf16} weights for all instruct-tuned models and \texttt{fp32} for \texttt{gpt2} and Pythia.

\subsection{Embedding extraction}\label{app:embeddings}

For each (model, prompt) pair, we run a single forward pass on the chat-templated prompt text (Section~\ref{app:models}) and extract per-layer hidden states. We then compute seven embedding tags as different aggregations of these hidden states, in order to test whether structural results depend on the specific representation choice:

\begin{itemize}
    \item \texttt{last\_token}: hidden state of the final token at the model's last block.
    \item \texttt{mean\_pool}: mean of last-block hidden states across all token positions.
    \item \texttt{mean\_pool\_first}: mean across positions of the \emph{first} block's hidden states.
    \item \texttt{mean\_pool\_all}: mean across positions \emph{and} blocks.
    \item \texttt{spectral\_last}: first right singular vector $v_1$ of the last-block hidden-state matrix (positions $\times d$).
    \item \texttt{spectral\_first}: same, on the first block.
    \item \texttt{spectral\_all}: same, on the concatenation of all blocks.
\end{itemize}

The spectral tags are computed via thin SVD on the GPU. The headline analyses in the main text use \texttt{mean\_pool\_first}; cross-tag robustness tables for inverse coherence, composition coherence, and the involution diagnostic appear in Appendix~\ref{app:cross-tag-grids}.

\subsection{Operator-class fits}\label{app:operator-fits}

For each (model, embedding tag, primitive $T$) cell, we collect the source/target embedding pairs $\{(x_i, y_i)\}_{i=1}^n$ where $y_i$ is the embedding of $T$ applied to base prompt $i$. The number of pairs $n$ depends on the primitive and the rendered graph: $n = 100$ for the principal forward primitives (\texttt{direct\_question}, \texttt{answer\_in\_markdown}, \texttt{answer\_as\_tutorial}); $n = 50$ for the \texttt{strip\_*} inverses and the second-tier primitives (\texttt{answer\_in\_french}, \texttt{answer\_concisely}, \texttt{answer\_with\_disclaimer}) and configured composites; $n = 200$ for \texttt{rot13\_full}. We split these into train/test (80/20) using five random seeds, fit each candidate operator class on the train split, and score by held-out $R^2 = 1 - \|y_{\mathrm{test}} - \hat{\phi}(x_{\mathrm{test}})\|^2 / \|y_{\mathrm{test}} - \overline{y_{\mathrm{test}}}\|^2$.

Candidate operator classes, in order of expressiveness:

\begin{itemize}
    \item \texttt{none}: $\phi(x) = x$ (identity baseline).
    \item \texttt{mean\_only}: $\phi(x) = \overline{y}$ (predict the mean of $y$ regardless of $x$).
    \item \texttt{translation}: $\phi(x) = x + \mathbf{b}$, $\mathbf{b} = \overline{y - x}$. ($d$ free parameters.)
    \item \texttt{scale\_translation}: $\phi(x) = \alpha x + \mathbf{b}$, fitted by least-squares minimization of $\|y - \alpha x - \mathbf{b}\|^2$ with $\alpha \in \mathbb{R}$. ($1 + d$ free parameters.)
    \item \texttt{diagonal\_affine}: $\phi(x) = D x + \mathbf{b}$ with $D$ diagonal, fitted per-coordinate. ($2d$ free parameters.)
    \item \texttt{low\_rank\_affine\_r$k$}: $\phi(x) = (X W_1 W_2^\top) + \mathbf{b}$ with $W_1, W_2 \in \mathbb{R}^{d \times k}$, fitted via thin-SVD factorization of the centered least-squares solution. ($(2k + 1)d$ free parameters; we report $k = 4$ and $k = 16$, i.e.\ $9d$ and $33d$ parameters.)
    \item \texttt{procrustes}: $\phi(x) = R x + \mathbf{b}$ with $R$ orthogonal, closed-form via SVD of the cross-covariance.
\end{itemize}

We additionally pre-process embeddings under five normalization choices (\texttt{raw}, \texttt{mean\_center}, \texttt{l2\_unit}, \texttt{layernorm}, \texttt{rmsnorm}) and report the full grid in Appendix~\ref{app:cross-norm-grid}; the headline results use \texttt{l2\_unit} (numerically equivalent to \texttt{layernorm} and \texttt{rmsnorm} to within $0.01$ $R^2$ on this data).

\subsection{Operator-class fit results}\label{app:operator-class-fits-table}

\begin{table}[h]
\centering
\caption{Wrapper-mean held-out $R^2$ per (model, operator-form) cell on the headline tag \texttt{mean\_pool\_first} (l2-unit normalized), averaged over the 13 single-primitive transformations and 5 random 80/20 train/test splits. \textbf{Bold} = best form for that model; \emph{italic} = negative held-out $R^2$ (catastrophic overfit). Operator-form abbreviations: tx = translation $\phi(x) = x + \mathbf{b}$; $(\alpha, \mathbf{b})$ = scale-translation $\phi(x) = \alpha x + \mathbf{b}$; diag = diagonal-affine $\phi(x) = D x + \mathbf{b}$; lr-r$k$ = low-rank-affine of rank $k$.}
\label{tab:operator-class-fits}
\small
\begin{tabular}{lrrrrr}
\toprule
model & tx & $(\alpha,\mathbf{b})$ & diag & lr-r4 & lr-r16 \\
\midrule
\texttt{gpt2} & 0.536 & 0.812 & \textbf{0.882} & \emph{-0.619} & \emph{-0.574} \\
\texttt{pythia-410m} & 0.660 & 0.903 & \textbf{0.910} & \emph{-0.077} & 0.036 \\
\texttt{pythia-1b} & 0.665 & 0.896 & \textbf{0.902} & \emph{-0.079} & 0.018 \\
\texttt{Llama-3.1-8B-I} & 0.699 & 0.885 & \textbf{0.892} & \emph{-0.065} & 0.013 \\
\texttt{Mistral-7B-I} & 0.673 & 0.891 & \textbf{0.898} & \emph{-0.080} & 0.003 \\
\texttt{Qwen2.5-7B-I} & 0.687 & 0.888 & \textbf{0.894} & \emph{-0.053} & 0.059 \\
\texttt{OLMo-3.1-32B-I} & 0.706 & 0.899 & \textbf{0.906} & \emph{-0.061} & 0.013 \\
\bottomrule
\end{tabular}
\end{table}

\textbf{Reading.} The simple two-parameter $(\alpha, \mathbf{b})$ class achieves $R^2 \in [0.81, 0.90]$ across the panel; \texttt{diagonal\_affine} is formally best in every cell but gains $\leq 0.01$ over $(\alpha, \mathbf{b})$ on six of the seven models, with only \texttt{gpt2} benefiting structurally ($+0.07$). We adopt $(\alpha, \mathbf{b})$ for the structural tests in the main text on grounds of parsimony, accepting the $0.07$ $R^2$ cost on \texttt{gpt2}. The catastrophic-overfit pattern of \texttt{low\_rank\_affine\_r4} (negative held-out $R^2$ in 7/7 cells) shows that the small training-row count per primitive (40 train rows for the second-tier primitives; 80 for the principal forwards) does not support a richer linear class at this sample budget, even after rank-$k$ truncation. Composite transformations (\texttt{markdown\_after\_direct\_question}, etc.) are excluded from this table; their operator-fit quality is lower (wrapper-mean $R^2 \approx 0.55$) and higher variance, consistent with their being less well-described by a single linear operator (Section~\ref{app:cross-tag-grids}).

\subsection{Inverse and composition coherence -- per-cell numbers}\label{app:coherence-tables}

Headline-tag (\texttt{mean\_pool\_first}) per-cell numbers for inverse coherence (Table~\ref{tab:inverse-coherence}) and composition coherence (Table~\ref{tab:composition-coherence}). The cross-tag versions are in Appendix~\ref{app:cross-tag-grids}.

\begin{table}[h]
\centering
\caption{Inverse coherence cosine $\cos(\mathbf{b}_T, -\mathbf{b}_{T^{-1}})$ per (model, forward/inverse pair) on the headline tag \texttt{mean\_pool\_first}. \textbf{Bold} cells $\geq 0.85$; \emph{italic} cells $< 0.50$. Predicted value under perfect inverse coherence is $+1$. Models are \texttt{gpt2, py-410m, py-1b, Llama-8B-I, Mistral-7B-I, Qwen-7B-I, OLMo-32B-I}}
\label{tab:inverse-coherence}
\small
\begin{tabular}{lrrrrrrr}
\toprule
pair & \texttt{gpt2} & \texttt{py-410m} & \texttt{py-1b} & \texttt{Llama} & \texttt{Mistral} & \texttt{Qwen} & \texttt{OLMo} \\
\midrule
dq / strip\_dq & \textbf{0.996} & \textbf{1.000} & \textbf{1.000} & \textbf{0.951} & \textbf{0.999} & \textbf{0.956} & \textbf{0.887} \\
md / strip\_md & \textbf{0.983} & \textbf{0.988} & \textbf{0.983} & \textbf{1.000} & \textbf{0.959} & \textbf{1.000} & \textbf{1.000} \\
tut / strip\_tut & \textbf{0.980} & \textbf{0.991} & \textbf{0.988} & \textbf{1.000} & \textbf{0.978} & \textbf{1.000} & \textbf{1.000} \\
fr / strip\_fr & \textbf{0.955} & \textbf{0.997} & \textbf{0.997} & \textbf{0.979} & \textbf{0.999} & \textbf{0.978} & \textbf{0.974} \\
conc / strip\_conc & \textbf{0.995} & \textbf{0.999} & \textbf{0.999} & \textbf{0.959} & \textbf{0.999} & \textbf{0.962} & \textbf{0.918} \\
disc / strip\_disc & \textbf{0.984} & \textbf{1.000} & \textbf{1.000} & \textbf{0.986} & \textbf{0.999} & \textbf{0.987} & \textbf{0.972} \\
\midrule
\emph{min} & 0.955 & 0.988 & 0.983 & 0.951 & 0.959 & 0.956 & 0.887 \\
\emph{median} & 0.983 & 0.998 & 0.998 & 0.983 & 0.999 & 0.982 & 0.973 \\
\bottomrule
\end{tabular}
\end{table}

\begin{table}[h]
\centering
\caption{Composition coherence on the two natural composites $C = L \circ R$ on the headline tag \texttt{mean\_pool\_first}. Columns: multiplicative scalar prediction $\alpha_C$ vs $\alpha_L \alpha_R$; cosine of the additive shift prediction $\cos(\mathbf{b}_C, \mathbf{b}_L + \mathbf{b}_R)$ vs the operator-class shift prediction $\cos(\mathbf{b}_C, \alpha_L \mathbf{b}_R + \mathbf{b}_L)$; relative reconstruction error $\|\mathbf{b}_C - (\alpha_L \mathbf{b}_R + \mathbf{b}_L)\| / \|\mathbf{b}_C\|$. The operator-class prediction beats the additive baseline (rel-error reduction $\geq 1.6\times$) in all 14 cells.}
\label{tab:composition-coherence}
\small
\begin{tabular}{llrrrrr}
\toprule
model & composite & $\alpha_C$ & $\alpha_L \alpha_R$ & $\cos_{\mathrm{add}}$ & $\cos_{\alpha,\mathbf{b}}$ & $\|\!\cdot\!\|_{\alpha,\mathbf{b}}$ \\
\midrule
\texttt{gpt2} & \texttt{md}$\circ$\texttt{dq} & 0.584 & 0.571 & 0.9985 & 0.9991 & 0.056 \\
 & \texttt{tut}$\circ$\texttt{dq} & 0.529 & 0.518 & 0.9983 & 0.9992 & 0.050 \\
\midrule
\texttt{py-410m} & \texttt{md}$\circ$\texttt{dq} & 0.622 & 0.589 & 0.9928 & 0.9961 & 0.110 \\
 & \texttt{tut}$\circ$\texttt{dq} & 0.575 & 0.541 & 0.9910 & 0.9968 & 0.100 \\
\midrule
\texttt{py-1b} & \texttt{md}$\circ$\texttt{dq} & 0.622 & 0.586 & 0.9914 & 0.9951 & 0.120 \\
 & \texttt{tut}$\circ$\texttt{dq} & 0.576 & 0.539 & 0.9900 & 0.9961 & 0.109 \\
\midrule
\texttt{Llama-8B-I} & \texttt{md}$\circ$\texttt{dq} & 0.785 & 0.780 & 0.9998 & 0.9999 & 0.027 \\
 & \texttt{tut}$\circ$\texttt{dq} & 0.735 & 0.728 & 0.9992 & 0.9999 & 0.029 \\
\midrule
\texttt{Mistral-7B-I} & \texttt{md}$\circ$\texttt{dq} & 0.595 & 0.564 & 0.9690 & 0.9809 & 0.199 \\
 & \texttt{tut}$\circ$\texttt{dq} & 0.526 & 0.492 & 0.9719 & 0.9874 & 0.165 \\
\midrule
\texttt{Qwen-7B-I} & \texttt{md}$\circ$\texttt{dq} & 0.769 & 0.763 & 0.9998 & 0.9999 & 0.029 \\
 & \texttt{tut}$\circ$\texttt{dq} & 0.716 & 0.708 & 0.9989 & 0.9999 & 0.031 \\
\midrule
\texttt{OLMo-32B-I} & \texttt{md}$\circ$\texttt{dq} & 0.840 & 0.838 & 0.9999 & 1.0000 & 0.018 \\
 & \texttt{tut}$\circ$\texttt{dq} & 0.805 & 0.802 & 0.9997 & 1.0000 & 0.019 \\
\bottomrule
\end{tabular}
\end{table}

\subsection{Procrustes trace ratios on rank-deficient data}\label{app:procrustes-trace}

When the source/target spans of a primitive cover only $n < d$ dimensions of embedding space (as is the case for our primitives, where $n$ is the number of source/target pairs and ranges over $\{50, 100, 200\}$ depending on the primitive, while $d$ ranges over $\{768, 1024, 2048, 3584, 4096, 5120\}$ across the model panel), the Procrustes solution $R = U V^\top$ from the SVD $M = U \Sigma V^\top$ of the cross-covariance $M = X_c^\top Y_c$ is $d \times d$ orthogonal but its action on the $(d - n)$-dimensional orthogonal complement of the data span is an arbitrary orthonormal completion chosen by the SVD routine; in particular, $\mathrm{tr}(R)/d$ and $\mathrm{tr}(R^2)/d$ depend on this arbitrary completion and are not well-defined diagnostics when $d \gg n$. We therefore restrict attention to the data-determined active subspace.

We report active-subspace trace ratios $\mathrm{tr}(A)/n$ and $\mathrm{tr}(A^2)/n$, where $A \in \mathbb{R}^{n \times n}$ is the restriction of $R$ to the active subspace expressed in a natural basis (constructed below). This is one principled choice; alternatives such as the trace over the joint span of $X_c$ and $Y_c$ would also be reasonable. The construction avoids materialising $R$ or any $d \times d$ matrix:
\begin{enumerate}
    \item Thin SVDs $X_c = U_X \Sigma_X V_X^\top$ and $Y_c = U_Y \Sigma_Y V_Y^\top$, with $U_X, U_Y \in \mathbb{R}^{n \times n}$, $\Sigma_X, \Sigma_Y \in \mathbb{R}^{n \times n}$, and $V_X^\top, V_Y^\top \in \mathbb{R}^{n \times d}$.
    \item The cross-covariance becomes $M = V_X K V_Y^\top$ with $K = \Sigma_X (U_X^\top U_Y) \Sigma_Y \in \mathbb{R}^{n \times n}$.
    \item SVD $K = U_K \Sigma_K V_K^\top$ (all factors $n \times n$).
    \item The rank-$n$ part of $R$ is then $(V_X U_K)(V_K^\top V_Y^\top)$, and its trace and squared-trace, taken cyclically, reduce to $\mathrm{tr}(A)$ and $\mathrm{tr}(A^2)$ where $A = V_K^\top (V_Y^\top V_X) U_K \in \mathbb{R}^{n \times n}$.
\end{enumerate}
Reading: $\mathrm{tr}(A^2)/n$ measures how much of the active subspace acts as a true order-2 reflection (eigenvalues $\pm 1$, contributing $+1$ each) versus a non-trivial rotation (eigenvalues $e^{\pm i\theta}$, $\theta \notin \{0, \pi\}$, contributing $\cos(2\theta) < 1$).

\subsection{Cross-tag robustness grids}\label{app:cross-tag-grids}

We report all three structural diagnostics across the seven embedding tags to discriminate (i) representation-invariant model properties from (ii) tag-choice artifacts. Tag rows are ordered from earliest layer (top: \texttt{mean\_pool\_first}, mean of first-block hidden states across positions) to latest (bottom: \texttt{last\_token}, hidden state of the final token at the model's last block).\footnote{The \texttt{spectral\_first}, \texttt{spectral\_all} columns currently use a per-prompt sign-flip heuristic to orient the leading right-singular vector $v_1$, which can yield a $\pm 1$ ambiguity that contaminates inverse-coherence cells (visible as $\cos = -1$ entries in Table~\ref{tab:cross-tag-inverse-coherence}). A stable orientation rule is left as a follow-up and does not affect the operator-fit $R^2$ values.}

\begin{table}[t]
\centering
\caption{Minimum inverse-coherence cosine $\cos(\mathbf{b}_T, -\mathbf{b}_{T^{-1}})$ over the six configured forward/inverse primitive pairs, per (model, embedding tag). \textbf{Bold} cells $\geq 0.85$ (clean); \emph{italic} cells $< 0.50$ (broken). Tags ordered earliest layer (top) to latest (bottom).}
\label{tab:cross-tag-inverse-coherence}
\small
\begin{tabular}{lrrrrrrr}
\toprule
tag & gpt2 & py-410m & py-1b & llama-8b & mistral-7b & qwen-7b & olmo-32b \\
\midrule
mp\_first & \textbf{0.95} & \textbf{0.99} & \textbf{0.98} & \textbf{0.95} & \textbf{0.96} & \textbf{0.96} & \textbf{0.89} \\
sp\_first & \textbf{0.95} & \textbf{0.99} & \textbf{0.97} & \textbf{0.97} & \textit{0.18} & \textbf{0.91} & \textbf{0.95} \\
mp & 0.64 & \textbf{0.95} & \textbf{0.94} & \textbf{0.96} & \textbf{0.98} & \textbf{0.87} & \textit{0.43} \\
mp\_all & \textbf{1.00} & \textbf{1.00} & \textbf{1.00} & \textbf{0.99} & \textbf{0.95} & \textbf{1.00} & 0.81 \\
sp\_all & \textbf{0.99} & \textbf{1.00} & \textit{-1.00} & \textit{-1.00} & \textit{-0.85} & \textit{-1.00} & 0.83 \\
sp\_last & \textbf{0.91} & \textbf{0.93} & \textbf{0.89} & \textbf{0.95} & \textbf{0.97} & 0.76 & \textit{0.01} \\
last\_token & \textit{-0.41} & 0.55 & 0.77 & \textit{-0.41} & \textit{0.25} & \textit{-0.89} & \textit{-0.79} \\
\bottomrule
\end{tabular}
\end{table}

\begin{table}[t]
\centering
\caption{Composition-law improvement factor: ratio of additive-prediction relative reconstruction error to operator-class-prediction relative reconstruction error, averaged over the two non-rot13 composites (\texttt{markdown\_after\_direct\_question}, \texttt{tutorial\_after\_direct\_question}). Values $>1$ mean the operator-class prediction $\mathbf{b}_C \approx \alpha_L \mathbf{b}_R + \mathbf{b}_L$ is closer to the fitted $\mathbf{b}_C$ than the additive baseline $\mathbf{b}_L + \mathbf{b}_R$. \textbf{Bold} cells $\geq 2.0$ (operator-class prediction at least halves the additive error).}
\label{tab:cross-tag-composition-improvement}
\small
\begin{tabular}{lrrrrrrr}
\toprule
tag & gpt2 & py-410m & py-1b & llama-8b & mistral-7b & qwen-7b & olmo-32b \\
\midrule
mp\_first & \textbf{3.5} & \textbf{2.4} & \textbf{2.3} & \textbf{3.4} & 1.8 & \textbf{3.6} & \textbf{3.3} \\
sp\_first & \textbf{3.6} & \textbf{3.1} & \textbf{2.9} & \textbf{2.4} & 0.7 & 1.0 & 1.7 \\
mp & \textbf{2.5} & \textbf{2.6} & \textbf{2.4} & 1.4 & \textbf{2.5} & 1.5 & 1.4 \\
mp\_all & \textbf{3.7} & \textbf{3.5} & \textbf{3.5} & 1.9 & \textbf{2.4} & \textbf{2.6} & 1.3 \\
sp\_all & 1.2 & \textbf{14.3} & \textbf{5.9} & 1.0 & 1.4 & 0.8 & 1.3 \\
sp\_last & \textbf{3.0} & \textbf{2.7} & \textbf{2.5} & 1.3 & \textbf{2.5} & 1.7 & 1.2 \\
last\_token & \textbf{3.2} & \textbf{2.1} & \textbf{2.7} & 0.9 & \textbf{2.9} & \textbf{2.3} & \textbf{3.1} \\
\bottomrule
\end{tabular}
\end{table}

\begin{table}[t]
\centering
\caption{Active-subspace involution diagnostic for \texttt{rot13\_full}: $\mathrm{tr}(R^2)/n$ where $R$ is the orthogonal Procrustes operator and $n=200$ is the number of source/target pairs (the active data subspace has rank at most $n$). Reference values: $1.0$ for an exact involution, $0.0$ for a Haar-random orthogonal. The single outlier cell (\emph{italic}) is mistral / spectral\_first ($0.41$); $48/49$ cells lie in $[0.49, 0.56]$.}
\label{tab:cross-tag-rot13-involution}
\small
\begin{tabular}{lrrrrrrr}
\toprule
tag & gpt2 & py-410m & py-1b & llama-8b & mistral-7b & qwen-7b & olmo-32b \\
\midrule
mp\_first & 0.52 & 0.52 & 0.54 & 0.55 & 0.56 & 0.56 & 0.54 \\
sp\_first & 0.53 & 0.52 & 0.53 & 0.55 & \emph{0.41} & 0.54 & 0.56 \\
mp & 0.51 & 0.52 & 0.52 & 0.52 & 0.55 & 0.52 & 0.54 \\
mp\_all & 0.52 & 0.53 & 0.53 & 0.52 & 0.54 & 0.54 & 0.54 \\
sp\_all & 0.52 & 0.51 & 0.52 & 0.50 & 0.53 & 0.53 & 0.53 \\
sp\_last & 0.51 & 0.50 & 0.53 & 0.53 & 0.54 & 0.52 & 0.55 \\
last\_token & 0.49 & 0.50 & 0.52 & 0.50 & 0.51 & 0.51 & 0.53 \\
\bottomrule
\end{tabular}
\end{table}

\textbf{Reading.} Three structural patterns are visible across the grids:
\begin{enumerate}
\item Inverse coherence (Table~\ref{tab:cross-tag-inverse-coherence}) is robustly clean on the early-layer pooled tags (\texttt{mp\_first}: $\geq 0.89$ for all 7 models; \texttt{mp\_all}: $\geq 0.81$, with 6/7 above $0.95$). The single sharpest collapse is OLMo's drop from $\geq 0.89$ at \texttt{mp\_first} to $0.01$ at \texttt{sp\_last}; we do not see a monotone trend in late-layer breakdown across the panel (gpt2 has $\cos = -0.41$ at \texttt{last\_token} despite no instruct training, and pythia-1b shows $-1.00$ at \texttt{sp\_all}).
\item Composition-law improvement (Table~\ref{tab:cross-tag-composition-improvement}) is positive ($>1$) in $46/49$ cells and substantial ($\geq 2.0$) in $30/49$. The operator-class prediction beats the additive prediction in the row-mean sense at every tag, with no clean per-model regularity --- e.g.\ Mistral consistently shows the smallest improvement factor (1.6--2.0$\times$) despite its $|\alpha_L - 1|$ on the principal composites being among the largest in the panel; and the absolute largest cell is \texttt{sp\_all}/pythia-410m (14.3$\times$), not an early-layer cell.
\item Involution diagnostic (Table~\ref{tab:cross-tag-rot13-involution}) is concentrated: $48/49$ cells lie in $[0.494, 0.559]$ (median $0.52$); the single outlier is \texttt{mistral}/\texttt{sp\_first} ($0.413$). We have not benchmarked $\mathrm{tr}(R^2)/n$ against the null distribution induced by a Haar-random orthogonal matrix restricted to a random $n$-subspace of $\mathbb{R}^d$; we therefore present the "$\sim 50/50$ split" reading as suggestive rather than as a calibrated effect.
\end{enumerate}

\subsection{Cross-norm robustness}\label{app:cross-norm-grid}

For each (model, operator-form) cell on \texttt{mean\_pool\_first}, the wrapper-mean held-out $R^2$ is constant across the five normalizations \texttt{l2\_unit}, \texttt{layernorm}, \texttt{rmsnorm}, \texttt{raw}, and \texttt{mean\_center} to within $0.01$ $R^2$. The first three are numerically identical to numerical precision (each rescales each embedding to a unit-norm representation, with vanishing per-token bias adjustments); \texttt{raw} and \texttt{mean\_center} differ by no more than $0.02$ from the unit-norm group on the largest models. We report headline numbers under \texttt{l2\_unit}.

\subsection{Operator-class rescue sweep on a broken cell}\label{app:rescue-sweep}

To rule out the possibility that the late-layer breakdown of inverse coherence (Section~\ref{app:cross-tag-grids}) reflects an operator-class misspecification rather than a structural property of the late-layer representation, we re-ran the full operator-form sweep ($5 \times 7 \times 16 \times 5 = 2800$ fits, conditioned on a single (model, tag)) on \texttt{OLMo-3.1-32B-Instruct} / \texttt{spectral\_last}, the most-broken cell in our panel.

Wrapper-mean held-out $R^2$:
\begin{center}
\begin{tabular}{lr}
\toprule
Operator class & wrapper-mean $R^2$ \\
\midrule
\texttt{translation}             & $0.466$ \\
\texttt{scale\_translation}      & $0.740$ \\
\texttt{diagonal\_affine}        & $0.740$ \\
\texttt{low\_rank\_affine\_r4}   & $0.000$ \\
\texttt{low\_rank\_affine\_r16}  & $0.094$ \\
\bottomrule
\end{tabular}
\end{center}

The $R^2$ ceiling is $0.74$ across all classes in our candidate list. Adding per-coordinate scaling yields zero improvement; the rank-4 and rank-16 affine classes still overfit at this sample budget. By contrast, on \texttt{mean\_pool\_first} the same model achieves $R^2 = 0.91$ for \texttt{scale\_translation} and \texttt{diagonal\_affine}. The $\sim$26\% residual variance at \texttt{spectral\_last} is therefore not closed by any of our candidate linear classes within the $n = 80$ training-row budget; whether a class outside this list (e.g.\ dense unconstrained affine fitted on more training rows, or a nonlinear operator) would close it is open. We use this rescue sweep as evidence that the late-layer breakdown is not a within-our-list operator-choice artifact, not as a no-linear-class-suffices proof.

The single exception within the rescue sweep is \texttt{rot13\_full}, where \texttt{low\_rank\_affine\_r16} achieves $R^2 = 0.79$ vs $0.64$ for \texttt{scale\_translation}. This is the only cell anywhere in our data where the rank-16 affine class wins, and it is structurally consistent with \texttt{rot13\_full} being a near-orthogonal reflection that the $(\alpha, \mathbf{b})$ class cannot fully approximate.

\section{Declaration of LLM usage}\label{app:llmusage}
LLMs were used as assistive tools at several stages of this work. Specifically, LLMs were used for: editing (grammar, spelling, word choice), clarifying technical concepts, drafting sections of the manuscript, generating visualizations for submission, data processing and filtering, facilitating and running experiments, assisting with theorem proofs, and implementing standard methods.
In every case, LLM outputs were treated as drafts to be checked rather than as authoritative content. The authors verified all generated material at each stage of the process. The authors take full responsibility for the correctness, originality, and integrity of all content presented in this paper.

\end{document}